\documentclass[11pt, a4paper]{report}

\usepackage[catalan, english, spanish]{babel}
\usepackage[utf8]{inputenc}
\usepackage[T1]{fontenc}

\usepackage[globalproofs,plaineqn,fancyexamples,framed, spanish]{theorems}
\usepackage{graphicx, color}
\usepackage{colortbl}
\usepackage{dsfont}

\usepackage{amssymb}
\usepackage{mathabx}

\newcommand{\paginablanco}{\newpage\thispagestyle{empty}~\newpage}
\definecolor{green}{rgb}{0,0.7,0}
\definecolor{blue}{rgb}{0,0,0.5}
\definecolor{red}{rgb}{0.8,0.2,0.2}
\definecolor{orange}{rgb}{0.8,0.5,0.2}
\definecolor{violet}{rgb}{0.5,0.2,0.5}

\begin{document}

\thispagestyle{plain}

% Estilo de cabecera y pie de página que se utiliza
\newcommand{\autor}{Javier Insa Cabrera}
\newcommand{\director}{José Hernández Orallo}
\newcommand{\titulo}{Análisis de primeros prototipos de tests de inteligencia universales: evaluando y comparando algoritmos de IA y seres humanos}
\newcommand{\logo}{\begin{figure}[h!]\centering\includegraphics[width=0.7\textwidth]{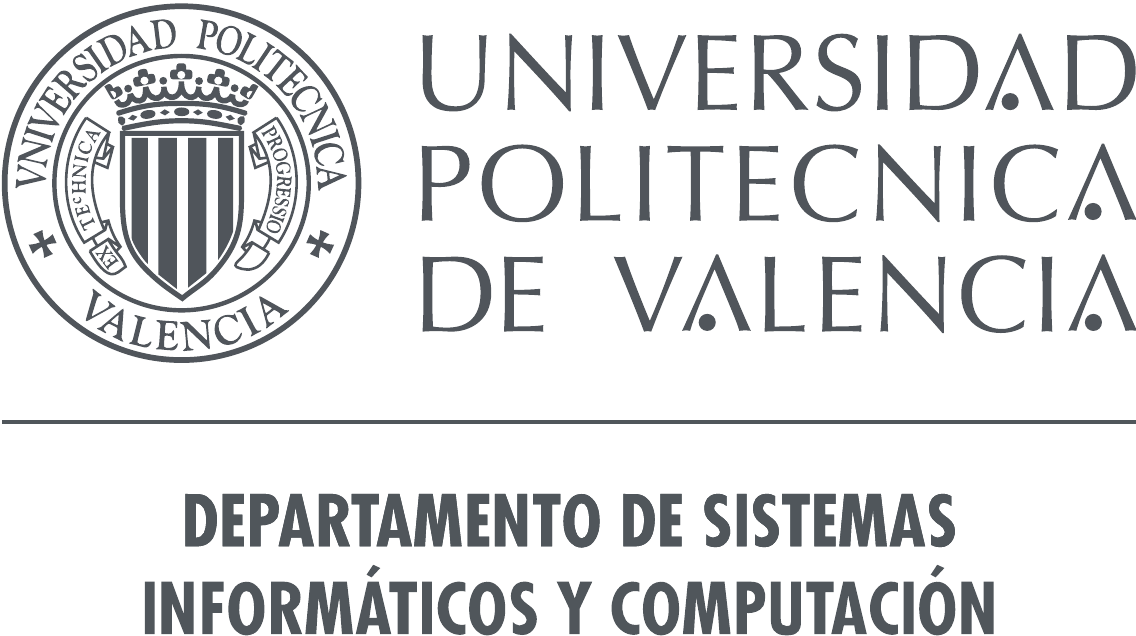}\end{figure}}
\newcommand{\fecha}{\today}

\thispagestyle{empty}

\begin{center}
\logo

\vspace{3cm}
{\setlength{\baselineskip}{2\baselineskip} \LARGE \titulo \par}
\end{center}

\vspace{2.5cm}
\begin{center}
%\begin{flushright}
\begin{tabular}{rl}
\multicolumn{2}{c}{\Large Tesis de máster}\\[0.75cm]
\large Autor: & \large \autor\\[0.25cm]
\large Director: & \large \director
\end{tabular}
%\end{flushright}
\end{center}

\vspace{2.5cm}
\begin{flushright}
Valencia, \fecha
\end{flushright}

\newpage
\thispagestyle{empty}
~
\newpage

% Comienza el resumen en castellano al estar como idioma seleccionado
\chapter*{Resumen}
\pagenumbering{roman}
La Inteligencia Artificial (IA) siempre ha intentado emular la mayor virtud del ser humano: su \emph{inteligencia}. Sin embargo, y aunque han habido multitud de esfuerzos por conseguirlo, a simple vista nos percatamos de que su inteligencia apenas llega a asemejarse a la de los seres humanos.

Los métodos existentes hoy en día para evaluar la inteligencia de la IA, se centran en utilizar técnicas empíricas para medir el rendimiento de los algoritmos en ciertas tareas \emph{concretas} (e.g., jugar al ajedrez, resolver un laberinto o aterrizar un helicóptero). No obstante, estos métodos no resultan apropiados si lo que queremos es evaluar la inteligencia general de la IA y, aun menos, si pretendemos compararla con la de los seres humanos.

En el proyecto ANYNT se ha diseñado un nuevo método de evaluación que trata de evaluar la IA a partir de \emph{nociones computacionales} bien conocidas utilizando \emph{problemas lo más generales posibles}.

Este nuevo método sirve para evaluar la inteligencia general (la cual nos permite aprender a resolver cualquier nuevo tipo de problema al que nos enfrentamos) y no para evaluar únicamente el rendimiento en un conjunto de tareas específicas. Este método no solo se centra en medir la inteligencia de algoritmos, sino que está pensado para poder evaluar cualquier tipo de sistema inteligente (seres humanos, animales, IA, ¿extraterrestres?, \dots) y a la vez situar sus resultados en una misma escala y, por lo tanto, poder compararlos.

Este nuevo enfoque de evaluación de la IA permitirá (en un futuro) evaluar y comparar cualquier tipo de sistema inteligente conocido o aun por construir/descubrir ya sean artificiales o biológicos.

En esta tesis de máster pretendemos comprobar que este nuevo método ofrece unos resultados coherentes al evaluar algoritmos de IA mediante el diseño e implementación de prototipos de test de inteligencia universales y su aplicación a diferentes sistemas inteligentes (algoritmos de IA y seres humanos). Del estudio analizamos si los resultados obtenidos por dos sistemas inteligentes distintos se encuentran correctamente situados en la misma escala y se proponen cambios y refinamientos a estos prototipos con el fin de, en un futuro, poder lograr un test de inteligencia verdaderamente universal.

\paginablanco

% Cambia el idioma activo al inglés
\selectlanguage{english}
% Comienza el resumen en inglés al estar como idioma seleccionado
\chapter*{Abstract}
Artificial Intelligence (AI) has always tried to emulate the greatest virtue of humans: their \emph{intelligence}. However, although there have been many efforts to reach this goal, at a glance we notice that the intelligence of AI systems barely resembles that of humans.

Today, available methods that assess AI systems are focused on using empirical techniques to measure the performance of algorithms in some \emph{specific} tasks (e.g., playing chess, solving mazes or land a helicopter). However, these methods are not appropriate if we want to evaluate the general intelligence of AI and, even less, if we compare it with human intelligence.

The ANYNT project has designed a new method of evaluation that tries to assess AI systems using well known \emph{computational notions} and \emph{problems which are as general as possible}.

This new method serves to assess general intelligence (which allows us to learn how to solve any new kind of problem we face) and not only to evaluate performance on a set of specific tasks. This method not only focuses on measuring the intelligence of algorithms, but also to assess any intelligent system (human beings, animals, AI, aliens?, \dots), and letting us to place their results on the same scale and, therefore, to be able to compare them.

This new approach will allow us (in the future) to evaluate and compare any kind of intelligent system known or even to build/find, be it artificial or biological.

This master thesis aims at ensuring that this new method provides consistent results when evaluating AI algorithms, this is done through the design and implementation of prototypes of universal intelligence tests and their application to different intelligent systems (AI algorithms and humans beings). From the study we analyze whether the results obtained by two different intelligent systems are properly located on the same scale and we propose changes and refinements to these prototypes in order to, in the future, being able to achieve a truly universal intelligence test.

\paginablanco

% Cambia el idioma activo al catalán
\selectlanguage{catalan}
% Comienza el resumen en valenciano al estar como idioma seleccionado
\chapter*{Resum}
La Intel·ligència Artificial (IA) sempre ha intentat emular la major virtut dels éssers humans: la seua \emph{intel·ligència}. No obstant això, i encara que hi ha hagut multitud d'esforços per aconseguir-ho, a simple vista ens adonem que la seua intel·ligència apenes arriba a ssemblar-se a la dels éssers humans.

Els mètodes existents hui en dia per avaluar la intel·ligència dels sistemes d'IA, se centren en utilitzar tècniques empíriques per a mesurar el rendiment dels algoritmes en certes tasques \emph{concretes} (e.g., jugar als escacs, resoldre un laberint o aterrar un helicòpter). No obstant això, aquests mètodes no resulten apropiats si el que volem és avaluar la intel·ligència general dels sistemes d'IA i, encara menys, si pretenem comparar-la amb la dels éssers humans.

En el projecte ANYNT s'ha dissenyat un nou mètode d'avaluació que tracta d'avaluar la IA a partir de \emph{nocions computacionals} ben conegudes utilitzant \emph{problemes el més generals possibles}.

Aquest nou mètode servix per avaluar la intel·ligència general (la qual ens permet aprendre a resoldre qualsevol nou tipus de problema al qual ens enfrontem) i no per avaluar únicament el rendiment en un conjunt de tasques específiques. Aquest mètode no només es centra en mesurar la intel·ligència d'algoritmes, sinó que està pensat per a poder avaluar qualsevol tipus de sistema intel·ligent (éssers humans, animals, IA, ¿extraterrestres?, ...) i al mateix temps situar els seus resultats en una mateixa escala i, per tant, poder comparar-los.

Aquest nou enfocament d'avaluació de la IA permetrà (en un futur) avaluar i comparar qualsevol tipus de sistema intel·ligent conegut o inclús per construir/descubrir ja siguen artificials o biològics.

En aquesta tesi de màster pretenem comprovar que aquest nou mètode oferix uns resultats coherents en avaluar algoritmes d'IA per mitjà del disseny i la implementació de prototips de test d'intel·ligència universals i la seua aplicació a diferents sistemes intel·ligents (algoritmes d'IA i éssers humans). De l'estudi analitzem si els resultats obtinguts per dos sistemes intel·ligents distints es troben correctament situats en la mateixa escala i es proposen canvis i refinaments a aquests prototips a fi de, en un futur, poder aconseguir un test d'intel·ligència verdaterament universal.

% Cambia el idioma activo al español
\selectlanguage{spanish}

\paginablanco

% Índice general
\tableofcontents

\paginablanco

\chapter{Introducción}
\pagenumbering{arabic}
El objetivo de la \emph{Inteligencia Artificial} (IA) ha sido, desde siempre, dotar de comportamiento a las máquinas (o algoritmos) para que lleguen a realizar tareas que en los seres humanos requieren de inteligencia. El fin último es que parezcan más inteligentes (o al menos desde nuestro punto de vista).

Uno de los objetivos propuestos por la IA es que los algoritmos sean capaces de \emph{aprender} a resolver nuevos problemas por sí solos en lugar de que se les tenga que asistir continuamente cada vez que se enfrenten a un nuevo problema. En el momento en el que se construyan estos algoritmos capaces de aprender, habremos construido un sistema al que realmente podremos calificar como \emph{inteligente}.

Uno de los principales problemas que existe en la IA, es que no existe ningún método para medir o evaluar esta inteligencia, y si existen, no están enfocados a medir propiamente la inteligencia, sino más bien el rendimiento que tienen los algoritmos en aprender a resolver ciertas tareas predefinidas.

En lugar de construir mejores métodos para evaluar a los algoritmos, la disciplina se ha centrado casi exclusivamente en diseñar nuevos algoritmos (o mejorar los ya existentes) de aprendizaje, planificación, reconocimiento de formas, razonamiento, etc, para que funcionen mejor en las tareas en las que se les va a evaluar. Para ello, los diseñadores \emph{ajustan} sus algoritmos para que obtengan buenos resultados en estas tareas.

Aunque cada vez se obtienen mejores resultados en estas evaluaciones, no por ello significa que haya habido algún avance significativo en la IA \emph{general}, sino más bien, que los diseñadores de algoritmos comprenden cada vez mejor los detalles y entresijos de las tareas a las que se enfrentan sus algoritmos y cómo deben \emph{ajustarlos} para que obtengan mejores resultados.

\section{Motivación}\label{sec:motivacion}
Entre los distintos métodos que existen hoy en día para evaluar la inteligencia, encontramos a los tests psicométricos. Estos tests tienen una larga historia \cite{spearman1904general}, son efectivos, fáciles de administrar, rápidos y bastante estables cuando se utilizan sobre un mismo individuo (humano) a través del tiempo. De hecho, han proporcionado una de las mejores definiciones prácticas de inteligencia: ``la inteligencia es lo que se mide con los tests de inteligencia''. Sin embargo, los tests psicométricos son antropomórficos: no pueden evaluar la inteligencia de otros sistemas que no sean el Homo sapiens, también son estáticos y están basados en un tiempo de prueba límite. Nuevas aproximaciones en la psicometría, como la `Teoría de Respuesta al Ítem' o en inglés `Item Response Theory' (IRT), permiten la selección de items basándose en la habilidad estimada y adaptando el test al nivel del individuo que se está examinando. La Teoría de Respuesta al Ítem es una herramienta prometedora, pero estos modelos aun no son la corriente principal de estos tests en psicometría (\cite{embretson2000psychometric}).

Este es uno de varios esfuerzos que han intentado establecer ``a priori'' cómo debería ser un test de inteligencia (p.~ej. \cite{embretson1998cognitive}) y a partir de ahí encontrar adaptaciones para distintos tipos de sujetos. Sin embargo, se necesitan (en general) diferentes versiones de los tests psicométricos para evaluar niños de distintas edades y para evaluar adultos con diversas patologías, ya que los tests psicométricos para los Homo sapiens adultos confían en que los sujetos posean de antemano ciertas habilidades y conocimientos.

Lo mismo ocurre para otros animales. Los psicólogos comparativos y otros científicos en el área de la cognición comparada normalmente diseñan tests específicos para distintas especies. Podemos ver un ejemplo de estos tests especializados para niños y chimpancés en \cite{Herrmann-etal07}. También se ha podido comprobar que los tests psicométricos no funcionan para las máquinas en el estado actual de la inteligencia artificial \cite{sanghidowe2003computer}, ya que se pueden engañar utilizando programas relativamente simples y especializados. Sin embargo, la principal desventaja de los tests psicométricos para evaluar sujetos distintos a los humanos, es que no poseen tras ellos una definición matemática.

Alan M. Turing propuso el primer test de inteligencia para máquinas \cite{turing1950}, el juego de la imitación (comúnmente conocido como el test de Turing) \cite{oppydowe2008}. En este test, a un sistema se le considera inteligente si es capaz de imitar a un humano (p.~ej. ser indistinguible de un humano) durante un periodo de tiempo y siguiendo un tema en un diálogo (teletexto) con uno o más jueces. Aunque ha sido ampliamente aceptado como referencia para comprobar si la inteligencia artificial alcanzará la inteligencia de los humanos, ha generado amplios debates y se han sugerido muchas variantes y alternativas \cite{oppydowe2008}. El test de Turing y las ideas relacionadas presentan varios problemas para ser un test de inteligencia para máquinas: (1) el test de Turing es antropomórfico (mide la humanidad, no la inteligencia), (2) no es gradual (no proporciona una puntuación), (3) no es práctico (cada vez es más fácil engañar a los jueces sin demostrar inteligencia y requiere mucho tiempo para obtener evaluaciones fiables), y (4) requiere de un juez humano.

Una aproximación más reciente y singular de un test de inteligencia para máquinas es el llamado CAPTCHA `Completely Automated Public Turing test to tell Computers and Humans Apart' (o en castellano `test de Turing público completamente automatizado para distinguir entre humanos y computadores') \cite{von2004telling}\cite{von2008recaptcha}. Los CAPTCHAs son problemas de reconocimiento de caracteres donde las letras aparecen distorsionadas y estas distorsiones hacen difícil que las máquinas (bots) puedan reconocer las letras. El objetivo inmediato de un CAPTCHA es distinguir entre humanos y máquinas, mientras que el objetivo final es prevenir que los bots y otro tipo de máquinas o programas sean capaces de crear cuentas, publicar comentarios u otro tipo de tareas que sólo los humanos deberían poder hacer. El problema con los CAPTCHAs es que cada vez se están haciendo más y más difíciles para los humanos, ya que los bots se están especializando y mejorando en ser capaces de leerlos. Cada vez que se desarrolla una nueva técnica CAPTCHA, aparecen nuevos bots capaces de pasar el test. Esto fuerza a los desarrolladores de CAPTCHA a cambiarlos una y otra vez, y así continuamente. A pesar de que los CAPTCHAs funcionan razonablemente bien hoy en día, en unos 10 o 20 años, se necesitarán hacer las cosas tan difíciles y generales que los humanos necesitarán más tiempo y varios intentos para conseguir resolverlos.

También existen otras propuestas menos conocidas para la medición de la inteligencia, muchas de las cuales son informales o meramente filosóficas, y ninguna de ellas ha sido puesta en práctica. Desde un punto de vista ingenieril, también han habido varias propuestas. Podemos destacar la serie de `workshops on Performance Metrics' para sistemas inteligentes (ver p.~ej. \cite{madhavan2009performance}) mantenido desde el 2000, el cual típicamente se refiere a escenarios, sistemas o aplicaciones muy limitadas o especializadas. Y, finalmente, ha habido un uso impreciso y general del término `Cociente de Inteligencia de Máquinas' o en inglés Machine Inteligence Quotient (MIQ). Se ha utilizado de distintas maneras en \cite{zadeh1994fuzzy}\cite{zadeh2008toward}\cite{ulinwa2008machine}, especialmente en el área de los sistemas difusos, pero sin ninguna definición precisa. En cualquier caso, la noción de MIQ es inapropiada debido a que el cociente\footnote{Originalmente el cociente era un cociente real de la edad mental estimada de un niño en comparación con su edad biológica. Hoy en día la `Q' en IQ es una reliquia histórica, el cociente es una puntuación normalizada de acuerdo a las puntuaciones de una población, típicamente asumiendo una distribución normal con una media igual a 100. Esta normalización es mucho más difícil de hacer para máquinas, ya que no está claro sobre qué máquinas promediar. Es esta última interpretación \emph{relativa} del cociente de inteligencia y su aplicación en máquinas lo que criticamos aquí.} se obtiene en psicometría a partir de una población de individuos de una especie en cierta etapa dada de su desarrollo (niños, adultos), el cual es posible para humanos, pero no tiene sentido alguno para sistemas de inteligencia artificial ya que no existe ninguna noción de especie ni ninguna muestra homogénea.

\section{Carencias de las técnicas anteriores}
Todos los métodos de evaluación anteriores resultan inapropiados si queremos evaluar la inteligencia de algoritmos (o máquinas). Es cierto que con ellos es posible comprobar si una máquina parece o no humana (test de Turing) o si son capaces de reconocer caracteres igual de bien que los humanos (CAPTCHAs), pero resultan inútiles si queremos proporcionar siquiera una estimación de su inteligencia. Estas técnicas, como mucho, pueden darnos una idea general de algunas de sus capacidades, pero no nos ofrecen ningún valor de su capacidad de aprendizaje.

Por otro lado, los `workshops on Perfornance Metrics' sí tratan de evaluar la inteligencia y la capacidad de aprendizaje de los algoritmos. El problema en este tipo de métricas de rendimiento, es que se suelen elegir problemas específicos sin ningún fundamento teórico tras su elección, por lo que, realmente, solo evalúan las capacidades necesarias para resolver estos problemas concretos y no las capacidades generales de los algoritmos. Sin embargo, este tipo de evaluaciones siguen siendo útiles, ya que permiten evaluar si los algoritmos son capaces de resolver y desenvolverse correctamente en tareas específicas de la vida real, pero dejan de ser útiles cuando tratamos de evaluar su capacidad general de aprendizaje.

\section{Objetivos}\label{sec:Objetivos}
Durante los últimos años se han intentado construir algunos métodos de evaluación que midan la inteligencia general de los algoritmos. Entre ellos podemos encontrar el proyecto ANYNT\footnote{http://users.dsic.upv.es/proy/anynt/}, el cual ha diseñado un nuevo sistema de evaluación cuyo objetivo es el de evaluar a estos algoritmos construyendo tests a partir de nociones computacionales bien conocidas. Para conseguir evaluar la inteligencia general, en este proyecto se ha diseñado e implementado una clase de entornos lo más general posible, la cual es la encargada de proporcionar los entornos en donde se evaluará a los agentes.

El objetivo principal de esta tesis es comprobar si la clase de entornos desarrollada en el proyecto ANYNT realmente cumple con las expectativas puestas en ella. Para ello, comprobamos si se obtienen resultados coherentes al evaluar distintos sistemas inteligentes (en concreto se han evaluado a algunos humanos y se ha implementado y evaluado un algoritmo de IA conocido como Q-learning).

En esta tesis no intentamos comprobar si la clase de entornos implementada cumple o no con las propiedades formales co las que se diseñó. Estas propiedades ya fueron comprobadas en un trabajo previo\footnote{En este trabajo previo se implementó gran parte de la estructura de la aproximación a la clase de entornos y, posteriormente, se realizaron los experimentos oportunos para comprobar que estas propiedades se cumplían. Podemos ver la memoria de este trabajo en `http://users.dsic.upv.es/proy/anynt/Memoria.pdf'.} realizado en este proyecto, en donde ya se concluyó que la clase de entornos las posee.

En esta tesis hemos analizado si la clase de entornos desarrollada funciona (tal y como se espera de ella) en la práctica a partir de los siguientes experimentos:
\begin{enumerate}
\itemsep=0px
\item La clase de entornos resulta factible en la práctica para evaluar sistemas de IA. Para concluir esto, evaluamos un algoritmo de IA estándar y comprobamos si sus resultados en las tareas generadas con nuestra clase de entornos son coherentes.
\item La diferencia en inteligencia de distintos sistemas que se han evaluado utilizando estos tests, se corresponde con la diferencia real esperada. Para ello, aplicamos los mismos tests a humanos y algoritmos de IA y comparamos los resultados obtenidos por ambos sistemas.
\end{enumerate}

Podemos ver los resultados de estos experimentos en los \mbox{Capítulos \ref{cap:Evaluacion} y \ref{cap:EvaluacionDistintosAlgoritmos}} respectivamente.

\section*{Organización de este documento}
El resto de esta memoria está organizada del siguiente modo: En el Capítulo~\ref{cap:Precedentes} veremos los conceptos básicos que necesitamos conocer para comprender este trabajo (aprendizaje por refuerzo, complejidad Kolmogorov, algoritmos de Markov, \dots) y las distintas evaluaciones de inteligencia que se han venido realizando en seres humanos y algoritmos de IA hasta la fecha. En el Capítulo~\ref{cap:MarcoGeneral} veremos el marco general de los tests desarrollados en el proyecto ANYNT. En el Capítulo~\ref{cap:Aproximacion} describimos la aproximación implementada de estos test. En el Capítulo~\ref{cap:Evaluacion} evaluamos un algoritmo clásico de IA (Q-learning) para comprobar que la clase de entornos desarrollada funciona coherentemente en la práctica para este tipo de sistema inteligente. En el Capítulo~\ref{cap:EvaluacionDistintosAlgoritmos} evaluamos conjuntamente dos sistemas inteligentes distintos: el Homo sapiens (los humanos) y un algoritmo de IA (nuevamente el Q-learning) y analizamos si los resultados obtenidos concuerdan con la diferencia real en inteligencia entre ambos sistemas. El Capítulo~\ref{cap:Perspectivas} propone algunas ideas sobre hacia donde debería encaminarse el proyecto según los resultados obtenidos en esta tesis. Y, finalmente, en el Capítulo~\ref{cap:Conclusiones} veremos a qué conclusiones hemos llegado con este trabajo y cual es el trabajo futuro en este proyecto.

\paginablanco

\chapter{Precedentes}\label{cap:Precedentes}
En este capítulo damos a conocer los conceptos básicos necesarios para comprender el trabajo realizado en el proyecto ANYNT y hacemos un pequeño repaso a los distintos tipos de evaluaciones que se han estado realizando históricamente en humanos, otros animales y máquinas.

\section{Conceptos básicos}
\subsection{Entornos}
Los entornos, entre muchas otras definiciones, se pueden considerar ``mundos'' (con sus propias características y propiedades) en donde cierta cantidad de individuos (o \emph{agentes}) pueden interactuar. Los entornos disponen de un estado interno, el cual se le proporciona a los agentes como parte de su `percepción' del entorno. Además, conforme los agentes van interactuando con el entorno, éste puede cambiar su estado en función de sus reglas de comportamiento y de las acciones que realicen los agentes.

Estos entornos se suelen utilizar para modelar varios tipos de problemas. Imaginemos, por ejemplo, un laberinto. Los laberintos no son más que un espacio cerrado con un único punto de entrada y otro de salida y un conjunto de pasillos y paredes que nos permiten movernos o no a través del laberinto. En algunos de estos laberintos se necesitan objetos (como por ejemplo una llave para abrir una puerta) para conseguir salir de ellos. Según esta definición, cualquier entorno (natural o artificial) con un espacio en donde podamos identificar paredes que nos interrumpan el paso, pasillos que podamos traspasar y llaves que nos permitan abrir puertas será un laberinto. Pero los laberintos son algo más que un conjunto de paredes y pasillos. En realidad son un problema en donde se debe descubrir, a partir de la entrada, cuál es el camino que nos conduce a la salida, pudiendo utilizar (en algunos casos) objetos para conseguirlo. Además, según como se construya el espacio y la capacidad del agente para manejar este tipo de problemas, el problema del laberinto puede hacerse más o menos complejo.

Además de los laberintos, otros muchos problemas pueden definirse utilizando entornos, como p.~ej. juegos de mesa o aparcar/aterrizar/atracar un vehículo. Cada uno de estos problemas (a su vez) puede modelarse de forma más o menos compleja, haciéndolos más o menos difíciles de resolver: el tamaño de las pistas de aterrizaje, haciendo las reglas de los juegos de mesa más severas, \dots.

Estos entornos también pueden ser de distintos tipos; como p.~ej. los entornos sociales, en donde varios agentes interactúan a la vez. Estos entornos suelen ser más complejos y por lo tanto más difíciles de describir, ya que no solo contienen un espacio y varios objetos y agentes, sino que además, estos agentes suelen tener un comportamiento complejo (lo cual suele ser difícil de describir). Este tipo de entornos se suelen utilizar para describir problemas en donde los agentes deberán cooperar o rivalizar con otros agentes.

Antes hablábamos de que los agentes pueden modificar el estado de un entorno con sus acciones. Este tipo de entornos son los llamados reactivos y suponen un mayor problema para los agentes, ya que, no solo deben moverse por el entorno para conseguir su objetivo, sino que también necesitan comprender cómo se modifica el entorno en función de sus acciones para resolver el problema.

Ahora imaginemos un entorno en donde combinemos las características de los entornos sociales y los reactivos. En este caso tenemos un problema muchísimo mayor que los anteriores por separado. Ya no solo el agente es capaz de modificar el entorno, sino que el resto de agentes también son capaces de hacerlo. Por lo que en este caso el agente deberá comprender correctamente al comportamiento del resto de agentes que modifican el entorno a la vez que deberá adaptar el suyo propio para conseguir resolver el problema.

Como podemos ver, las capacidades de estos entornos son lo suficientemente amplias como para formular cualquier tipo de problema imaginable. Sin embargo, la elección del tipo de entorno que se utilice hace que los problemas que se formulen tengan cierto tipo de propiedades y se necesiten de habilidades concretas por parte de los agentes para conseguir resolverlos. Por lo tanto, cuanto más generales sean estos entornos, es de esperar que se necesiten de mayores habilidades por parte del agente para poder resolverlos.

\subsection{Aprendizaje por refuerzo}\label{sec:AprendizajeRefuerzo}
El objetivo del aprendizaje por refuerzo (RL) es utilizar premios y castigos (recompensas positivas y negativas) para que el agente aprenda a resolver un problema. En función de los refuerzos que perciba, el agente reconfigura su comportamiento modificando las acciones que realizará en el futuro para así obtener el mayor número posible de premios. Como resultado a este proceso, el agente ``aprende'' qué debe hacer en estos problemas para resolverlos correctamente. Dicho de forma breve: en el aprendizaje por refuerzo se trata de conseguir que un agente actúe en un entorno de manera que maximice las recompensas que obtiene por sus acciones.

\begin{figure}[h!]
\centering
\includegraphics[width=0.75\textwidth]{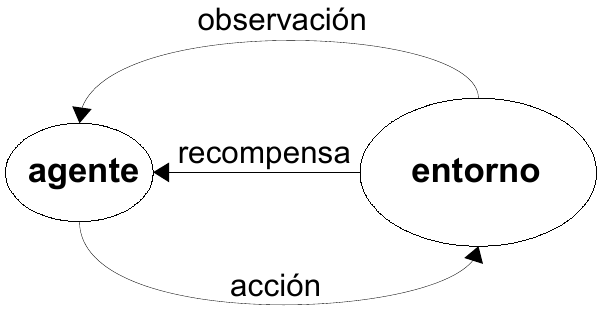}
\caption{Visión general de la interacción entre un agente y un entorno.}
\label{fig:Marco}
\end{figure}

La visión más general de este tipo de aprendizaje sitúa a un agente interactuando con un entorno a través de observaciones, acciones y recompensas. Generalmente se asumen entornos discretos; con tiempo, acciones y observaciones discretas\footnote{Es posible modelar muchos de los problemas de tiempo continuo utilizando entornos con tiempo ``discreto'', por lo que utilizar tiempos discretos no supone ninguna pérdida de generalidad.}. Las acciones están limitadas por un conjunto finito de símbolos $\cal{A}$ (p.~ej. $\{izquierda, derecha, arriba, abajo\}$), las recompensas se recogen a partir de cualquier subconjunto $\cal{R}$ de números racionales entre $0$ y $1$ y las observaciones también están limitadas por un conjunto finito $\cal{O}$ de posibilidades (p.~ej. una cuadrícula de celdas de \mbox{$n \times m$} con los objetos que se encuentran dentro). Las observaciones representan la situación actual del entorno, la cual puede ser completamente observable (existe una relación directa entre la observación que proporciona el entorno y su estado) o parcialmente observable (solo se observa parte del estado del entorno). En una interacción, una acción es la respuesta que realiza un agente a la observación que le ha proporcionado el entorno y, finalmente, el entorno le devolverá al agente una recompensa (o refuerzo) en función de su comportamiento. Estas interacciones de los agentes con los entornos se realizan en un tiempo discreto, de modo que en cada unidad de tiempo se realizará una interacción. Podemos ver una visión general de esta interacción en la Figura~\ref{fig:Marco}.

La señal de refuerzo puede ser inmediata o retardada; los refuerzos inmediatos representan una crítica para cada acción justo después de realizarla, por lo que la información aportada por el refuerzo en este caso es local a cada acción. Por el contrario, en el caso retardado, los refuerzos no se dan tras realizar cada acción, sino al completar la secuencia de acciones empleadas para resolver el problema, siendo (en este caso) una estimación global del comportamiento. También existen situaciones híbridas donde las recompensas positivas no son inmediatas tras realizar una acción en la buena dirección, pero tampoco existe una tarea o problema a resolver.

El agente observa el entorno para recoger información que le permita modificar su comportamiento, aprendiendo así a incrementar sus recompensas. Como dijimos anteriormente, el objetivo del aprendizaje por refuerzo es la utilización de las recompensas para que el agente ajuste su comportamiento (su función de agente). Por tanto, nuestro agente no será más que una función que, recibiendo como entrada una percepción del entorno, devolverá la siguiente acción a realizar.

Las aplicaciones del aprendizaje por refuerzo son múltiples, desde robots móviles que aprenden a salir de un laberinto, programas de ajedrez que aprenden cuáles son las mejores secuencias de movimientos para ganar un juego o un brazo robótico que aprende cómo mover sus articulaciones para lograr el movimiento final deseado.

Existen multitud de algoritmos para abordar el problema del aprendizaje por refuerzo. Algunos de estos algoritmos hacen suposiciones sobre los entornos en los que van a interactuar, como por ejemplo que sean `Procesos de decisión de Markov' en inglés `Markov Decision Process' (MDP), mientras que otros algoritmos tratan de abordar el problema contemplando entornos más generales. Uno de los algoritmos referentes en el aprendizaje por refuerzo es la técnica conocida como Q-learning \cite{watkins1992q}\cite{sutton1998reinforcement}.

\subsection[Q-learning]{Q-learning\protect\footnote{Esta definición de Q-learning ha sido obtenida de la Wikipedia.}}\label{sec:DefinicionQlearning}
Q-learning es una técnica de aprendizaje por refuerzo que trata de aprender una función acción-valor la cual indica la utilidad esperada al realizar una acción concreta en un estado dado. Uno de los puntos fuertes de Q-learning es que es capaz de comparar la utilidad esperada de las acciones disponibles sin la necesidad de tener un modelo del entorno, siendo a su vez una de sus limitaciones, ya que, al no disponer de un modelo, no es capaz de generalizar lo suficiente y no acaba de ``comprender'' bien el entorno con el que interactúa.

El modelo del problema consiste en un agente, un conjunto de estados ($S$) y un conjunto de acciones por estado ($A$). El agente puede moverse de un estado a otro realizando una acción y cada estado le proporciona al agente una recompensa (un número natural o real). Por lo que el objetivo del agente es maximizar su recompensa total y consigue esto aprendiendo qué acción ofrece la mayor recompensa para cada estado.

Por lo tanto, el algoritmo tiene una función que calcula la calidad (Quality) de una combinación estado-acción:

\[ Q : S \times A \rightarrow \mathds{R} \]

Antes de comenzar el aprendizaje, la matriz Q devuelve un valor prefijado elegido por el diseñador. Entonces, cuando el agente realiza una acción $a \in A$ en un estado $s \in S$ y se le proporcione la recompensa (el estado ha cambiado) se calcula el nuevo valor para ese estado y esa acción. El núcleo del algoritmo es una simple iteración de actualización de valores, en el cual se modifica el valor anterior haciendo una corrección basándose en la nueva información del siguiente modo:

\begin{centering}
$$Q(s_t, a_t) \leftarrow Q(s_t, a_t) + \alpha_t(s_t, a_t) \times \Big[\overbrace{R(s_{t+1}) + \gamma \max_{a_{t+1}}{Q(s_{t+1}, a_{t+1})}}^{Valor~aprendido} - Q(s_t, a_t)\Big]$$
\end{centering}

Donde $Q(s_t, a_t)$ es el valor de la matriz $Q$ para la acción $a_t$ en el estado $s_t$, $\alpha_t(s_t, a_t)$ es el ritmo de aprendizaje (learning rate) con \mbox{$0 < \alpha \leq 1$} (puede ser el mismo para todos los pares), $R(s_{t+1})$ es la recompensa observada en la interacción actual, $\gamma$ es el factor de descuento (discount factor) con \mbox{$0 \leq \gamma < 1$} y $\max_{a_{t+1}}{Q(s_{t+1}, a_{t+1})}$ es el máximo valor que se espera obtener en la siguiente interacción.

Cada vez que el agente tiene que elegir una acción ($a_t$) en un estado ($s_t$), consulta en su matriz $Q$ cuál es la acción que mayor valor tiene para ese estado y, tras recibir la recompensa correspondiente a la acción realizada, se actualiza su valor ($Q(a_t, s_t)$) con la fórmula antes vista.

\subsubsection{Ritmo de aprendizaje $\alpha$ (Learning rate)}
El ritmo de aprendizaje (con $0 < \alpha \leq 1$) determina hasta qué punto la nueva información adquirida sobrescribirá a la información antigua. Un factor de 0 hará que el agente no aprenda nada (por eso $\alpha$ nunca podrá ser igual a 0), mientras que un factor de 1 hará que el agente considere únicamente la información más reciente.

\subsubsection{Factor de descuento $\gamma$ (Discount factor)}
El factor de descuento (con \mbox{$0 \leq \gamma < 1$}) determina la importancia de las recompensas futuras. Un factor de 0 hará al agente ``oportunista'', considerando solo las recompensas actuales, mientras que con un factor próximo a 1 hará que se esfuerce por obtener recompensas a más largo plazo. Si el factor de descuento se igualara o excediera a 1, los valores de la matriz Q divergerían, por lo que esta situación suele evitarse.

\subsection{Complejidad Kolmogorov}\label{sec:ComplejidadKolmogorov}
En la teoría de información algorítmica, la complejidad Kolmogorov de un objeto es la cantidad de recursos computacionales necesarios para describir ese objeto. A la complejidad Kolmogorov también se la conoce como `complejidad descriptiva', `complejidad Kolmogorov–Chaitin' o `complejidad del tamaño del programa'.

Para definir la complejidad Kolmogorov primero tenemos que describir un lenguaje de descripción para cadenas. Esta descripción puede ser, por ejemplo, la de un lenguaje de programación como lo es C o Java. Si $p$ es un programa (escrito en C o Java, por ejemplo) ejecutado en una máquina de referencia $U$ que obtiene como salida la cadena $x$, entonces $p$ es una descripción para $x$ en la máquina de referencia $U$.

La relevancia de la máquina de referencia $U$ depende sobretodo del tamaño de $x$. Ya que cualquier máquina que sea Turing-completa puede emular a cualquier otra, se mantiene que para todo par de máquinas $U$ y $V$, existe una constante $c(U, V)$ (la longitud del emulador de $U$ en $V$) que solo depende de $U$ y $V$ y no de $x$, tal que para todo $x$, \mbox{$|K_U(x) - K_V(x)| \leq c(U, V)$}. Por lo que la constante $c(U, V)$ será relativamente menos significativa conforme la longitud de $x$ sea más grande.

Una vez que encontramos el programa que obtiene $x$ cuya longitud es mínima (con respecto al resto de programas que obtienen x), entonces la longitud de esta descripción es la denominada complejidad Kolmogorov.

\begin{definition} {\bf Complejidad Kolmogorov}\label{def:kolmogorov}
\[ K_U(x) := \min_{p \textrm{ tal que } U(p)=x} |p| \]
donde $|p|$ denota la longitud en bits de $p$ y $U(p)$ denota el resultado de ejecutar $p$ en $U$.
\end{definition}

Para entenderlo mejor veamos dos ejemplos:

\begin{center}
{\footnotesize \ttfamily azazazazazazazazazazazazazazazazazazazazazazazazazazazazazaz}
{\footnotesize \ttfamily lnkjdfglkfgvoijmnfgvil}
\end{center}

Como podemos ver la primera cadena es una sucesión de ``az'' que se puede describir en lenguaje natural (en castellano) como ``az 30 veces'', utilizando un total de 11 símbolos (contando espacios). Sin embargo, para la segunda cadena (aún siendo una cadena bastante más pequeña que la primera) es bastante difícil encontrar alguna forma de describirla que utilice 11 símbolos o menos. Por lo que la segunda cadena es más compleja de describir que la primera según el lenguaje castellano.

El problema principal de la complejidad Kolmogorov es que no es computable debido al problema de la parada (halting problem, o más generalmente, el \emph{Entscheidungsproblem}). Imaginemos que tenemos un programa $p$ que proporciona $x$ cuya longitud es $|p| = n$, lo que significa que $n$ es una cota superior de la complejidad Kolmogorov de $x$, esto es $n \geq K(x)$. Sin embargo, aun es posible que existieran programas que proporcionen la cadena de salida $x$ utilizando una descripción cuya longitud es menor a $n$ y, sin embargo, tardasen un tiempo prácticamente infinito en ejecutarse. Por este motivo, en la práctica, esta complejidad es incomputable.

Existen variantes a la complejidad Kolmogorov que sí son computables. Una de ellas es la denominada complejidad de Levin $Kt$.

\begin{definition} {\bf Complejidad de Levin $Kt$}
\[ Kt_U(x) := \min_{p \textrm{ tal que } U(p)=x} \{ |p| + \log time(U,p,x) \} \]
donde $|p|$ denota la longitud en bits de $p$, $U(p)$ denota el resultado de ejecutar $p$ en $U$ y $time(U,p,x)$ denota el tiempo\footnote{Aquí el \emph{tiempo} no se refiere a tiempo físico sino a tiempo computacional, p.~ej. pasos computacionales utilizados por la máquina $U$. Esto es importante, ya que la complejidad de un objeto no puede depender de la velocidad de la máquina en donde se ejecuta.} que $U$ tarda en ejecutar $p$ para producir $x$.
\end{definition}

Una vez añadimos el tiempo que se tarda en obtener la cadena $x$, cualquier programa que tarde un tiempo infinito ($y$) en ejecutarse será más complejo que cualquier otro programa conocido que consiga proporcionar $x$ en un tiempo finito, ya que $\lim_{y \to \infty} log(y) = \infty$.

Para más información sobre la complejidad Kolmogorov véase \cite{Li-Vitanyi08}.

\subsection{Distribución universal}\label{sec:DistribucionUniversal}
A partir de la complejidad Kolmogorov podemos definir la probabilidad universal para una máquina $U$ del siguiente modo:
\begin{definition} {\bf Distribución universal}
Dada una máquina prefijada\footnote{Para una definición conveniente de la probabilidad universal, se requiere que $U$ sea una máquina prefijada (véase p.~ej.~\cite{Li-Vitanyi08} para más detalles). Nótese también que incluso para máquinas prefijadas existen infinitas entradas para $U$ que obtienen como salida $x$, por lo que $P_U(x)$ es un estricto límite inferior de la probabilidad de que $U$ obtenga $x$ como salida (dada una entrada aleatoria).} $U$, la probabilidad universal de la cadena $x$ se define como:
\[ P_U(x) := 2^{-K_U(x)} \]
\end{definition}
la cual da mayor probabilidad a los objetos cuya menor descripción sea más pequeña y da menor probabilidad a los objetos cuya menor descripción sea más grande. Cuando $U$ es universal (es decir, Turing-completa), esta distribución es similar (en función de una constante) a la distribución universal para cualquier otra máquina universal distinta al poder emularse entre sí.

\subsection{Algoritmos de Markov}\label{sec:AlgoritmoMarkov}
Un algoritmo de Markov es un sistema de reescritura de cadenas que utiliza reglas de tipo gramaticales para operar con cadenas de símbolos. Los algoritmos de Markov se han demostrado que son Turing-completos, lo que significa que a partir de su simple notación son adecuados como modelo general de computación.

\begin{definition} {\bf Algoritmo de Markov}

Un algoritmo de Markov es una tripleta $\left\langle \Sigma, \Gamma, \Pi \right\rangle$ donde $\Sigma$ es el alfabeto de la cadena de entrada, $\Gamma$ es el alfabeto del algoritmo con $\Sigma \subseteq \Gamma$ y $\Pi$ es una secuencia ordenada de reglas (o también llamadas producciones) de la forma $LHS \rightarrow RHS$ (reglas normales) o $LHS \rightarrow\cdotp RHS$ (reglas terminantes) donde $LHS$ y $RHS$ representan palabras (posiblemente de longitud 0) en $\Gamma^*$.

El algoritmo procesa una cadena de entrada $s \in \Sigma^*$ de la siguiente forma:

\begin{enumerate}
\itemsep=0px
\item\label{itm:AlgoritmoMarkovPrincipio} Comprueba las reglas en orden descendente, comprobando si alguna subcadena de `$s$' (incluyendo la cadena vacía) concuerda con la parte izquierda ($LHS$) de las reglas.
\item Si ninguna subcadena de `$s$' concuerda con ninguna parte izquierda ($LHS$) de las reglas se termina el algoritmo.
\item Si alguna parte izquierda ($LHS$) concuerda con una o más subcadenas en `$s$' reemplazar en esta iteración del algoritmo la subcadena situada más a la izquierda en `$s$' por la parte derecha de la regla ($RHS$).
\item Si la regla usada era una regla terminante, se termina el algoritmo.
\item Si no, volver al paso \ref{itm:AlgoritmoMarkovPrincipio}.
\end{enumerate}
\end{definition}

Vamos a ver un ejemplo de un algoritmo de Markov que transforma un número en binario en su correspondiente número unario (utilizamos `|' para representar cada unidad), donde $\Sigma = \{1, 0\}$, $\Gamma = \Sigma \cup \{|\}$ y con las siguientes reglas $\Pi$:

\begin{enumerate}
\itemsep=0px
\item $|0 \rightarrow 0||$
\item $1 \rightarrow 0|$
\item $0 \rightarrow$
\end{enumerate}

Ahora veamos el resultado de ejecutar el anterior algoritmo de Markov utilizando `101' (5 en binario) como cadena de entrada. Para aclarar la ejecución del algoritmo subrayaremos la subcadena transformada e indicaremos la regla que se ha utilizado en cada paso.

\begin{itemize}
\itemsep=0px
\item $\underline{1}01 \rightarrow_2 \underline{0|}01$
\item $0\underline{|0}1 \rightarrow_1 0\underline{0||}1$
\item $00||\underline{1} \rightarrow_2 00||\underline{0|}$
\item $00|\underline{|0}| \rightarrow_1 00|\underline{0||}|$
\item $00\underline{|0}||| \rightarrow_1 00\underline{0||}|||$
\item $\underline{0}00||||| \rightarrow_3 00|||||$
\item $\underline{0}0||||| \rightarrow_3 0|||||$
\item $\underline{0}||||| \rightarrow_3 |||||$
\end{itemize}

\section{Evaluación de inteligencia biológica}
En esta sección vemos dos de las disciplinas de la psicología que buscan evaluar la inteligencia de las especies biológicas en general y de la especie humana en particular.

\subsection{Psicometría}
Desde hace mucho tiempo, los psicólogos han tratado de medir las capacidades y cualidades de los seres humanos para poder compararlas con las del resto de individuos de su misma especie. Uno de los motivos de querer medirlas, es que algunas personas claramente disponen de mayores capacidades y cualidades que otras, y el ser humano, en su afán de investigación y comprensión de la mente humana, pretende ser capaz de medirlas de una forma eficaz y objetiva.

A partir de este afán de investigación apareció la psicometría (\cite{embretson2000psychometric}). La psicometría es la disciplina que se encarga de la medición en psicología, y su finalidad es construir y utilizar adecuadamente los tests (o pruebas) y las escalas, de tal modo que se garantice su fiabilidad, validez y aplicación adecuada para medir los diferentes aspectos psicológicos; como lo son: el conocimiento, las habilidades sociales, la personalidad, \dots

La psicometría engloba la construcción de tests que sean válidos y fiables\footnote{Un test es `válido' si mide el atributo que realmente pretendía medir, mientras que es `fiable' si lo mide siempre de igual manera, obteniendo los mismos resultados para un sujeto en distintas evaluaciones de una misma prueba.} para medir los diferentes aspectos psicológicos de los seres humanos ofreciendo un valor tras la medición. De entre los distintos tipos de aspectos, el primero que se trató de medir fue la inteligencia y, posteriormente, se construyeron nuevos tests para medir el resto de aspectos cognitivos y de personalidad de los seres humanos.

Para realizar estos tests, se debe conocer (1) qué se pretende medir y (2) las características que mide cada test. Si se utilizase un test para medir la motivación de una persona, entonces los resultados obtenidos no nos servirán para tomar ninguna decisión sobre, por ejemplo, su personalidad, por lo que los resultados de cada test son relativos y únicamente válidos al objeto de estudio. Además, no existe ningún método teórico para construirlos. Los tests se construyen por prueba y error, formándolos a partir de los ejercicios que han ofrecido los resultados esperados tras evaluar a distintos individuos.

Estos tests se deberán ajustar al tipo de individuo que se pretenda evaluar dependiendo de sus características personales (edad, cultura, \dots), por lo que se precisa de un psicólogo experto que diseñe los tests (o que seleccione alguno previamente construido que se ajuste correctamente a los individuos) e interprete los resultados cada vez que se quiera realizar una nueva medición.

Uno de los puntos fuertes de estos tests psicométricos es que sus resultados nos permiten hacer comparaciones de las capacidades de una persona con respecto a otra, o incluso nos permiten ver la evolución de sus capacidades en diferentes momentos de su vida.

También existen algunos tipos de evaluaciones que pretenden medir la inteligencia general de los seres humanos, como por ejemplo ``la prueba de inteligencia de Cattell'', conocida como ``Test de factor <<g>>'' que busca medir la inteligencia concebida como una capacidad mental general, o ``factor g'', y mediante tareas no verbales, eliminar la influencia de habilidades ya cristalizadas como la fluidez verbal y otros aprendizajes adquiridos.

Podemos ver una de las aplicaciones prácticas de estos tests en las grandes empresas. En los últimos años, se han utilizado para encontrar a los aspirantes que mejor se adecuan a sus puestos de trabajo vacantes. En este proceso se suelen realizar distintos tests a los aspirantes para tener un valor de sus capacidades y, a partir de sus resultados, poder decidir de una forma objetiva cuál es el aspirante cuyos conocimientos, capacidades sociales, inteligencia, etc. se corresponden mejor a las necesidades de la empresa.

\subsection{Psicología comparada}
%http://www.apuntesdepsicologia.com/ramas-de-la-psicologia/psicologia-comparada.php
Otra disciplina de la psicología es la denominada `Psicología comparada' \cite{morgan1894introduction} (puede verse una versión más moderna en \cite{morgan2006introduction}). La psicología comparada aborda el estudio del comportamiento animal, al igual que otras disciplinas, como la etología, la psicología animal, y recientemente la sociobiología, todas originadas en la misma base teórica. El origen está en la teoría de la evolución de las especies de Darwin y Wallace \cite{darwin-originofspecies-1859}.

La psicología comparada, conocida también como psicología animal, es una disciplina de la psicología que pretende conocer la conducta animal y los mecanismos que la provocan mediante el estudio de diversas especies y comparando los comportamientos estudiados con los humanos.

La psicología comparada estudia especialmente la conducta de los simios por ser la especie más cercana al hombre, ya que compartimos alrededor del $95\%$ de los genes con los chimpancés y los bonobos, por lo cual están considerados las especies más próximas al ser humano. Pero también se estudia la conducta de otros animales, por ser más fáciles de criar, y que han aportado datos interesantes a los psicólogos como, p.~ej. los ratones de laboratorio.

El tipo de investigación de la psicología comparada puede estudiar conductas condicionadas (como las experiencias de Pavlov), pero también se ocupa de la anatomía y las estructuras cerebrales.

La psicología comparada promueve el estudio experimental de la psicología del aprendizaje y condicionamiento animal y humano. Investigando los procesos básicos del aprendizaje y condicionamiento clásico y el instrumental, incluyendo las relaciones de los procesos de aprendizaje, memoria, atención, motivación, cognición, comparadas en organismos animales y humanos, contemplando también las bases neurobiológicas.

Existe un problema al tratar de realizar este tipo de tests en psicología comparada. Este es un problema de comunicación, ya que no es posible decirles verbalmente que es lo que se espera de ellos durante su evaluación. Es por ello que se les suele ofrecer algún tipo de recompensa (como por ejemplo comida) cada vez que consigan resolver algún tipo de tarea y así condicionarles para que realicen el test.

\section{Evaluación de inteligencia de máquinas}
Debido a la aparición de la inteligencia artificial, se han propuesto varios tipos de evaluaciones para tratar de comprobar (en algunos casos) cuán inteligentes son estas máquinas, y (en otros) si su inteligencia es comparable a la de los humanos.

Veamos algunas de estas propuestas.

\subsection{Test de Turing}
Una de las primeras pruebas que se formularon fue propuesta por Alan M. Turing en 1950 en su `juego de la imitación' \cite{turing1950} (comúnmente conocido como el `test de Turing'). Turing cambió la pregunta natural de si una máquina piensa por si una máquina puede ser indistinguible de un sistema que piense. Para ello el juego supone 3 participantes: (1) un ser humano, (2) una máquina que deberá hacerse pasar por humano y (3) un juez que deberá averiguar cual de los dos anteriores es la máquina. Debido a que en este juego no se desea evaluar si una máquina es físicamente parecida a un humano, sino si es mentalmente capaz de parecerse a un humano, el juego debe realizarse sin existir un contacto directo entre el juez y los otros dos participantes, de modo que no deba influir la apariencia física con las decisiones que tome el juez.

El test propuesto por Turing ha creado mucha controversia en los últimos años \cite{oppydowe2008}. Se ha llegado a varias conclusiones a partir de este test: (1) el test requiere de un juez (humano) para realizarse, por lo que no puede hacerse automáticamente, (2) el test no proporciona un valor medible de la inteligencia, simplemente se contesta si parece o no humano (no es gradual) y (3) realmente el test no evalúa si la máquina es o no inteligente, sino que mide su `humanidad'.

Ya que este test no mide la inteligencia de las máquinas sino su humanidad, no resulta adecuado para este propósito. Sin embargo, se han creado muchas variantes del test, como el Total Turing Test (TTT) \cite{Harnad89} o el Inverted Turing Test (ITT) \cite{Watt96naivepsychology}. Incluso un multimillonario (Loebner) ha creado un concurso con este test, dotando de una sustancial cantidad económica (100.000 $\$$) para los creadores del programa que consiga engañar a varios de sus jueces instruidos para descubrir al impostor. En la actualidad, todavía ningún programa ha conseguido obtener este premio.

\subsection{CAPTCHAs}
\begin{figure}[h!]
\centering
\includegraphics[width=0.75\textwidth]{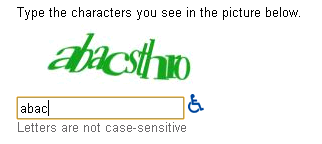}
\caption{Texto distorsionado generado por un CAPTCHA}
\label{fig:captcha}
\end{figure}

Una de las variantes al test de Turing más conocida es el CAPTCHA (Completely Automated Public Turing test to Tell Computers and Humans Apart) \cite{von2004telling}\cite{von2008recaptcha}. Básicamente, estos test son capaces de distinguir entre humanos y máquinas generando textos distorsionados. Estos textos resultan fácilmente comprensibles por los humanos, mientras que, sin embargo, resultan ser extremadamente complejos de reconocer para las máquinas en el estado del arte actual. Podemos ver un ejemplo de estos tests en la Figura~\ref{fig:captcha}.

Desde su aparición en 2004, estos CAPTCHAs se han utilizado en formularios por internet con múltiples finalidades: evitar que las máquinas sean capaces de descargar archivos de forma masiva, crear múltiples cuentas de correo, postear comentarios en foros, etc. Sin embargo, existen algunos programas que se están especializando en entender estos CAPTCHAs, por lo que, para seguir siendo útiles, deben actualizarlos continuamente haciéndolos cada vez más difíciles de resolver. Actualmente estos CAPTCHAs son capaces de distinguir entre humanos y máquinas, pero dentro de unos años (debido a esta constante actualización) los problemas generados por los CAPTCHAs serán tan complejos que ni siquiera los humanos serán capaces de resolverlos, por lo que se deberán buscar nuevos métodos que nos permitan diferenciar mejor a los humanos de las máquinas a través de internet.

\subsection{Machine Intelligence Quotient (MIQ)}
Hoy en día, los aparatos electrónicos resultan cada vez más `inteligentes'. Podemos ver lavadoras que intentan consumir el mínimo posible de agua en función de la carga que tienen, neveras que conservan mejor los alimentos dependiendo del tipo de alimento que se encuentre en cada zona, etc. Estas son máquinas que tratan de realizar su cometido de una forma `óptima' e `inteligente' y, tras su aparición, L. A. Zadeh acuñó el término de `Machine Intelligence Quotient' (MIQ) para comparar estás máquinas con capacidades `inteligentes' que comenzaron a aparecer a partir de 1990; las cuales manifestaban una capacidad `impresionante' para razonar, tomar decisiones inteligentes y aprender de la experiencia (\cite{zadeh1994fuzzy}).

Sin embargo, este término realmente fue sugerido por Zadeh para referirse a estas máquinas cada vez más `inteligentes'. En realidad, su objetivo real era dar a conocer los avances y posibilidades de la lógica difusa para controlar estos aparatos.

Aunque este término parece corresponderse con un tipo de cociente de inteligencia en las máquinas, en realidad no se aplica a ningún tipo de evaluación ni posee ningún concepto matemático que lo defina. Solo existe como un concepto intuitivo de las capacidades de las máquinas actuales en comparación con las de épocas anteriores.

No obstante, podemos ver algunos intentos para definir formalmente este MIQ. Por ejemplo, en uno de estos intentos se mide la inteligencia de las máquinas a través de 3 perspectivas distintas: técnica, organizacional y personal, para posteriormente calcular el MIQ (para más detalles véase \cite{ulinwa2008machine}). En otros proponen estimar el MIQ a través de un método más orientado a partir de los humanos. En \cite{journals/tsmc/ParkKL01} se sugiere que, como algunas máquinas de control no pueden realizar todo el trabajo por ellas mismas, sino que requieren de alguna intervención humana, la MIQ debería representar la inteligencia de las máquinas desde el punto de vista de los humanos. Como se puede ver, ni siquiera entre estos intentos parece que se pueda obtener una estimación clara y consensuada de este término.

\subsection{Test de compresión}
David L. Dowe propuso en 1997 añadir un nuevo requisito computacional que consideraba necesario para la definición de inteligencia \cite{DoweHajek97a}. Este nuevo requisito (que habría que añadir al Test de Turing) no estaba directamente relacionado sobre el comportamiento de las máquinas, sino más bien en su capacidad inductiva de aprendizaje. Básicamente Dowe sugirió que para que una máquina fuera inteligente también debía ser capaz de aprender de forma inductiva la representación del dominio del test en el que se le está evaluando. De este modo, si dos programas ofrecen los mismos resultados, entonces el programa más comprimido sería el más inteligente.

Posteriormente Matthew V. Mahoney propuso un idea similar en 1999 \cite{conf/aaai/Mahoney99}. En este artículo se ofrecía también una solución a un problema del test de Turing, el cual consistía en que únicamente era capaz de obtener un valor binario de si la máquina parece o no una persona y no un valor numérico que nos indique cuán humanas parecen. Estos test se basan en las distintas capacidades de compresión de textos que poseen las máquinas y los humanos. Los humanos son capaces de comprimir mejor debido a su mayor conocimiento general y estructura lógica del lenguaje, por lo que (en estos test) la máquina cuya capacidad de compresión se acerque más a la capacidad de compresión humana, será la que más humana parezca.

Este tipo de compresión ha sido añadido como un reto en el ``Hutter Prize''\footnote{http://prize.hutter1.net/}. Este reto consiste en comprimir 100 MBytes de información de la Wikipedia, de modo que la compresión que mejore la mejor marca obtendrá el premio.

\subsection{C-test}
José Hernández Orallo propuso en 1998 el C-test (\cite{HernandezOrallo-MinayaCollado98}\cite{HernandezOrallo00b}\cite{journals/aicom/Hernandez-Orallo00}), el cual trata de evaluar la inteligencia universal basándose en nociones computacionales bien conocidas, como lo son: la teoría de la inducción de Solomonoff para predecir el siguiente carácter y la complejidad Kolmogorov (véase \ref{sec:ComplejidadKolmogorov}) para concluir la dificultad de cada ejercicio. Por tanto, los C-tests miden la capacidad de inferencia inductiva de los sujetos.

En esta aproximación, los ejercicios que se proponen en los tests están formados por una secuencia de caracteres, y el objetivo es averiguar el siguiente carácter que mejor sigue la secuencia.

En la Figura~\ref{fig:C-test} vemos un ejemplo de estos tests.

\begin{figure}[!h]
\centering
{\sffamily\small
$$
\begin{array}{llll}
k=9  & : & $a, d, g, j, \dots$ & $Respuesta: m$ \\
k=12 & : & $a, a, z, c, y, e, x, \dots$ & $Respuesta: g$ \\
k=14 & : & $c, a, b, d, b, c, c, e, c, d, \dots$ & $Respuesta: d$  \\
\end{array}
$$
}
\caption{Ejercicios de secuencia de caracteres con distintas dificultades ($k$) en el C-test.}
\label{fig:C-test}
\end{figure}

Posteriormente se propuso una factorización para medir otras características (capacidades deductivas, analógicas, \dots) \cite{HernandezOrallo00b}.

\subsection{Juegos competitivos}
En 2002, H. Masum et al. propusieron en \cite{conf/gecco/MasumCO02} el método `Turing Ratio' que hace más énfasis en tareas y juegos en lugar de en tests cognitivos. Parten como premisa que ``\dots hacerlo bien en un amplio rango de tareas es una definición empírica de `inteligencia'{''}. Para ello, llegan a la conclusión de que es necesario encontrar un conjunto de tareas que sean capaces de medir habilidades importantes, que admitan distintas estrategias para poder resolverlas (siendo algunas estrategias cualitativamente mejores que otras) y que sean reproducibles y relevantes con el paso del tiempo. Sugieren un sistema de medida del rendimiento a través de comparaciones por parejas entre sistemas de IA, de una forma similar al utilizado para clasificar a los jugadores en el sistema de clasificación internacional de ajedrez.

Con este planteamiento se hace evidente el hecho de que encontrar estas tareas no resulta fácil, ya que no solo es necesario obtener un conjunto significativo de tareas, sino que también es necesario poder medir su relevancia, su importancia y cuan amplia es cada tarea con respecto a la habilidad que se está evaluando.

\subsection{Los test de IQ clásicos no sirven para evaluar la IA}
En 2003, Sanghi y Dowe se propusieron comprobar si los tests de IQ clásicos eran o no apropiados para medir la inteligencia de algoritmos de IA \cite{sanghidowe2003computer}. Para ello, implementaron algunos algoritmos simples con la finalidad de que hicieran estos test de IQ.

Estos algoritmos obtuvieron unos resultados parecidos a los que obtendría un ser humano. Por lo que se demostró algo que ya se sospechaba, que los tests de IQ clásicos no sirven para medir la inteligencia de cualquier sistema inteligente a parte del ser humano. Esta fue la primera demostración empírica de que los tests psicométricos no son válidos para evaluar máquinas.

\subsection{Inteligencia universal}
Algunos trabajos de Legg y Hutter (p.~ej. \cite{legg2005universal}\cite{LeggHutter07}) han dado una nueva definición de la inteligencia de las máquinas, denominada ``Inteligencia Universal'', la cual está basada, al igual que algunas propuestas anteriores, en la complejidad Kolmogorov y la teoría de predicción (o ``inferencia inductiva'') de Solomonoff. La idea principal es que la inteligencia de un agente se evalúa como algún tipo de suma (o promedio ponderado) de rendimientos en todos los posibles entornos.

Una de las contribuciones más relevantes de \cite{LeggHutter07} es que su definición de \emph{inteligencia universal} permite evaluar formalmente el rendimiento teórico de algunos agentes: un agente aleatorio, un agente especializado, \dots, hasta incluso un agente super inteligente.

Su definición se basa en la idea de que la inteligencia se puede ver como el rendimiento en una variedad de entornos, y sugieren que para esta variedad de entornos se utilicen todos los entornos posibles. Después, dado un agente $\pi$, su inteligencia universal $\Upsilon$ puede obtenerse con la siguiente definición:

\begin{definition} {\bf Inteligencia Universal \cite{LeggHutter07}}\label{def:InteligenciaUniversal}
\[ \Upsilon(\pi, U) := \sum^\infty_{\mu=i}p_U(\mu)\cdot V^\pi_\mu = \sum^\infty_{\mu=i}p_U(\mu)\cdot E\left(\sum_{i=1}^\infty r^{\mu,\pi}_i\right) \]
donde $\mu$ es cualquier entorno codificado en una máquina universal $U$, con $\pi$ como el agente a evaluarse. Con $p_U(\mu)$, asignamos la probabilidad de cada entorno, a pesar de que estas probabilidades no sumarán 1.
\end{definition}

Al tener un número infinito de entornos, no se puede asignar una distribución uniforme a los entornos. La solución es utilizar una distribución universal en una máquina dada $U$, como puede verse en la Sección~\ref{sec:DistribucionUniversal}, adaptando apropiadamente la codificación de los entornos como cadenas en $U$, y asumiendo que la clase de entornos es recursivamente enumerable (véase página 63 de \cite{Legg08} para más detalles).

Y, finalmente, ya que tenemos una suma de un número infinito de entornos, para evitar que las recompensas acumuladas $V^\pi_\mu$ sean infinitas se impone una restricción a todos los entornos:

\begin{definition} {\bf Entornos con recompensas limitadas}
Un entorno $\mu$ es limitado por las recompensas si y solo si para todo $\pi$:
\[ V^\pi_\mu = E\left(\sum_{i=1}^\infty r^{\mu,\pi}_i\right) \leq 1 \]
\end{definition}

El valor numérico dado para cada recompensa $r_i$ se utiliza para computar la recompensa esperada global $V^\pi_\mu$.

\chapter{Marco general del sistema de evaluación}\label{cap:MarcoGeneral}
En este capítulo vemos el marco general/conceptual en el que se basa el proyecto ANYNT y el diseño sugerido para construir un sistema de evaluación a partir de este marco.\footnote{Parte del material de este capítulo ha sido extraído a partir del artículo \cite{HernandezOralloDowe2010}.}

La evaluación y medición de la inteligencia están estrechamente relacionadas con la noción de test, lo que, como hemos visto, es el día a día de la psicometría. En el capítulo anterior hemos visto multitud de sistemas de evaluación que pretenden medir la inteligencia. Mientras algunos de estos tests intentan medir ciertas capacidades cognitivas concretas, otros han intentado realizar una evaluación más general de estas capacidades (podemos ver, por ejemplo, los tests de factor <<g>> en psicometría como un intento de evaluar estas capacidades generales). Sin embargo, varios de estos sistemas de evaluación resultan incompletos para evaluar la inteligencia general, como por ejemplo: la falta de un resultado gradual (test de Turing), la interpretación sólo empírica de los resultados obtenidos (psicometría o workshops on Performance Metrics), etc.

Desde un punto de vista científico, se puede pensar sobre cuál es la desiderata que se espera de un procedimiento de medición de inteligencia, y, a partir de ahí, ver si esta desiderata es físicamente alcanzable. Una visión científica ambiciosa de las propiedades que debería tener una medición de la inteligencia sería:

\begin{itemize}
\item Debe ser ``universal'' en el sentido de que no debería ser diseñada solo para medir a un tipo particular de sistema inteligente. Debe permitirnos medir cualquier tipo de sistema inteligente (biológico o computacional).
\item En lugar de utilizar métodos empíricos para realizar las evaluaciones, debe derivarse de nociones computacionales bien fundadas con formalizaciones precisas de los conceptos involucrados en la medición, tales como: entorno, acción, objetivo, puntuación, tiempo, etc.
\item Uno de los problemas de muchos tests actuales es que solo miden algunas habilidades concretas de la inteligencia, como el razonamiento o la memoria. El test debe ser significativo en el sentido de que lo que se esté midiendo represente la noción más general de inteligencia.
\item Debe ser capaz de evaluar y puntuar cualquier sistema inteligente existente o cualquier sistema que pueda ser construido en el futuro, permitiendo así la comparación entre distintas generaciones de sistemas inteligentes incluso por encima de la inteligencia humana.
\item La medición debería servir para medir cualquier nivel de inteligencia y cualquier escala de tiempo del sistema. Debe ser capaz de evaluar sistemas tanto brillantes como ineptos (cualquier nivel de inteligencia) así como sistemas muy lentos o muy rápidos (cualquier escala de tiempo).
\item La calidad (precisión) de la evaluación debería ser ajustable y debería depender sobretodo del tiempo que se le proporcione a la evaluación. La evaluación puede ser interrumpida en cualquier momento, produciendo una aproximación a la puntuación de la inteligencia. De este modo cuanto más tiempo se deje para la evaluación, más aproximado será el resultado que se obtenga (test \emph{anytime}).
\end{itemize}

El sistema de evaluación propuesto en el proyecto ANYNT para evaluar la inteligencia general, proporciona un test que pretende reunir todas las propiedades anteriormente descritas: puede ser válido tanto para sistemas artificiales como biológicos, de cualquier grado de inteligencia y cualquier velocidad, no es antropomórfico (no es exclusivo para humanos), es gradual (proporciona un valor estimado de la inteligencia del sujeto), es \emph{anytime}, está basado exclusivamente en nociones computacionales (como la complejidad Kolmogorov, véase la Sección~\ref{sec:ComplejidadKolmogorov}) y también es significativo, ya que promedia la capacidad de obtener buenos resultados en distintos tipos de entornos.

Para conseguir esto, el proyecto ANYNT se ha basado en algunos de los trabajos presentados en el capítulo anterior (e.g., tests de compresión, c-tests e inteligencia universal), además de otros trabajos previos en la medición o definición de la inteligencia \cite{DoweHajek97a}\cite{DoweHajek97b}\cite{DoweHajek98}\cite{HernandezOrallo-MinayaCollado98}\cite{HernandezOrallo00a}\cite{legg2005universal}\cite{LeggHutter07}. Estos trabajos se basan en las nociones de inferencia inductiva, predicción, compresión y aleatoriedad y están todos propiamente formalizados y comprendidos en el contexto de la Teoría Algorítmica de la Información (AIT), complejidad Kolmogorov, la navaja de Occam o el principio de la Longitud Mínima del Mensaje (MML) \cite{Li-Vitanyi08}\cite{Wallace-Boulton68}\cite{Wallace-Dowe99}\cite{Wallace05}.

El punto de partida es el mismo que en \cite{LeggHutter07}; se establece un marco de trabajo en donde un agente interactúa con un conjunto de entornos generados a partir de una distribución, siendo estos entornos los ejercicios que se realizarán durante el test. Esta es una idea bastante simple y natural, ya que la inteligencia se puede ver como el rendimiento en una variedad de entornos. También parece atractivo si no nos preocupamos sobre el número de entornos ni sobre la cantidad de tiempo que se requiere en cada uno de ellos. Sin embargo, si se quiere que la evaluación sea fiable con una pequeña muestra de entornos y un pequeño número de interacciones en cada uno de ellos, entonces se deben controlar algunos problemas, por lo que es necesario establecer algunas condiciones sobre los entornos: (1) se necesita definir una medida de complejidad para entornos apropiada y computable; la complejidad del entorno permitirá distinguir su dificultad y situar correctamente los resultados, (2) deben ser discriminativos; distintos tipos de sistemas inteligentes deberán obtener distintos resultados en concordancia con su inteligencia real, (3) deben reaccionar inmediatamente y deben estar balanceados (en términos de la puntuación esperada) para poder asegurar que sus resultados son apropiados; es necesario disponer de un valor de referencia en todos los entornos con el que poder comparar los resultados de los agentes y (4) se necesita diseñar una manera de derivar una muestra de estos entornos que se ajuste al nivel de inteligencia del sujeto; es necesario diseñar un mecanismo que nos permita alcanzar inequívocamente el nivel de inteligencia apropiado para cada sujeto y, por lo tanto, obtenga resultados más estables conforme más nos ajustemos a su nivel de inteligencia.

\section{Dificultad de los entornos}
Una importante cuestión es cómo elegir los entornos en donde se evaluará a los agentes. Una primera aproximación sería utilizar una distribución uniforme para obtener un entorno de entre todos los posibles entornos existentes. Sin embargo, al existir infinitos entornos en donde poder evaluar a los agentes, la probabilidad de obtener uno de todos estos entornos sería \mbox{$1/\infty = 0$}.

Otra aproximación para elegir los entornos sería utilizar una distribución universal (véase la Sección~\ref{sec:DistribucionUniversal})\footnote{Para el cálculo de la complejidad puede utilizarse una variante computable de la complejidad Kolmogorov como, por ejemplo, el $Kt$ de Levin (véase la Sección~\ref{sec:ComplejidadKolmogorov} para más detalles).}, evitándose así cualquier sesgo particular hacia un individuo específico, especie o cultura, haciendo el test universal para cualquier posible tipo de sujeto. Con esta distribución, los entornos más simples se obtendrían con mayor probabilidad, mientras que los entornos más complejos tendrían menor probabilidad. Sin embargo, simplemente utilizando esta distribución se obtendrían una y otra vez los entornos simples (nótese que un entorno $\mu$ con $K(\mu) = 1$ aparecería la mitad de las veces, ya que $2^{-1}=$ $ $0,5), esta forma de obtener los entornos aun limitaría o proporcionaría bajas probabilidades a entornos muy complejos. Desde luego, deberán descartarse los entornos repetidos para no volver a ser tenidos en cuenta durante la evaluación.

Antes de seguir, hay que clarificar las nociones de simple/fácil y complejo/difícil que vamos a utilizar. Por ejemplo, solo eligiendo un entorno con un $K$ alto no nos asegura que el entorno sea en realidad complejo.

\begin{example}
Considérese, por ejemplo, que un entorno relativamente simple $\mu_1$ con un $K$ alto, tiene un comportamiento sin ningún patrón hasta el ciclo $i$ = 1.000 y entonces se repite este comportamiento indefinidamente. $K$ será alto, ya que se deben codificar 1.000 ciclos, pero el patrón es relativamente simple (una repetición) si el agente tiene una gran memoria e interactúa durante miles de ciclos.

Ahora considérese un segundo entorno $\mu_2$ con un $K$ alto que funciona como sigue: para cualquier acción, proporcionar la misma observación $o$ y recompensa $r$, excepto cuando la interacción $i$ es potencia de 2. En este caso (y solo en este caso), la observación y la recompensa dependerán de una fórmula compleja en función de las anteriores acciones. Es fácil ver que en un amplio número de ciclos el comportamiento del entorno es simple, mientras que solo en unos pocos ciclos el entorno requerirá de un comportamiento complejo.

Finalmente considérese, p.~ej. un entorno $\mu_3$ con solo dos posibles acciones tal que toda secuencia de acciones lleva a ``subentornos'' muy simples, excepto una secuencia específica de acciones $a_1a_2\dots{}a_n$ que conduce a un subentorno complejo. Lógicamente, la complejidad de este entorno es alto, pero solo 1 acción de $2^n$ combinaciones hará accesible y visible para el agente el subentorno complejo. Por lo tanto, con una probabilidad de $(2^n -1)/2^n$, el agente verá este entorno $\mu_3$ como muy simple.
\end{example}

Las cuestiones aparecidas con los dos primeros entornos $\mu_1$ y $\mu_2$ también pueden aparecer con cadenas, pero el tercer entorno $\mu_3$ nos muestra que la noción de fácil o difícil no es la misma para cadenas que para entornos. En el caso de un entorno, un agente solo llegará a explorar una parte (generalmente) pequeña.

\begin{figure}[h!]
\centering
Entorno con $K$ alto $\Longleftarrow$ Entorno intuitivamente complejo (difícil)\\
Entorno con $K$ bajo $\Longrightarrow$ Entorno intuitivamente sencillo (fácil)
\caption{Relación entre $K$ y la complejidad intuitiva.}
\label{fig:RelacionComplejidadDificultad}
\end{figure}

Como muestra la Figura~\ref{fig:RelacionComplejidadDificultad}, la relación es unidireccional: dada una $K$ baja, podemos afirmar que el entorno parecerá simple. No obstante, dado un entorno intuitivamente complejo, la $K$ deberá ser necesariamente alta.

Dada la relación vista en la Figura~\ref{fig:RelacionComplejidadDificultad}, solo a través de un $K$ alto encontraremos entornos complejos, pero no todos ellos serán difíciles. Desde la perspectiva del agente, sin embargo, esta situación se hará más acusada, ya que muchos entornos con una $K$ alta contendrán patrones difíciles a los que será difícil que el agente pueda acceder a través de sus interacciones. Como resultado, los entornos serán probabilísticamente simples. Esto significa que los entornos normalmente son vistos más simples de lo que realmente son, ya que muchos de los patrones que incluyen pueden ser (probabilísticamente) inaccesibles. Por lo tanto, solo se mostrarán patrones muy simples y, a partir de ahí, la mayor parte de la inteligencia medida vendrá a partir de entornos con un $K$ bajo.

\section{Selección de entornos}
Muchos entornos (tanto simples como complejos) serán completamente inútiles para evaluar la inteligencia, e.g., entornos que paran de interactuar, entornos con recompensas constantes o entornos que son muy similares a otros entornos previamente utilizados. Incluir estos entornos en la muestra de entornos resulta un gasto de recursos, ya que deberán ser siempre descartados: habrá que obtener una muestra más correcta para un proceso de test más eficiente. La cuestión es determinar un criterio no arbitrario para excluir algunos entornos. Por ejemplo, la definición de la inteligencia universal propuesta por Legg y Hutter (véase la Definición~\ref{def:InteligenciaUniversal} de la Página~\pageref{def:InteligenciaUniversal}) fuerza a que los entornos interactúen infinitamente, y ya que la descripción debe ser finita, debe existir un patrón. Este patrón puede ser (o no) aprendido por el examinado. Sin embargo, esto obviamente incluye entornos de modo que ``siempre proporcionen la misma observación y recompensa'' y también los entornos que ``proporcionan observaciones y recompensas aleatoriamente\footnote{En la práctica, esto debería ser pseudo-aleatorio, lo cual implica que en realidad existe un patrón.}'', los cuales suponen un problema. No obstante, estos últimos tienen una gran complejidad si asumimos entornos deterministas. En ambos casos, el comportamiento para cualquier agente en estos entornos sería prácticamente el mismo, por lo que, no poseerán ninguna capacidad discriminativa y, por lo tanto, serán inútiles para discriminar entre agentes.

En un entorno interactivo, un requisito obvio para que un entorno sea discriminativo es que lo que haga el agente debe tener consecuencias en sus recompensas. En \cite{Legg08} se ha desarrollado una taxonomía de entornos, y se presenta el concepto de entornos MDP (Markov Decision Process) ergódicos. Los entornos MDP ergódicos se caracterizan por hacer que cualquier posible observación sea alcanzable (en uno o más ciclos) a partir de cualquier estado. Esto significa que en cualquier estado las acciones de los agentes pueden recuperarle de una mala decisión. Estos entornos MDP ergódicos son una gran restricción, ya que muchos entornos reales no proporcionan ``segundas oportunidades''. Si siempre fueran posibles las ``segundas oportunidades'', el comportamiento de los agentes tendería a ser más precipitado y menos reflexivo. Además, parece más fácil aprender y tener éxito en esta clase de entornos que en una clase más general.

En su lugar, se restringe que los entornos sean sensibles a las acciones de los agentes. Esto significa que una mala acción pueda hacer que el agente acabe en una parte del entorno de donde nunca pueda volver (no ergódico), pero que al menos las acciones que haga el agente puedan modificar las recompensas que obtenga en este subentorno. Más concretamente, se quiere que los agentes sean capaces de influenciar en sus recompensas en cualquier punto de cualquier subentorno. Esto significa que no se puede alcanzar un punto a partir del cual las recompensas se proporcionen independientemente de lo que haga el agente. Esto puede formalizarse del siguiente modo:

\begin{definition} {\bf Sensible a las recompensas}

Dado un entorno determinista\footnote{La restricción de entornos deterministas es porque, de lo contrario, la definición debería definirse en términos de la suma de las recompensas esperadas. Esto es perfectamente posible, pero por motivos de simplicidad se asumen entornos deterministas.} $\mu$, decimos que es $n$-acciones sensible a las recompensas si, para toda secuencia de acciones $a_1a_2\dots{}a_k$ de longitud $k$, existe un entero positivo $m \leq n$ de modo que existan dos secuencias de acciones $b_1b_2\dots{}b_m$ y $c_1c_2\dots{}c_m$ cuya suma de las recompensas que se obtiene por la secuencia de acciones $a_1a_2\dots{}a_kb_1b_2\dots{}b_m$ es distinta\footnote{Se pueden establecer varios niveles de sensibilidad mediante el establecimiento de un umbral mínimo en esta diferencia.} a la suma de las recompensas de la secuencia $a_1a_2\dots{}a_kc_1c_2\dots{}c_m$.
\label{def:SensibleRecompensas}
\end{definition}

Nótese que la Definición~\ref{def:SensibleRecompensas} no significa que cualquier acción tenga un impacto en las recompensas (directa o indirectamente), sino que en cualquier punto/instante siempre existen como mínimo dos secuencias de acciones distintas (en $n$ interacciones) que el agente puede elegir para obtener recompensas distintas. Esto significa que el agente puede estar atascado (obteniendo malas recompensas) en un entorno durante algún tiempo si no realiza las acciones correctas, pero al menos siempre puede elegir obtener mejores recompensas en estas situaciones. En otras palabras, en cualquier momento el agente puede luchar para incrementar las recompensas que obtenga (o al menos evitar que decrezcan).

\section{Recompensas}
Cuando hay que evaluar a un agente, es necesario determinar si se necesita alguna restricción en la distribución de las recompensas y, más especialmente, como se agregan las recompensas. La forma más común de evaluar el rendimiento de un agente $\pi$ en un entorno $\mu$ es calcular el valor esperado de la suma de todas las recompensas, p.~ej.:

\begin{definition} {\bf Recompensa acumulada esperada}
\[  V^{\pi}_\mu = E\left( \sum_{i=1}^\infty r_i^{\mu,\pi} \right) \]
\end{definition}

Esta no es la única opción, ya que también se podría determinar si se le da más relevancia a las recompensas inmediatas o más a largo plazo.

Existe una relación interesante entre la sensibilidad de las recompensas y el tipo de recompensas permitidas en los entornos. En \cite{LeggHutter07}, se discute sobre cómo distribuir las recompensas a través del entorno, y se llega a una sola restricción: la recompensa total debe ser menor o igual que 1. Principalmente la razón es que, de lo contrario, todos los agentes podrían acumular en el límite un valor infinito, y todos los agentes podrían obtener la misma puntuación. Nótese que con esta definición original, un entorno que proporcione una recompensa de 1 tras la primera acción y que, a partir de entonces, siempre proporcionase 0 cumpliría con esta restricción, pero resultaría un entorno inútil. La condición de ser sensible a las recompensas hace que esto no sea posible ya que, en cualquier momento, parte del total de la recompensa debe surgir a partir de cualquier subentorno. Esto no excluye, en principio, un entorno que proporciona recompensas de 0 durante miles de acciones iniciales y a partir de ese momento comience a proporcionar recompensas. Esta es la razón por la cual típicamente se especifica un número bajo de $n$ en la definición de entorno con $n$-acciones sensible a las recompensas, la cual implica que debe poderse `consumir' alguna recompensa tras cada $n$ o menos acciones. Un caso extremo es cuando $n = 1$, donde se debe proporcionar alguna cantidad de estas recompensas para cualquier posible acción en cualquier momento.

Obviamente, en la práctica un agente no puede interactuar infinitamente en los entornos. Limitar el número de interacciones es fácil de hacer. Simplemente es necesario establecer un límite de interacciones $n_i$ para cada entorno para la obtención de las recompensas\footnote{Aparte de tener un número finito de interacciones, también se asume que el entorno sea determinista, por lo que esta definición puede ser calculada.}:

\begin{definition} {\bf Recompensas acumulativas con un número finito de interacciones}

\[ V^\pi_\mu(n_i) = \sum_{k=1}^{n_i} r_k^{\mu,\pi} \]
donde $\pi$ es el agente que interactúa en el entorno $\mu$ obteniendo en la interacción $k$ una recompensa de $r_k^{\mu,\pi}$, durante $n_i$ interacciones.
\end{definition}

Sería recomendable hacer que el entorno fuera $n_r$-acciones sensible a las recompensas, con $n_r \leq n_i$ de modo que sus acciones tengan un efecto en las recompensas en el periodo delimitado de interacciones.

Con el objetivo de garantizar una serie de propiedades de convergencia y de balance frente a agentes aleatorios, se imponen una serie de restricciones sobre las recompensas y los entornos.

La primera idea es utilizar recompensas simétricas, las cuales pueden estar en un rango entre -1 y 1, p. ej.:

\begin{definition} {\bf Recompensas simétricas}

\[ \forall i: -1 \leq r_i \leq 1 \]
\end{definition}

Nótese que esto no impide que la recompensa acumulada en un cierto punto sea mayor que 1 o menor que -1. Por lo que, si hacemos muchas acciones, podemos tener una recompensa acumulada mayor que 1. Observando implementaciones físicas, las recompensas negativas no tienen por qué estar asociadas con castigos, lo cual se considera poco ético para individuos biológicos. Por ejemplo, si estamos evaluando un simio, las recompensas desde -1 a -1/3 podrían implicar no dar nada, desde -1/3 a 1/3 dar una pieza de fruta, y desde 1/3 a 1 dos piezas. O una recompensa negativa puede implicar eliminar una fruta adjudicada previamente.

Si nos fijamos en las recompensas simétricas, también es de esperar que los entornos sean simétricos, o más precisamente, que sean balanceados en como proporcionan las recompensas. Esto puede verse del siguiente modo: en un test fiable, queremos que muchos (si no todos) los entornos ofrezcan una recompensa de 0 para agentes aleatorios. La siguiente definición lo formaliza.

\begin{definition} {\bf Entorno balanceado}

Un entorno $\mu$ está balanceado si y solo si:

\begin{itemize}
\itemsep=0px
\item Sus recompensas son: $\forall i : -1 \leq r_i \leq 1$.
\item Dado un agente aleatorio $\pi_r$, se mantiene la siguiente igualdad\footnote{Nótese que el valor esperado debe ser en el límite $0$. Incluso para entornos deterministas, la serie de recompensas acumuladas podría ser divergente, p.~ej., el valor $\lim_{n_i \rightarrow \infty} V^{\pi_r}_\mu(n_i)$ podría no existir y se asume un límite Cesàro ($\lim_{n_i \rightarrow \infty} \frac{1}{n_i} \sum_{j=1}^{n_i} V^{\pi_r}_\mu(n_i)$).}:
\end{itemize}

\[  V^{\pi_r}_\mu = E\left( \sum_{i=1}^\infty r_i^{\mu,\pi_r} \right) = 0 \]

Donde $r_i^{\mu,\pi_r}$ se refiere a la recompensa en la interacción $i$ que ha obtenido $\pi_r$ en el entorno $\mu$, $E$ es el valor esperado y $V^{\pi_r}_\mu$ se refiere a la media de recompensas obtenida por $\pi_r$.
\label{def:EntornoBalanceado}
\end{definition}

Esto excluye tanto a entornos hostiles como a entornos benévolos, p.~ej. entornos donde realizando acciones aleatorias se conseguiría más recompensas negativas (respectivamente positivas) que positivas (respectivamente negativas). En muchos casos no es difícil probar que un entorno particular está balanceado. Para entornos complejos, la restricción previa se puede comprobar experimentalmente. Otra aproximación es proporcionar una máquina de referencia que únicamente genere entornos balanceados.

Nótese que las modificaciones previas en las recompensas ahora nos permiten usar una media en lugar de una recompensa acumulada para calcular el resultado de los entornos, es decir:

\begin{definition} {\bf Recompensa media}

Dado un entorno $\mu$, con $n_i$ como el número de interacciones completadas, entonces la recompensa media para el agente $\pi$ se define como:

\[  v^{\pi}_\mu(n_i) = \frac{V^{\pi}_\mu(n_i)}{n_i} \]
\end{definition}

\section{Interacciones prácticas}
Si queremos asegurar que las interacciones terminen en un tiempo corto, es necesario limitar un tiempo máximo para cada salida. Para ello se propone una aproximación a la $Kt$ de Levin. Primero definimos $\Delta ctime(U, p, a_{1:i})$ como el tiempo necesario para imprimir el par $\langle r_{i+1},o_{i+1} \rangle$ tras la secuencia de acciones $a_{1:i}$, es decir el tiempo de ciclo de respuesta. A partir de aquí se establece el límite superior para el tiempo de cómputo máximo que el entorno puede consumir para generar la recompensa y la observación después de la acción del agente.

\begin{definition} {\bf Complejidad $Kt$ ponderando los pasos de la interacción}

\[ Kt^{max}_U(\mu,n) := \min_{p \textrm{ tal que } U(p)=\mu} \left\{ l(p) + \log \left( \max_{a_{1:i}, i\leq n}(\Delta ctime(U,p,a_{1:i})) \right) \right\} \]
\end{definition}
\noindent  lo que significa la suma de la longitud del entorno más el logaritmo del máximo tiempo de respuesta de este entorno con la máquina $U$. Nótese que este límite superior puede usarse en la implementación de entornos, especialmente para hacer su generación computable. Para hacer esto definimos su complejidad para un número límite de ciclos $n$, haciendo su definición computable. Este límite $n$ no es solo necesario para la computación, también es práctico en algunos otros casos donde la computabilidad no es un problema pero no existe un máximo. Por ejemplo, considérese un entorno cuya salida $i$-ésima dependa del cálculo de si el número $i$ es primo o no. En este caso, el máximo $\Delta ctime$ no está delimitado y, por tanto el $Kt_U^{max}$ de esta secuencia sería infinito. Por lo tanto, el entorno sería descartado si no establecemos un límite de $n$.

La función de complejidad previa asegura que el tiempo de respuesta en cualquier interacción con un entorno está delimitado.

Ahora vamos a dar a una definición más refinada de la Inteligencia Universal:% utilizando los entornos sensibles a las recompensas, las recompensas simétricas, el entorno balanceado y la media de las recompensas:

\begin{definition} {\bf Inteligencia Universal (conjunto finito de entornos balanceados y sensibles a las recompensas, número finito de interacciones, complejidad $Kt^{max}$) con puntuación ajustada}
\[ \Upsilon(\pi, U, m, n_i) := \frac{1}{m \cdot n_i}\sum_{\mu \in S} v^\pi_\mu(n_i) \]
donde $S$ es un subconjunto finito de $m$ entornos balanceados siendo $n_i$-acciones sensible a las recompensas extraído con $p^t_U(\mu) := 2^{-Kt^{max}_U(\mu, n_i)}$.
\end{definition}

\section{Tests adaptativos}
Obviamente si queremos evaluar de una forma práctica la inteligencia de un agente, necesitamos que el test se haga en un periodo finito de tiempo y, por lo tanto, de entornos.

Aparentemente existen muchas opciones para incorporar el tiempo. Una aproximación podría ser, por ejemplo, utilizar un tiempo global $\tau$ para los entornos. Está claro que conforme más tiempo físico se deje, generalmente, un agente puede obtener más observaciones (muy rápidamente, p.~ej. moviéndose aleatoriamente al principio) y después tratar de usar su conocimiento durante el resto del test. Por lo tanto, los resultados esperados mejorarán conforme crezca $\tau$, mientras que se espera que empeoren conforme crezca la complejidad de los entornos (mientras $\tau$ se mantiene constante).

Utilizar dos parámetros distintos para el número de entornos ($m$) y el tiempo global de cada entorno ($\tau$) para cada entorno es un sesgo importante (aparte del efecto que tienen en la medida). Si se deja muchos entornos con un pequeño número de interacciones en cada uno, muchos agentes serán incapaces de encontrar el patrón y tomar ventaja de ello. Si se deja muchas interacciones para que los agentes encuentren el patrón solo se explorarían unos pocos entornos durante la evaluación. Por lo que existen también razones prácticas para encontrar una compensación entre los parámetros $m$ y $\tau$. Si se quiere evaluar un sistema con una alta inteligencia, estos valores deberían elegirse en concordancia: aparentemente deberían utilizarse pocos entornos pero más complejos y con más tiempo para interactuar en ellos. Por el contrario, si se quiere evaluar un sistema menos competente, deberían utilizarse muchos entornos simples con posiblemente menos tiempo de interacción. Por lo que esto sugiere ajustar los parámetros en función de las expectativas puestas en el sujeto que se quiera evaluar. La cuestión es que los tests no se deberían poder configurar independientemente antes de proporcionarlos, ya que la complejidad y el tiempo de los ejercicios deberían ser establecidos de antemano dependiendo del sujeto, y, por lo tanto, los resultados no serían comparables. En su lugar, es necesario que los tests sean adaptativos al nivel de inteligencia del examinado, y esta adaptación debe realizarse automáticamente y de igual modo para todos.

No solo se quiere que el test sea adaptativo, también que, dada una pequeña cantidad de tiempo poder aproximar la inteligencia del agente. Por otro lado, si se le diese más tiempo a la evaluación, se tendría una mejor aproximación, y también sería interesante poder parar la evaluación en cualquier momento y obtener la aproximación. En otras palabras, es preferible que el test sea \emph{anytime} (un algoritmo anytime es un algoritmo que puede pararse en cualquier momento, dando una recompensa razonable para el tiempo dado).

Una posibilidad es definir dos contadores: uno para la complejidad del entorno y otro para el tiempo físico que se puede interactuar en el entorno antes de cambiar a otro distinto. Ya que no se sabe la inteligencia o la velocidad del agente, se necesita comenzar con problemas muy simples y muy poco tiempo para interactuar con el entorno y que progresivamente se produzcan entornos más complejos y se vaya dejando más tiempo para cada entorno. Pero ya que muchos individuos solo seran capaces de aprender e interactuar razonablemente bien en algunos pequeños rangos de complejidad con un tiempo límite, debe haber un mecanismo que también disminuya la complejidad del entorno si el agente no obtiene buenas puntuaciones.

De este modo, con un test \emph{anytime} es posible parar el test en cualquier momento obteniendo una aproximación de la inteligencia del agente.

%Los test adaptativos no son solo, en la práctica, una buena idea para obtener una puntuación fiable en poco tiempo, sino que también son una salida al dilema de tener que elegir entre entornos simples o complejos y del tiempo requerido para cada uno de ellos. Los tests están diseñados para evaluar las habilidades de supuestos agentes inteligentes (artificiales, humanos, no-humanos, animales o extraterrestres) de un modo fiable.

%Aunque la idea es medir la inteligencia general (lo cual se conseguiría si se utilizasen máquinas universales no sesgadas para generar el conjunto de entornos), los tests que introducen aquí se pueden utilizar para evaluar capacidades más especializadas siempre que se elija apropiadamente la clase de entornos. Por ejemplo, si los entornos se eligieran de modo que las acciones solo afectasen a las recompensas pero no a las observaciones (entornos pasivos), los tests serían capaces de evaluar la capacidad de predecir secuencias (ignorando capacidades de planificación). De forma similar, utilizando distintas clases de entornos, se podría evaluar eficientemente el rendimiento en varias tareas (clasificación, laberintos, juegos de mesa, etc.) en muchas (si no en todas) las áreas de la inteligencia artificial.

\section{Clase de entornos Lambda ($\Lambda$)}
La medición de la inteligencia general requiere que los sujetos sean evaluados en un conjunto de problemas lo mas general posible, en donde puedan demostrar todas sus capacidades. Es por ello que en el proyecto ANYNT se ha diseñado una clase de entornos ($\Lambda$) \cite{HernandezOrallo09b}\footnote{Parte del material de esta sección ha sido extraído a partir de este artículo.} que se utilizará en este sistema de evaluación y que se espera sea lo suficientemente general. Es esta clase de entornos la que ponemos a prueba en esta tesis de máster; comprobando si es realmente adecuada para evaluar la inteligencia general o, por el contrario, sigue sin ser lo suficientemente general.

Esta clase de entornos trata de ser lo más general posible a la vez que incorpora las propiedades que vimos en el capítulo anterior para asegurar que los entornos generados sean realmente significativos para una evaluación apropiada. Los entornos generados por esta clase de entornos están compuestos por un espacio formado por celdas y sus conexiones entre celdas (acciones), y varios objetos (los agentes también son considerados objetos) que se sitúan en el espacio ocupando una de las celdas, con la posibilidad de moverse a través de su topología.

Empezaremos definiendo cómo deben ser el espacio y los objetos y, a partir de ahí, especificaremos las observaciones, las acciones y las recompensas de su clase de entornos. Después veremos qué propiedades posee la clase de entornos al haberlo diseñado de este modo y, finalmente, cómo hacen para generar estos entornos.

Antes de nada, es necesario definir algunas constantes que afectan a cada entorno. Estas constantes son: \mbox{$n_a = |A| \geq 2$} la cual se refiere al número de acciones, \mbox{$n_c \geq 2$} es el número de celdas y $n_\omega$ el número de objetos (sin incluir al agente que se va a evaluar ni a dos objetos especiales llamados \emph{Good} y \emph{Evil} que veremos más adelante).

\subsection{Espacio}
El espacio está definido como un grafo etiquetado dirigido de $n_c$ nodos (o vértices), donde cada nodo representa una celda. Los nodos están numerados desde $1$ hasta $n_c$ de la siguiente forma: $C_1, C_2, ..., C_{n_c}$. Desde cada celda tenemos $n_a$ flechas (o aristas) salientes, cada una de ellas se denota como \mbox{$C_i\rightarrow_\alpha{}C_j$}, lo cual significa que la acción \mbox{$\alpha \in A$} sale desde la celda $C_i$ hasta la celda $C_j$. Todas las flechas salientes desde $C_i$ se denotan como $(C_i)$. Una restricción del espacio es que en cada celda (como mínimo) dos de sus flechas salientes deben conducir a celdas distintas. Más formalmente \mbox{$\forall C_i : \exists r_1,r_2 \in (C_i)$} de tal modo que \mbox{$r_1 = C_i \rightarrow_{a_m}C_j$} y \mbox{$r_2 = C_i \rightarrow_{a_n}C_k$} con \mbox{$C_j \neq C_k$} y \mbox{$a_m \neq a_n$}. Además,  todas las celdas deben tener una acción reflexiva, de modo que \mbox{$\forall{}i : \exists{}r \in (C_i)$} de tal modo que \mbox{$r = C_i \rightarrow_{a_1} C_i$} (siendo $a_1$ siempre la acción reflexiva).

Un camino desde $C_i$ hasta $C_m$ es una secuencia de flechas \mbox{$C_i\rightarrow{}C_j$}, \mbox{$C_j\rightarrow{}C_k$}, \mbox{...}, \mbox{$C_l\rightarrow{}C_m$}. Todo grafo debe ser fuertemente conectado, de modo que todas sus celdas deben estar conectadas (p.~ej. debe haber un recorrido a través del grafo que pase por todos sus nodos), o, en otras palabras, para cada par de celdas $C_i$ y $C_j$ existe un camino desde $C_i$ hasta $C_j$ y viceversa. Nótese que esto implica que todas las celdas deben tener al menos una flecha de entrada (y de salida) desde (y hacia) cualquier otra celda.

Dada la definición previa, la topología del espacio puede ser muy variada. Puede incluir desde una simple cuadrícula\footnote{Se pueden diseñar todo tipo de cuadrículas, desde cuadrículas $2$ x $2$ hasta $n$ x $n$, pasando por cuadrículas toroidales y en 3D.} hasta topologías mucho más complejas dependiendo únicamente del número de celdas y de las flechas que existan entre sus celdas. En general el número de acciones $n_a$ es un factor que influencia mucho más en la topología que el número de celdas $n_c$.

Nótese, por ejemplo, que debido a que al menos una flecha debe ser reflexiva y al menos otra tiene que conducir a una celda diferente, entonces si $n_a = 2$ significa que el espacio necesariamente representa un anillo de celdas.

\subsection{Objetos}
Las celdas pueden contener objetos a partir de un conjunto de objetos predefinidos $\Omega$, con \mbox{$n_\omega = |\Omega|$}. Los objetos, denotados por $\omega_i$, pueden ser animados o inanimados, pudiéndose percibir en función de las reglas que posee cada objeto. Un objeto es inanimado (durante un cierto periodo o indefinidamente) cuando realiza la acción $a_1$ (acción reflexiva) de forma repetida. Los objetos pueden realizar acciones en el espacio siguiendo cualquier comportamiento, incluso pueden ser o no deterministas. Los objetos pueden ser reactivos y pueden actuar con diferentes acciones dependiendo de sus observaciones. Los objetos realizan una y solo una acción en cada interacción con el entorno (excepto los objetos especiales \emph{Good} y \emph{Evil}, los cuales pueden realizar varias acciones simultáneamente).

Aparte del agente evaluado $\pi$, como ya hemos mencionado existen dos objetos especiales llamados Good y Evil, representados por $\oplus$ y $\ominus$ respectivamente cuando son vistos por el agente evaluado $\pi$. Sin embargo, son indistinguibles para el resto de objetos (incluyéndose a sí mismos), por lo que en sus observaciones están representados por el mismo símbolo $\odot$.

Good y Evil deben tener el mismo comportamiento. Por el mismo comportamiento no se refieren a que realicen los mismos movimientos, sino que tienen una misma \emph{lógica} o \emph{programa} tras ellos que los hace actuar o moverse por el espacio. Nótese que Good y Evil se ven entre ellos de la misma forma (igual que el resto de objetos exceptuando a $\pi$). Por ejemplo, si Good tiene el programa ``Hacer la acción $a_2$ a menos que $\odot$ esté situado en la celda a la que lleva la acción $a_2$. En este caso, hacer la acción $a_1$'', entonces Evil debería tener el programa ``Hacer la acción $a_2$ a menos que $\odot$ esté situado en la celda a la que lleva la acción $a_2$. En este caso, hacer la acción $a_1$''. Para el primero $\odot$ se refiere a Evil y para el segundo $\odot$ se refiere a Good.

\label{MotivoEntornoEstocastico}
Los objetos pueden compartir una misma celda, excepto Good y Evil, los cuales no pueden estar en una misma celda al mismo tiempo. Si su comportamiento hace que ambos acaben en la misma celda, entonces uno (elegido aleatoriamente con la misma probabilidad) se mueve a la celda a la que pretendía llegar y el otro se mantiene en la celda en la que se encontraba anteriormente. Debido a esta propiedad el entorno es estocástico (no determinista).

Los objetos $\oplus$ y $\ominus$ pueden ejecutar muchas acciones en una sola interacción, realizando cualquier secuencia de acciones (no vacía). Una razón para hacer esto es evitar que el agente evaluado comience a seguir a Good como una forma fácil y óptima de obtener recompensas positivas en muchos entornos.

Al inicializarse el entorno los objetos se sitúan aleatoriamente en las celdas. Esta es otra fuente de comportamiento estocástico.

Aunque Good y Evil tienen el mismo comportamiento, la celda inicial en la que se asigna (aleatoriamente) a cada uno de ellos puede determinar una situación en donde su comportamiento sea muy simétrico con respecto al agente $\pi$. Por ejemplo, considérese el siguiente ejemplo:

\begin{example}\label{ex:motivo_ciclos}
Imagínese el siguiente espacio de la Figura~\ref{fig:anillo} y considérese un comportamiento de $\oplus$ y $\ominus$ de tal modo que realicen la acción $r$ si y solo si el agente $\pi$ comparte una celda con cualquiera de ellos (nótese que esto puede percibirse en las observaciones). De lo contrario, realizan la acción $l$.

\begin{figure}[h!]
\centering
\includegraphics{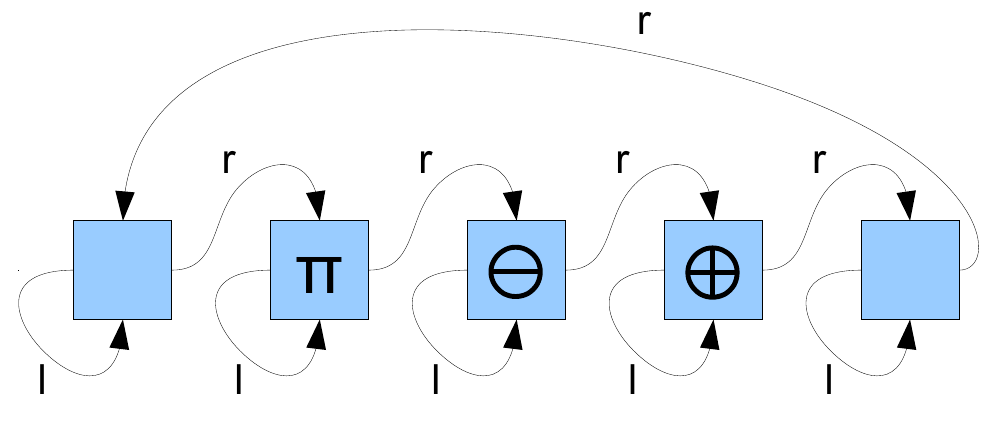}
\caption{Un espacio anillo donde el estado inicial puede ser crítico.}
\label{fig:anillo}
\end{figure}

A partir del estado representado en la Figura~\ref{fig:anillo}, es fácil ver que la situación relativa de los tres objetos solo se puede ver como ($\pi$, $\ominus$, $\oplus$) o cuando el agente comparte una celda con $\ominus$, seguido por $\oplus$ a la derecha ($\pi\ominus$, $\oplus$). Consecuentemente, en este entorno, es imposible para $\pi$ compartir una celda con $\oplus$, aun a pesar de que tanto $\oplus$ como $\ominus$ tienen el mismo comportamiento. En estos casos el estado inicial es crítico.
\end{example}

Siguiendo el ejemplo previo, en \cite{HernandezOrallo09b} definen una {``cláusula de ciclo''} la cual funciona de la siguiente forma: Dado un entorno con $n_a$ acciones, y $n_c$ celdas, se calcula un número aleatorio $n$ entre $1$ y ${n_c}^{n_a}$ (siguiendo una distribución uniforme), y entonces después de $n$ interacciones, se intercambian las posiciones de $\oplus$ y de $\ominus$. Entonces se computa otro nuevo número aleatorio (de la misma forma) y se vuelven a intercambiar las posiciones transcurrido ese nuevo número de interacciones, y así continuamente. La razón de que el número sea aleatorio es para evitar que el ciclo coincida con ningún otro ciclo o patrón que esté presente en el comportamiento de los objetos, como por ejemplo ``estate cerca de $\omega_1$ durante $n$ interacciones, tras lo cual debes cambiar de comportamiento y huir de $\omega_1$ hasta pasados $n$ nuevas interacciones y volver a seguir nuevamente a $\omega_1$''. El objetivo de esta cláusula es evitar la relevancia del estado inicial en el entorno, ya que, como hemos visto anteriormente, podría ser crítico.

Finalmente, las primeras interacciones con el entorno pueden tener lo que los autores llaman ``basura inicial'' \cite{HernandezOrallo09aTR}. Considérese, p.~ej. un comportamiento para $\oplus$ y $\ominus$ el cual sea ``empezar haciendo $a_1a_2a_0a_1a_1a_1a_0a_2a_2a_0a_1a_1a_0a_0$ y a partir de este momento hacer siempre $a_0a_1$''. La primera parte del comportamiento sigue una secuencia que a primera vista parece aleatoria y solo cuando se alcanza el patrón (la segunda parte del comportamiento), tiene sentido comenzar a evaluar el comportamiento de un agente. Consecuentemente, se sugiere dejar que un agente aleatorio interactúe durante $n$ interacciones para sobrepasar mucha de la basura inicial (si existiera) y entonces empezar con la evaluación. El valor de $n$ se puede obtener de la misma forma que se hizo para el valor de la ``cláusula de ciclo''.

\subsection{Recompensas}
Para las recompensas se trabajará con la noción de \emph{recompensa de la celda} denotada por $r(C_i)$ y la noción de \emph{rastro}. Inicialmente, \mbox{$r(C_i) = 0$} para todo $i$. Las recompensas de las celdas se actualizan en cada movimiento de $\oplus$ y de $\ominus$. En cada interacción se deja caer $r_i^\oplus$ en la celda en la que se encuentre $\oplus$ y se deja caer $-r_i^\ominus$ en la celda en la que se encuentre $\ominus$ y el resto de recompensas se dividen entre $2$. Por lo tanto, una forma intuitiva de ver esto es imaginando que $\oplus$ y $\ominus$ van dejando un rastro positivo y negativo de recompensas respectivamente. El agente $\pi$ se \emph{come} las recompensas que va encontrando en las celdas que ocupa. Con que se \emph{come} se quiere decir que, tras obtener la recompensa, la recompensa de la celda se establece a $0$. Si $\oplus$ o $\ominus$ llegan a una celda en donde se encuentre $\pi$, la recompensa de la celda se mantiene a $0$ mientras ambos objetos compartan la celda. En otras palabras: $\oplus$ y $\ominus$ no dejan caer recompensas si comparten la celda con $\pi$.

Los valores de $r_i^\oplus$ y de $-r_i^\ominus$ que $\oplus$ y $\ominus$ dejan caer en el resto de situaciones también forman parte del comportamiento de $\oplus$ y de $\ominus$ (el cual es el mismo, sin que esto signifique que \mbox{$r_i^\oplus = r_i^\ominus$} para todo $i$). Solo se impone una limitación en como se generan estos valores, \mbox{$\forall{}i : 0 < r_i^\oplus \leq 1$} y \mbox{$\forall{}i : 0 < r_i^\ominus \leq 1$}. Una elección típica es simplemente hacer que \mbox{$r_i^\oplus = r_i^\ominus = 1/2$} para todo $i$, aunque se necesitan comportamientos de las recompensas mucho más complejos para codificar entornos también más complejos. Finalmente, nótese que las recompensas no son parte de las observaciones, por lo que no se pueden observar (directamente) por ningún objeto (incluyendo a $\pi$).

Cuando $\pi$ se mueve a una celda, obtiene la recompensa de la celda que se encuentra en esa celda, p.~ej. la recompensa acumulada se actualiza de tal modo que \mbox{$\rho = \rho + r(C_i)$}.

Con este modo de calcular las recompensas, el orden completo de las recompensas, observaciones y acciones se puede ver de la siguiente forma:

\begin{enumerate}
\itemsep=0pt
\item \mbox{$\rho = 0$}
\item \mbox{$\forall{}i : r(C_i) \leftarrow 0$}
\item Situar los objetos aleatoriamente de modo que $\oplus$ y $\ominus$ estén en celdas diferentes.
\item Se le proporciona $\rho$ a $\pi$.
\item \mbox{$i \leftarrow 0$}
\item\label{it:buclePrincialClaseLambda} \mbox{$i \leftarrow i + 1$}
\item Para cada $C_j$ comprobar si contiene a $\oplus$ y no contiene a $\pi$. En ese caso, \mbox{$r(C_j) \leftarrow +r_i^\oplus$}.
\item Para cada $C_j$ comprobar si contiene a $\ominus$ y no contiene a $\pi$. En ese caso, \mbox{$r(C_j) \leftarrow -r_i^\ominus$}.
\item Se producen y se entregan las observaciones a todos los objetos.
\item El objeto $\pi$ actúa y se mueve a una celda $C_\pi$.
\item El resto de objetos actúa.
\item \mbox{$\rho \leftarrow \rho + r(C_\pi)$}
\item \mbox{$r(C_\pi) \leftarrow 0$}
\item \mbox{$\forall{}j : r(C_j) \leftarrow r(C_j)/2$}
\item Si no ha terminado la evaluación volver a \ref{it:buclePrincialClaseLambda}.
\item \mbox{$\rho \leftarrow \rho/i$}
\end{enumerate}

\subsection{Codificación de observaciones y acciones}\label{sec:observaciones}
Las observaciones son una secuencia de celdas con su contenido. Las celdas están ordenadas en función de su número. Cada elemento en la secuencia muestra la presencia o ausencia de cada objeto, incluido el agente evaluado. Adicionalmente, cada celda alcanzable (utilizando una acción desde la posición actual) se muestra en la observación. En particular, el contenido de cada celda es una secuencia de objetos, donde $\pi$ debe aparecer antes que $\oplus$ y $\ominus$, y después aparecen el resto de objetos en orden en función de su índice. Después vienen las posibles acciones también ordenadas por sus índices, y denotadas con $A_i$, en lugar de $a_i$. La razón de utilizar un alfabeto distinto en las observaciones se verá más adelante. Cada secuencia se separa del resto por el símbolo `:'.

Por ejemplo, si tenemos \mbox{$n_a=2$}, \mbox{$n_c=4$} y \mbox{$n_\omega=2$} entonces la siguiente secuencia \mbox{`$\pi\omega_2A_1 : \ominus : \oplus\omega_1A_2 :$'} es una posible observación para el agente evaluado $\pi$. El significado de la secuencia es que en la celda 1 tenemos al agente evaluado y al objeto $\omega_2$, en la celda 2 tenemos a Evil, en la celda 3 tenemos a Good y al objeto $\omega_1$ y la celda 4 está vacía. Además, podemos ver que utilizando la acción $a_1$ nos quedamos en la misma celda y podemos ir a la celda 3 con la acción $a_2$. Para el resto de objetos (incluyendo a Good y a Evil) la misma observación se vería como \mbox{`$\pi\omega_2A_1 : \odot : \odot\omega_1A_2 :$'}.

\subsection{Propiedades}\label{sec:PropiedadesClaseEntorno}
En esta sección demostramos que la clase de entornos cumple con las propiedades descritas en secciones anteriores. Las propiedades que veremos son: (1) la clase de entornos es sensible a las recompensas; el agente puede realizar acciones de modo que pueda afectar a las recompensas que reciba (véase la Definición~\ref{def:SensibleRecompensas} en la Página~\pageref{def:SensibleRecompensas}) y (2) la clase de entornos está balanceada; un agente aleatorio tendría una recompensa acumulada esperada igual a 0 (véase la Definición~\ref{def:EntornoBalanceado} en la Página~\pageref{def:EntornoBalanceado}). Las demostraciones para esta clase de entornos han sido extraídas de \cite{HernandezOrallo09b}.

A continuación vemos la proposición que demuestra como la clase de entornos es sensible a las recompensas:

\begin{proposition}\label{pro:SensibleRecompensas}
La clase de entornos $\Lambda$ es sensible a las recompensas.
\begin{proof}
Ya que el grafo está fuertemente conectado, todas las celdas son alcanzables por $\pi$. Para demostrar la proposición tenemos que proceder por casos:

\begin{enumerate}
\item Si la celda $C_j$ donde va a parar $\pi$ tiene una recompensa distinta de $0$, p.~ej. \mbox{$r(C_j) \neq 0$}. Entonces el agente puede quedarse quieto (siempre existe esta acción en cualquier entorno) poniendo la recompensa de la celda a $0$. En este caso, vamos a los siguientes casos:
\item Si la celda $C_j$ donde va a parar $\pi$ tiene una recompensa $0$, p.~ej. \mbox{$r(C_j) = 0$}, y tanto $\oplus$ como $\ominus$ se encuentran en otra celda diferente, digamos $C_k$ y $C_l$. Entonces el agente tiene las siguientes opciones:

\begin{itemize}
\item Quedarse siempre quieto: Obteniendo siempre una recompensa de 0 (independientemente de si $\oplus$ o $\ominus$ vayan a la celda, ya que no pueden dejar recompensas si comparten celda con $\pi$).
\item Moverse a otra celda: En este caso, ya que debe existir un camino (el grafo es fuertemente conectado) a ambas celdas ($C_k$ y $C_l$), si va hasta $C_k$, pueden ocurrir muchas cosas distintas por el camino:

\begin{itemize}
\item Si $\pi$ obtiene alguna recompensa en este periodo: entonces sabemos que dejar $C_j$ nos puede dar una recompensa distinta que quedarnos en ella.
\item Si no: entonces llega a $C_k$ sin ningún cambio en sus recompensas. Pero ni $\oplus$ ni $\ominus$ pueden dejar la recompensa de $C_k$ a $0$ ya que para todo $i$, \mbox{$0 < r_i^\oplus \leq 1$} y \mbox{$0 < r_i^\ominus \leq 1$} y no dejan de dar recompensas hasta que se encuentren con $\pi$ en la misma celda (nótese que $\oplus$ y $\ominus$ dejan caer sus recompensas después de que $\pi$ haya consumido la recompensa).
\end{itemize}
\end{itemize}

Por lo que en este caso, el agente $\pi$ puede elegir entre una recompensa de $0$ y otra distinta a $0$.

\item Si la celda $C_j$ donde va a parar $\pi$ tiene una recompensa de $0$, p.~ej. \mbox{$r(C_j) = 0$}, y $\ominus$ comparte la celda con él y $\oplus$ está en una celda distinta $C_k$. Entonces el agente tiene las siguientes opciones:

\begin{itemize}
\item Quedarse siempre quieto: obteniendo siempre una recompensa igual a 0.
\item O moverse a otra celda: Si se mueve a otra celda debe existir un camino (el grafo es fuertemente conectado) a la celda $C_k$. Pueden ocurrir muchas cosas distintas por el camino, pero ni $\oplus$ ni $\ominus$ pueden poner esta recompensa a $0$. Por lo tanto, $\pi$ obtiene una recompensa positiva o negativa distinta de $0$.
\end{itemize}
Entonces, en este caso, $\pi$ puede elegir entre una recompensa de $0$ y otra distinta a $0$.

\item Si la celda $C_j$ donde va a parar $\pi$ tiene una recompensa de $0$, p.~ej. \mbox{$r(C_j) = 0$}, y $\oplus$ comparte la celda con él y $\ominus$ está en una celda distinta $C_k$. Este caso es simétrico al anterior y se justifica de la misma manera.
\end{enumerate}

Ya que los 4 casos previos contemplan todas las posibles opciones de manera exhaustiva y todos ellos permiten obtener recompensas distintas dependiendo de las acciones que $\pi$ elija, el entorno es sensible a las recompensas.
\end{proof}
\end{proposition}

Y finalmente veamos que los entornos están balanceados:

\begin{proposition}\label{pro:Balanceada}
La clase de entornos $\Lambda$ está balanceada.
\begin{proof}
Por la definición de $\Lambda$, una recompensa siempre se encuentra entre $[-1, 1]$ y la recompensa acumulada $\rho$ también se encuentra entre $[-1, 1]$ ya que dividimos por el número de interacciones. También sabemos que los objetos especiales $\oplus$ y $\ominus$ tienen el mismo comportamiento. El resto de objetos (excepto $\pi$) no puede percibir distinción alguna entre $\oplus$ y $\ominus$. Por lo tanto, las posibilidades de que $\oplus$ y $\ominus$ sean más favorables para $\pi$ o de que se comporten de forma distinta solo puede depender de dos cosas: el propio agente $\pi$ y el estado inicial. Como el estado inicial se elige de forma aleatoria, para un agente aleatorio que elige entre sus posibles acciones de forma aleatoria, el valor medio esperado de las recompensas para cada celda es $0$, ya que la probabilidad de cualquier acción y estado por parte de $\oplus$ y $\ominus$ es igualmente probable que su estado simétrico (todo igual excepto las posiciones y rastros de $\oplus$ y de $\ominus$, que se intercambian). Por lo tanto, la recompensa acumulada esperada es $0$.
\end{proof}
\end{proposition}

Nótese que la proposición anterior se mantiene debido a la elección aleatoria del estado inicial. Por ejemplo, en el Ejemplo~\ref{ex:motivo_ciclos}, el estado mostrado en la Figura~\ref{fig:anillo} muestra un estado inicial que hace que la recompensa esperada para un agente aleatorio no sea $0$ (es negativa). A continuación vemos si podemos obtener un resultado que diga que el entorno siempre está balanceado (y no solo debido a la elección aleatoria de las posiciones iniciales de los objetos).

\begin{proposition}\label{pro:FuertementeBalanceada}
La clase de entornos $\Lambda$ está fuertemente balanceada, por ejemplo, para cualquier interacción la recompensa acumulada esperada de un agente aleatorio (a partir de esa interacción) es $0$.
\begin{proof}
La demostración se basa en la anterior demostración. Sabemos que la recompensa acumulada esperada para un agente aleatorio solo depende de su estado inicial. Dependiendo del estado inicial, el valor esperado puede ser distinto de $0$ sin considerar intercambios. Pero, debido a la cláusula de ciclo, siempre hay una interacción en donde $\oplus$ y $\ominus$ se intercambian. Ya que la longitud del ciclo se elige de modo aleatorio y es independiente del comportamiento de $\oplus$ y $\ominus$, y que el agente se comporta de modo aleatorio (y por lo tanto independientemente del comportamiento de $\oplus$ y $\ominus$ y sus intercambios), entonces tenemos que en cualquier interacción $j$ podemos considerar una secuencia de recompensas \mbox{$r_j$,$r_{j+1}$,...}, ignorando los intercambios. Si todos suman $0$, entonces la proposición está probada. Si las recompensas acumuladas dan un valor positivo $p$, es porque en el límite hay más positivas que negativas (debido a la media del valor acumulado). Por lo tanto, cualquier futuro intercambio dará un valor de $-p$ (ignorando todos los siguientes intercambios). Recurrentemente, tenemos una secuencia infinita de \mbox{$p$,$-p$,$p$,$-p$,...}, cuyo valor esperado es 0. Se razona de igual modo cuando el valor esperado inicial es negativo.
\end{proof}
\end{proposition}

Las proposiciones previas muestran que el entorno sigue las propiedades requeridas para su uso en un test anytime \cite{HernandezOralloDowe2010}. Esto es importante, ya que se requiere que los entornos sean discriminativos (y que sean sensibles a las observaciones y a las recompensas es una condición importante para ello) y también se quiere que las recompensas no estén sesgadas (no se quiere que los agentes aleatorios saquen buenos resultados).

Otra propiedad que no es tan relevante para el test pero si que es interesante desde un punto de vista teórico (y especialmente para comparar algunas habilidades) es si la clase de entornos previa permite `aliasing perceptual'. La respuesta es un rotundo sí. Dada la complejidad del comportamiento de los objetos, incluso una observación completa del espacio no es suficiente (aun sabiendo el comportamiento de los objetos), ya que no conocemos su estado interno ni su memoria interna.

\subsection{Generación del entorno}
Finalmente, es necesario poder generar los entornos de manera automática y también poder calcular su complejidad. Para generar un entorno, es necesario generar el espacio (las constantes y la topología) y el comportamiento para todos los objetos (excepto el de $\pi$). Una primera aproximación podría ser utilizar (como de costumbre) gramáticas generativas (al menos esta aproximación parece fácil para construir el espacio). Sin embargo, si se va a utilizar una gramática generativa para codificar el espacio (y, por lo tanto, para calcular su complejidad), entonces podría elegirse una gramática generativa que no fuera universal aunque generase espacios válidos. En \cite{HernandezOralloDowe2010} los autores sugieren que esta idea no es del todo correcta. Con una gramática generativa no universal podría ocurrir que la complejidad de un espacio con $100$ celdas idénticas fuera $100$ veces la complejidad de un espacio con una sola celda. Pero esta es una complejidad que va en contra de la noción de la complejidad Kolmogorov (la cual se utilizará para medir la complejidad del entorno, véase la Sección~\ref{sec:ComplejidadKolmogorov}) y del ordenamiento de entornos que se está buscando. Nótese que la probabilidad de un entorno $e$ debe basarse en $K(e)$ (complejidad Kolmogorov de e) y no en su longitud. Una cuadrícula regular de \mbox{$100$ x $100$} es generalmente más simple que un espacio más enrevesado con un número mucho menor de celdas.

En su lugar en \cite{HernandezOrallo09b} proponen utilizar gramáticas generativas universales, en particular proponen utilizar algoritmos de Markov (véase la Sección~\ref{sec:AlgoritmoMarkov}\footnote{Aunque el algoritmo típicamente elige para el tercer paso del algoritmo la subcadena situada más a la izquierda (de entre las subcadenas en $s$ que concuerdan con la regla izquierda) para ser reemplazada, modifican el algoritmo para que elija la subcadena situada más a la derecha. El motivo que dan es que sus cadenas representan la interacción con entornos, las cuales normalmente crecen hacia la derecha, por lo que esta parte suele ser más importante. Este es un cambio completamente simétrico, por lo que cambiar esto no es más que una convención que no cambia el poder expresivo de estas máquinas.}).

Para entender mejor como se puede utilizar esta gramática y ver su capacidad expresiva, tras explicar como se codifican el espacio y los objetos, codificaremos los algoritmos de un entorno reactivo complejo.

\subsubsection{Codificar y generar espacios}\label{sec:MCCodYGenEspacios}
Primero se necesita obtener el número de celdas $n_c$ y de acciones $n_a$ a partir de alguna distribución utilizando cualquier codificación estándar para números naturales (p.~ej. la función \emph{log*} en \cite{Rissanen83}\cite{Wallace05}). El grafo del espacio se define con un algoritmo de Markov sin ninguna restricción en su definición, pero con la siguiente post-condición. El espacio generado tiene que representarse con una cadena siguiendo esta definición (utilizamos lenguaje regular para esta definición) $a_1a_2(+^*|-^*)$\dots$a_{n_a}(+^*|-^*)$ para cada celda dividiéndolas con un carácter separador `:'. Cada segmento entre `:' enumera a una celda, y la información en cada celda se compone de las flechas salientes (más precisamente la secuencia de acciones y a qué celda conduce cada acción, calculado por el número de símbolos $+$ o $-$. Esta sucesión de $+$ o $-$ es el desplazamiento desde la celda actual). Se utiliza un índice toroidal donde, p.~ej. \mbox{$1 - 2 = n_c - 1$} y se omite el desplazamiento cuando la acción conduce a la misma celda.

Con la descripción y definiciones anteriores es posible \emph{codificar} (inequívocamente) cualquier espacio. Sin embargo, si se quiere generar espacios utilizando algoritmos de Markov, la cosa es mucho más difícil, ya que una generación aleatoria de las reglas de Markov puede generar cadenas que no representen ningún espacio. Aunque pueden existir optimizaciones, ya que no se quiere perder generalidad ni la capacidad de medir la complejidad, los autores proponen dejar al algoritmo ejecutarse un número limitado de iteraciones y entonces pasarle un post-procesamiento (eliminar acciones repetidas, símbolos inválidos, etc.) y entonces comprobar si la cadena resultante representa a un espacio válido (sintáctica y semánticamente).

Para el cálculo de la complejidad, cualquier aproximación de la longitud del algoritmo de Markov (p.~ej. el número de símbolos del algoritmo entero) sería válido como una aproximación de su complejidad.

\subsubsection{Codificar y generar objetos}
Para generar los objetos, primero se obtiene el número de objetos $n_\omega$ a partir de alguna distribución y se codifica utilizando cualquier codificación para los números naturales como se menciona anteriormente. Después se tienen que obtener (con una distribución uniforme) y codificar sus celdas iniciales, codificando $n_\omega + 3$ números naturales ($n_w$ para los objetos + 3 para $\pi$, $\oplus$ y $\ominus$) delimitados entre $1$ y $n_c$ (no necesariamente diferentes). En el caso de que $\oplus$ y $\ominus$ estén en la misma celda, se recalculan sus números.

El comportamiento (que debe ser universal) es generado por otro algoritmo de Markov. Para disminuir la cantidad de reglas del algoritmo y mejorar su comprensión, vamos a enriquecer la notación con los símbolos: \mbox{`$[]$' = `opcionalidad'}, \mbox{`$( | )$' = `varias posibilidades'}, `$\chi$' = `cualquier símbolo del alfabeto del algoritmo' y `$(\chi\not\in \{\})$' = `cualquier símbolo del alfabeto del algoritmo excepto'. Si alguno de estos símbolos aparece con un subíndice (p.~ej. $(A_1|\pi)_1$) significa que todas las apariciones del símbolo con ese subíndice están ligadas (es decir, todas sus apariciones deben ser sustituidas por el mismo símbolo).

Veamos un ejemplo para entender como se interpretan estos símbolos. Con $n_a = 2$ y $n_w = 0$, la regla \mbox{$[(\chi_1\not\in \{A_1, \pi\})]A_1(\pi|\odot)_2 \rightarrow (\pi|\odot)_2[\chi_1]$} se sustituirá por las siguientes reglas de Markov:

\begin{enumerate}
\itemsep=0px
\item $A_2A_1\pi \rightarrow \pi{}A_2$
\item $A_2A_1\odot \rightarrow \odot{}A_2$
\item $\odot{}A_1\pi \rightarrow \pi\odot$
\item $\odot{}A_1\odot \rightarrow \odot\odot$
\item $A_1\pi \rightarrow \pi$
\item $A_1\odot \rightarrow \odot$
\end{enumerate}

Solo con la observación actual los objetos no tendrían acceso a la memoria y estarían muy limitados, por lo que se añaden dos símbolos especiales $\Delta$ y $\nabla$ que significan respectivamente: poner la cadena actual en memoria o recuperar la última cadena insertada en memoria. Además, si $\Delta$ se sitúa encima de algún símbolo, solo éste se guardará en memoria (esto solo ocurrirá si la regla ha sido lanzada); p.~ej. $A_1^\Delta$ solo guardará $A_1$. $\nabla$ se sustituye por la cadena recuperada de la memoria.

La cadena de entrada del algoritmo es la observación que le proporciona el entorno, codificada como hemos visto en la Sección~\ref{sec:observaciones}. Una vez el algoritmo de Markov se ha ejecutado, todos los símbolos que no están en el conjunto de acciones $A$ se eliminan de la cadena. Entonces, para todos los objetos (excepto para $\oplus$ y $\ominus$), la acción más a la derecha en la cadena es la acción que se realiza. Para $\oplus$ y $\ominus$ se utiliza toda la cadena.

Para el cálculo de la complejidad, sería válida alguna aproximación de la longitud del algoritmo (p.~ej. el número de símbolos del algoritmo entero).

Es importante remarcar que la complejidad de un conjunto de objetos no es la suma de las complejidades de los objetos. Alternativamente, los autores sugieren utilizar un bit extra en cada regla indicando si la regla es nueva o no y añadir identificadores a las reglas que se reutilizan para tener una forma de referirse a otras reglas en otros objetos. Consecuentemente, si 100 objetos comparten muchas reglas, la complejidad de todo el conjunto será mucho más pequeña que la suma de las partes.

\subsubsection{Codificación de un entorno complejo}
\begin{figure}[h!]
\centering
\includegraphics[width=1\textwidth]{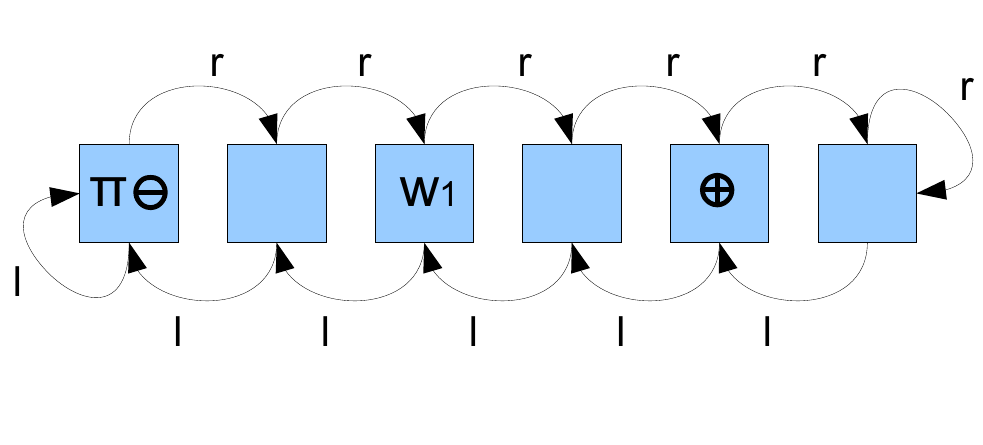}
\caption{Situación inicial de un entorno reactivo con un objeto ($\omega_1$) y con comportamiento complejo.}
\label{fig:EntornoCodificarInicial}
\end{figure}

Tras ver cómo se codifican los entornos vamos a mostrar un ejemplo para ver cómo se codificaría el entorno reactivo complejo de la Figura~\ref{fig:EntornoCodificarInicial}. En este entorno los objetos \mbox{Good ($\oplus$)} y \mbox{Evil ($\ominus$)} se mueven siempre tres celdas a la izquierda seguido de tres celdas a la derecha (una celda por cada acción) a menos que vayan a situarse en la misma celda en la que se encuentra el objeto $\omega_1$, en cuyo caso cambian su acción y se mantienen en la celda. El objeto $\omega_1$ se mantiene siempre en la misma celda a menos que coincida con $\pi$, en cuyo caso cambiará de celda del siguiente modo: si $\pi$ hubiera venido desde su acción `l' utilizará la acción `r' para alejarse y viceversa. Si solo existe una acción con la que abandonar la celda, será la que utilice independientemente de por donde hubiera venido $\pi$.

Lo primero que haremos será codificar el espacio. Veamos un algoritmo de Markov que lo genera.

\begin{enumerate}
\itemsep=0px
\item $L \rightarrow lr+:$
\item $C \rightarrow l-r+:$
\item $R \rightarrow\cdotp l-r$
\item $\rightarrow LCCCCR$
\end{enumerate}

Tras disparar la secuencia de reglas \mbox{4-1-2-2-2-2-3}, fácilmente vemos como la cadena generada \mbox{(\footnotesize \ttfamily lr+:l-r+:l-r+:l-r+:l-r+:l-r)} representa claramente al espacio deseado.

A continuación codificaremos todos los objetos presentes en el entorno. Veamos un algoritmo de Markov que genera el comportamiento de $\omega_1$.

\begin{enumerate}
\itemsep=0px
\item $S(L|R)^\Delta \rightarrow S$
\item $\pi[\odot]\omega_1 \rightarrow \nabla\Phi$
\item $\pi(\chi\not\in\{L, R, :\}) \rightarrow \pi$
\item $\pi(L|R)^\Delta \rightarrow$
\item $R\Phi \rightarrow\cdotp l$
\item $L\Phi \rightarrow\cdotp r$
\item $\rightarrow\cdotp s$
\end{enumerate}

Generar algún comportamiento para los objetos requiere de algoritmos más complejos. Una forma de codificar comportamientos es dividirlo en función de subcomportamientos más simples. En este algoritmo, la primera regla comprueba si alguna de las acciones $L$ o $R$ dejaría al objeto en la misma celda y, en este caso, guarda en la memoria que no podrá utilizarla para huir. La tercera regla elimina símbolos innecesarios de la cadena que representa a la celda en la que se encuentra $\pi$. La cuarta regla comprueba si $\pi$ está en alguna celda adyacente y, si fuera el caso, se guarda en memoria con qué acción le ha visto. La segunda regla comprueba si $\pi$ se encuentra con el objeto y, si se diera el caso, recupera el símbolo de la memoria (en verdad este símbolo representa la acción que \emph{no} deberá hacerse bajo ninguna circunstancia) y marca en la cadena que se deberá realizar un movimiento para escapar de $\pi$ ($\Phi$). La quinta y sexta reglas se lanzan únicamente cuando se ejecutó la segunda regla, es decir, cuando se detectó que $\pi$ se encontraba en la misma celda que $\omega_1$. Cuando se da esta situación se comprueba: (1) cuál fue la acción desde la que vino $\pi$ (si se lanzó la cuarta regla) o (2) la acción que no permitía al objeto huir de la celda (si se lanzó la primera regla). En ambos casos se ejecuta la acción contraria para escapar de $\pi$. La última acción se ejecutará siempre que el objeto no coincida con $\pi$.

Y finalmente codificamos el comportamiento de $\oplus$ y $\ominus$. Veamos un algoritmo de Markov que genera su comportamiento.\footnote{Hay que recordar que ambos objetos deben tener el mismo comportamiento, por lo que ambos comparten el mismo algoritmo de Markov.}

\begin{enumerate}
\itemsep=0px
% Aquí se comprueba si el objeto \omega_1 nos interrumpe el paso.
\item $\omega_1(\chi\not\in\{L, R, :\}) \rightarrow \omega_1$ % Borra símbolos innecesarios. Esta regla es innecesaria para este entorno, pero hace el comportamiento más general.
\item $\omega_1L \rightarrow E_I$ % Si \omega_1 está a la izquierda me apunto que esperar (E_I) si tenía que mover hacia  en esta interacción.
\item $\omega_1R \rightarrow E_D$ % Si \omega_1 está a la derecha me apunto que tengo que esperar (E_D) si tenía que mover hacia  en esta interacción.
% Aquí eliminamos los símbolos innecesarios para dejar la memoria limpia al guardar.
\item $(\pi|\omega_1|S|L|R|:) \rightarrow $ % Eliminar todos los símbolos innecesarios de la observación.
\item $\odot(E_I|E_D)_1 \rightarrow (E_I|E_D)_1\odot$ % Guardar E_I o E_D a la izquierda
\item $\odot\odot \rightarrow \Phi_I\nabla\Phi_F$ % Una vez eliminados todos los símbolos empezar a preprocesar.
%Comienza el preproceso para borrar la última acción que se hizo.
\item $[E_I|E_D]H(I|D)\Phi_F \rightarrow \Phi_F$ % Eliminar la última acción que se hizo.
\item $(I|D)_1\Phi_F \rightarrow \Phi_F(I|D)_1$ % Las acciones D o I no se tocan durante el preproceso.
%Elije la acción a realizar en esta interacción.
\item $\Phi_I\Phi_F \rightarrow H\Delta$ % Se terminó el preproceso, ahora elegimos la acción.
\item $(E_IHI|E_DHD) \rightarrow\cdotp s$ % Si la acción que toca hacer se marcó como que no se podía I = E_I o D = E_D, entonces quedarse en la celda.
\item $HI  \rightarrow\cdotp l$ % Hacer movimiento a la izquierda.
\item $HD \rightarrow\cdotp r$ % Hacer movimiento a la derecha.
% Si se ha acabado el patrón volver a empezarlo.
\item $H \rightarrow HIIIDDD\Delta$ % Si se ha gastado todo el patrón, volver a empezarlo.
\end{enumerate}

Para comprender este algoritmo hay que tener dos conceptos claros: (1) Debemos tener guardado en memoria el patrón que queremos realizar y cuál fue la última acción que hicimos. Esta información deberá guardarse en cada ejecución del algoritmo. (2) El algoritmo, tras recuperar el patrón de la memoria, debe ser capaz de reconocer que la primera acción ya se realizó en la interacción anterior y deberá eliminarlo antes de proseguir con la siguiente acción.

Con estos dos conceptos en mente podemos dividir el algoritmo en cuatro secciones. La primera sección (reglas 1, 2 y 3) comprueba si se puede llegar a $\omega_1$ con alguna de las acciones $L$ o $R$ y se marca que deberá evitarse realizar estas acciones con los símbolos $E_I$ y $E_D$ respectivamente. La segunda sección (reglas desde la 4 hasta la 6 izquierda) elimina el resto de símbolos de la cadena (nótese que solo queremos guardar el patrón de movimientos, por lo que hay que evitar mantener información innecesaria) y mantenemos el símbolo $E_I$ o $E_D$ a la izquierda, ya que es necesario para decidir la acción en esta interacción. La tercera sección (reglas desde la 6 derecha hasta la 9 izquierda) procesan el patrón recuperado de memoria para eliminar la acción que se hizo en la interacción anterior. Y finalmente, la cuarta sección (reglas desde la 9 derecha hasta la 13) guarda el patrón y la acción actual en memoria y selecciona la acción a realizar.

\begin{figure}[h!]
\centering
\includegraphics[width=1\textwidth]{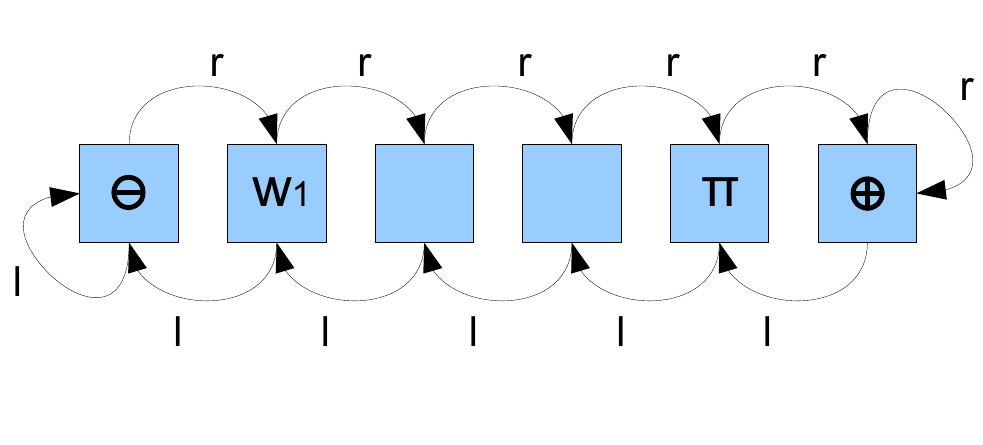}
\caption{Situación ideal para $\pi$ en el entorno reactivo de la Figura~\ref{fig:EntornoCodificarInicial}.}
\label{fig:EntornoCodificarIdeal}
\end{figure}

En la Figura~\ref{fig:EntornoCodificarIdeal} vemos la situación ideal a la que se puede llegar si $\pi$ realiza una secuencia de movimientos adecuada. Ya que $\omega_1$ bloquea el paso, \mbox{$\ominus$ (Evil)} no se moverá nunca de su celda y \mbox{$\oplus$ (Good)} dispondrá de las suficientes celdas para moverse sin ser bloqueado por $\omega_1$.

\paginablanco

\chapter{Aproximación implementada del sistema de evaluación}\label{cap:Aproximacion}
El marco conceptual anteriormente descrito, es una idea general de cómo se pueden construir tests universales. En lugar de implementar los tests tal y como sugieren los autores, en un trabajo previo\footnote{http://users.dsic.upv.es/proy/anynt/Memoria.pdf} se construyó gran parte de la arquitectura con algunas simplificaciones pero, sin por ello, renunciar a las características y propiedades más importantes que debe tener un test supuestamente universal. Por una parte se mantuvo la estructura principal: agentes, entornos y espacios y las interacciones entre ellos, mientras que por otro lado se simplificaron otros aspectos como: el comportamiento de los agentes, la generación de los espacios y la construcción de las observaciones.

En esta sección describimos la arquitectura que se realizó en este trabajo previo e incluimos las modificaciones que se han realizado en esta tesis de máster. En concreto, del trabajo previo vemos: (1) las principales diferencias entre el marco conceptual antes descrito y la implementación realizada, (2) la arquitectura construida del sistema de evaluación y (3) dos modos de uso del sistema de evaluación (uno construido para asistir en su desarrollo, y otro para evaluar algoritmos de IA). Además, en esta tesis se ha añadido: (1) varias modificaciones que dotan de mayor flexibilidad al sistema de evaluación, (2) una primera versión de un interfaz gráfico con el que evaluar a los seres humanos (esta interfaz ha sido diseñada con el objetivo de que fuera lo mas apropiada e intuitiva posible) y (3) se ha implementado un algoritmo de referencia de IA conocido como Q-learning, con el cual probaremos a la clase de entornos.

\section{Principales diferencias entre el marco conceptual y la implementación realizada}
En esta sección vemos las principales diferencias de la aproximación implementada del sistema de evaluación con respecto al marco general del capítulo anterior.

\subsection{Objetos y agentes}
Hemos hecho una jerarquización distinta entre objetos y agentes. Ambos son objetos, pero si el objeto es animado (es decir, tiene un comportamiento que le permite moverse a través del espacio) entonces es un agente. Este cambio no tiene ninguna implicación en el sistema, es una cuestión meramente de implementación.

\subsection{Comportamiento de los agentes}
De momento no hemos utilizado ningún lenguaje de especificación con el que generar el comportamiento de los agentes. Para ello, hemos implementado a mano varios tipos de comportamientos estándar: movimiento aleatorio, patrón de movimientos, etc.

Por otro lado, Good y Evil siguen dejando caer sus recompensas al compartir celda con el agente a evaluar. Un pequeño cambio que, sin embargo, hace inservible parte de la demostración de la Proposición~\ref{pro:SensibleRecompensas} la cual afirma que los entornos son sensibles a las recompensas. En esta proposición se demostraba cómo el agente puede elegir entre obtener una recompensa igual a $0$ u obtener una recompensa distinta de $0$ (positiva o negativa). Concretamente este cambio invalida la demostración de que un agente sea capaz de asegurarse una recompensa igual a $0$ simplemente quedándose quieto en una celda. Ya que Good y Evil siempre dejan caer sus recompensas, al compartir celda con el agente evaluado, éste no podrá asegurarse obtener una recompensa de $0$, ya que consumirá la recompensa que dejen Good o Evil. Debido a este cambio debemos reformular esta parte de la demostración.

\begin{proposition}
La aproximación a la clase de entornos $\Lambda$ implementada es sensible a las recompensas.
\begin{proof} {\bf Sketch.}
El cambio realizado incapacita el comportamiento del agente de quedarse parado en una celda para obtener una recompensa de $0$. Sin embargo, sigue siendo posible obtener esta recompensa si el agente modifica su comportamiento a uno aleatorio, ya que este comportamiento tiene una recompensa esperada igual a $0$ tal y como puede verse en las \mbox{Proposiciones~\ref{pro:Balanceada} y \ref{pro:FuertementeBalanceada}}.

Ya que se sigue demostrando que es posible elegir obtener una recompensa media igual a $0$ y el cambio realizado no invalida el resto de la demostración de la Proposición~\ref{pro:SensibleRecompensas}, la clase de entornos implementada sigue siendo sensible a las recompensas.
\end{proof}
\end{proposition}

\subsection{Generación del espacio}\label{sec:GeneracionEspacio}
Por el momento, no se genera un algoritmo de Markov para describir el espacio. En su lugar, se genera directamente la cadena que lo representa utilizando una combinación de las distribuciones geométrica y uniforme.

Existen dos formas para construir el espacio: (1) se proporciona manualmente su descripción o (2) se deja que la descripción se genere aleatoriamente y se comprueba que el espacio que representa está fuertemente conectado. En ambos casos es necesaria una descripción del espacio para posteriormente construir su topología (celdas y acciones).

A continuación vemos cómo se codifica esta descripción y el procedimiento que se sigue para su generación.

\subsubsection{Codificación del espacio}
Se ha utilizado la codificación para espacios del marco conceptual visto en la Sección~\ref{sec:MCCodYGenEspacios} en la Página~\pageref{sec:MCCodYGenEspacios}, con dos pequeñas modificaciones: (1) Utilizamos el símbolo `|' para separar las celdas. (2) La acción reflexiva será siempre la acción 0. Esta acción, al estar implícita en todos los espacios, se omite en la descripción.

Vamos a ver un ejemplo:

\begin{center}
{\footnotesize \ttfamily 1+2++3|1+23-|1+23|1+2-{}-3-}
\end{center}

Como podemos ver en esta descripción, el espacio se divide en 4 celdas y se pueden realizar 4 acciones (3 acciones más la acción reflexiva 0). Desde la primera celda se llega hasta la segunda celda tras realizar la acción 1, la acción 2 nos lleva a la tercera celda y la acción 3 deja al agente en la misma celda. Si vemos la descripción de la cuarta celda podemos ver como la acción 1 nos mueve a la primera celda (se asume una distribución toroidal), con la acción 2 llegamos a la segunda celda y la acción 3 nos devuelve a la tercera celda.

\begin{figure}[h!]
\centering
\includegraphics[width=0.70\textwidth]{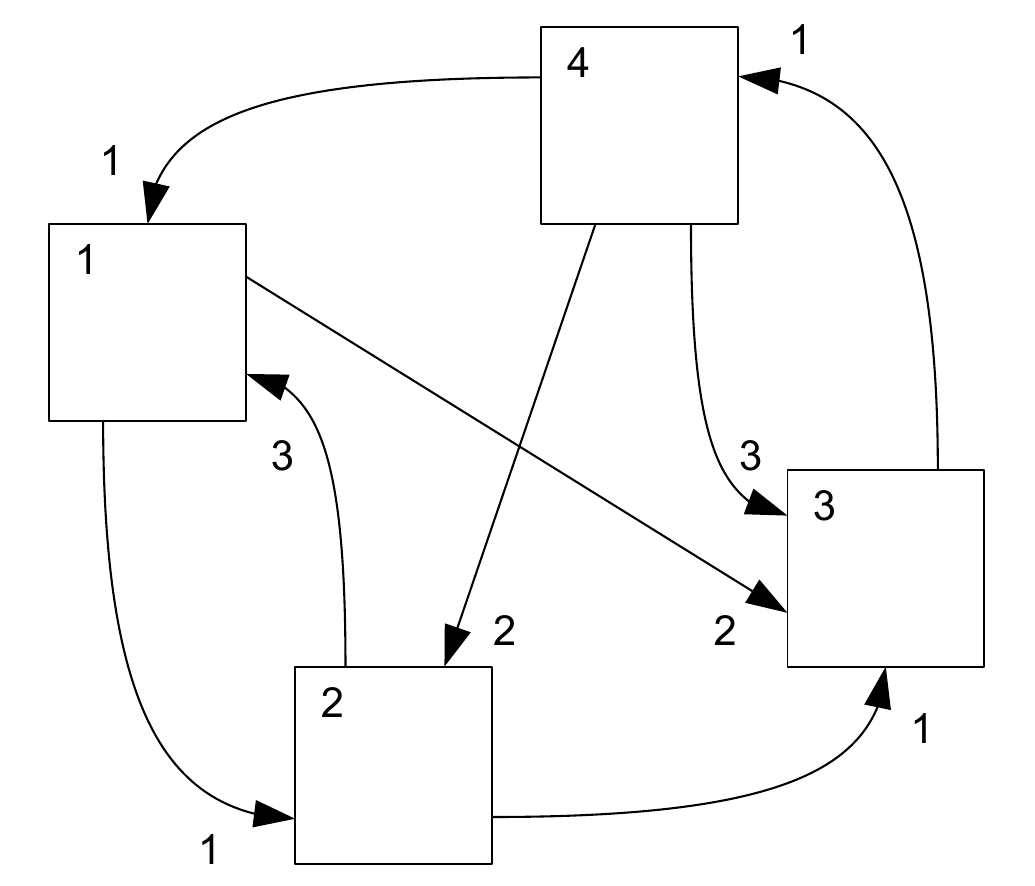}
\caption{Espacio representado por la cadena `{\footnotesize \ttfamily 1+2++3|1+23-|1+23|1+2-{}-3-}' omitiendo la acción reflexiva 0.}
\label{fig:espacio4celdas}
\end{figure}

En la Figura~\ref{fig:espacio4celdas} podemos ver una representación gráfica de este espacio.

\subsubsection{Generación aleatoria de la descripción}
Es posible generar la descripción del espacio aleatoriamente. Hemos utilizado los siguientes pasos para generarla:

\begin{enumerate}
\itemsep=0px
\item Se decide el número de celdas que componen el espacio ($n_c$): Para decidir el número de celdas se ha utilizado una distribución geométrica con $p = 1/2$. Para elegir el número de celdas, asociamos el número de intentos para obtener un éxito dentro la distribución con el número de celdas que tendrá el espacio (empezando a partir de 2 celdas) del siguiente modo: si la distribución tiene éxito en el primer intento el espacio tendrá 2 celdas, si tiene éxito al segundo intento tendrá 3 celdas, etc. Establecemos un número máximo de celda $n_{max}$, esto es útil para cuando desarrollemos el interfaz, que el número de celdas siempre nos pueda caber en pantalla. Normalizamos la distribución de modo que si se llega a la celda $n_{max}$ siempre tenga éxito.
\item Se decide el número de acciones que componen el espacio ($n_a$): De igual modo que con el número de celdas, utilizamos una distribución geométrica para elegir el número de acciones y normalizamos utilizando el número de celdas como el número máximo de acciones.
\item Se eligen las conexiones entre las celdas: Para elegir a qué celdas están conectadas las acciones se sigue el siguiente procedimiento: (1) utilizamos una distribución uniforme para elegir el símbolo ($+$ o $-$) y, (2) de nuevo una distribución uniforme entre 0 (la acción mantiene al agente en la celda) y el número de celdas para elegir el desplazamiento. Insertamos tantos símbolos ($+$ o $-$) tras la acción como indique el desplazamiento para formar una conexión. Este procedimiento se seguirá para todas las acciones en todas las celdas.
\end{enumerate}

\subsection{Observaciones}\label{sec:DifObservaciones}
Según el marco conceptual, a los agentes se les proporcionan las observaciones a través de una tira de caracteres como podemos ver en la Sección~\ref{sec:observaciones}. Sin embargo, para esta implementación, se optó por realizar una copia de la estructura del espacio (incluyendo los objetos situados en cada celda) y enviarles esta misma copia a todos los agentes como su observación del entorno. De este modo, todos los agentes tienen una misma visión completamente observable del entorno.\footnote{Aunque los agentes puedan ver toda la información del espacio (incluidas las recompensas), solo hacen uso de la información a la que deberían tener acceso como propone la definición de las observaciones descrita en el marco conceptual.}

\subsection{Parametrización del entorno $\Lambda$}
Además de estos cambios, en esta tesis de máster se han añadido varias características nuevas que permiten parametrizar algunos aspectos del sistema de evaluación, proporcionándole una mayor flexibilidad. Entre estas características están:

\begin{itemize}
\itemsep=0px
\item La compartición o no de la recompensa de la celda si varios agentes se encuentran en ella, por lo que si dos agentes se encuentran en una celda con una recompensa de 0,5, entonces se les repartirá 0,5/2 = 0,25 a cada uno de ellos.
\item El momento en el que se consumen las recompensas (siendo antes o después de actualizar las recompensas de las celdas), intuitivamente podemos entender este parámetro como la facilidad de obtener íntegras las recompensas que dejan caer Good y Evil (esto es, sin que el entorno las haya modificado). Si se establece que se consumen antes, con seguir un paso por detrás a Good o a Evil obtendremos su recompensa íntegra, mientras que si se consumen después, solo se conseguirán obtener cuando coincidamos con ellos en la misma celda.
\item La desaparición o no de las recompensas de las celdas una vez que los agentes las han consumido.
\item El número por el que se dividen las recompensas (factor de división) que se encuentren en el entorno en cada iteración\footnote{Este factor permite un amplio abanico de posibilidades al actualizar las recompensas: un factor de 1 implica que las recompensas se mantendrán intactas entre iteraciones, con un factor de 2 las recompensas se dividen a la mitad, un factor de $\infty$ eliminará las recompensas en cada interacción, \dots}.
\end{itemize}

Con la inserción de estas características, los entornos disponen de una mayor flexibilidad dependiendo de los parámetros que se seleccionen para cada una de sus características. Podemos apreciar parte de esta flexibilidad en el Ejemplo~\ref{ejm:FlexibilidadNuevosParametros}.

\begin{example}
Imaginemos un entorno con 5 agentes, en donde las recompensas no se comparten entre los agentes (todos obtienen la recompensa de la celda en la que se encuentren sin dividirla entre el resto de agentes que se encuentren en esa misma celda), se consumen antes de ser actualizadas (por lo que con seguir un paso por detrás a Good se conseguirá su recompensa íntegra), ni desaparecen tras ser consumidas por los agentes y con un factor de división igual a 2. Con esta configuración, Good y Evil van dejando un rastro de recompensas, las cuales, conforme más tiempo transcurra desde que fueron dejadas, se van haciendo más y más pequeñas ($1$, $1/2$, $1/4$, $1/8$, \dots).

En este entorno, todos los agentes pueden simplemente seguir a Good (conforme le vayan encontrando) como un comportamiento óptimo, de modo que todos obtengan la recompensa íntegra ($1$). Además, el rastro de recompensas no desaparecerá tras ser consumidas por los agentes, por lo que aun resultará más fácil averiguar por donde se encuentran Good y Evil.

Si modificamos este entorno de modo que se empiecen a compartir las recompensas entre los agentes, entonces, en la situación anterior, puesto que Good deja caer una recompensa de 1 y hay 5 agentes en la celda, cada uno obtendrá una recompensa igual a 1/5. Sin embargo, ya que las recompensas no desaparecen tras ser consumidas, las anteriores celdas por las que pasó Good seguirán con las recompensas 1/2, 1/4, 1/8, \dots (nótese que la compartición de las recompensas no modifica la recompensa de la celda, únicamente establece qué parte de la recompensa se lleva cada agente), por lo que existirán mejores recompensas que la que actualmente obtienen los agentes.

Este pequeño cambio hace que los agentes tengan que buscar una posición que les otorgue una mejor recompensa. Sin embargo, la recompensa que recibirán los agentes en cada celda va variando conforme se van modificando las posiciones de todos los agentes (ya que se comparten las recompensas en función del número de agentes), por lo que deberán competir para conseguir las mejores posiciones (y por lo tanto recompensas) posibles.

Si encima se modificase el factor de división de 2 a 3, entonces el rastro de recompensas sería: $1$, $1/3$, $1/9$, $1/27$ \dots, por lo que quedarían menos celdas que proporcionen recompensas mejores a $1/5$. En este caso, los agentes deberán competir con el resto de agentes en un conjunto de celdas aun más reducido si pretenden obtener buenas recompensas.
\label{ejm:FlexibilidadNuevosParametros}
\end{example}

\section{Arquitectura del sistema de evaluación}
La estructura del test está constituida por una serie de ejercicios para cada test. Cada uno de estos ejercicios se realiza en un entorno generado a partir de la clase de entornos, en donde se evaluarán a los agentes. Además, en los entornos generados por esta clase de entornos (más concretamente en sus espacios) podrán añadirse cualquier número de objetos (y de agentes).

Este sistema de evaluación está divido en dos partes: (1) una interfaz gráfica con la que el usuario (humano) podrá interactuar con el sistema y (2) la estructura interna que se ha diseñado para constituir el propio test (clase de entornos, ejercicios, espacio, objetos, \dots).

%Este sistema aun está en construcción y desarrollo, por lo que aun no se han implementado todas las características que debe poseer. Por el momento, debido a su temprano desarrollo, el sistema no está construido para adaptarse directamente al sujeto que pretende utilizarlo, por lo que cada vez que se quiera utilizar el sistema, deberá configurarse anteriormente para el tipo de sujeto que lo quiera utilizar.

Ya que el sistema sigue en desarrollo, la arquitectura del test aun no ha sido implementada por completo. Sin embargo, si que se ha diseñado una implementación preliminar con la que poder comenzar a realizar evaluaciones, a la vez que es posible seguir con su construcción. En concreto se ha implementado gran parte de la arquitectura necesaria para realizar un ejercicio (clase de entornos, espacio, objetos, agentes, test en el que se está realizando, \dots), pero sin llegar a construir definitivamente las transiciones entre ejercicios, por lo que, si quisiera realizarse un test, deberán seleccionarse a mano las características de los ejercicios que se deseen utilizar.

En el Apéndice~\ref{apx:DiagramaDeClases} podemos ver el diagrama de clases del diseño e implementación del test y una explicación de sus partes más importantes.

\section{Realización de un test}
Un test es un conjunto de ejercicios y cada ejercicio está representado por un entorno.

Por el momento no generamos los entornos siguiendo ninguna distribución adaptativa ni tampoco no adaptativa. Los entornos deberían generarse en función de la complejidad de los entornos anteriores y del resultado que hubiera obtenido el agente en ellos. Sin embargo, en esta aproximación, cada entorno se genera a partir de las propiedades que queremos que tenga (número de celdas, número de acciones, \dots).

Se han implementado dos métodos con los que determinar la duración que dispondrán los agentes para interactuar en cada entorno: (1) un número máximo de interacciones o (2) un tiempo máximo para interactuar con el entorno. Transcurrido este periodo las interacciones con el entorno habrán terminado.

\section{Realización de un ejercicio}
En esta sección vemos en detalle como se realiza un ejercicio con el sistema de evaluación desarrollado.

Antes de comenzar, debe prepararse el entorno donde se realizará el ejercicio: el espacio y los objetos que contiene.

\begin{enumerate}
\itemsep=0px
\item Se genera el espacio siguiendo los pasos descritos en la Sección~\ref{sec:GeneracionEspacio}.
\item Se generan todos los objetos (incluyendo a Good y Evil) y su comportamiento. Si su comportamiento sigue un patrón, se genera en función de un conjunto de acciones (normalmente las posibles en el entorno) y un parámetro de parada $p_{stop}$ siguiendo estos pasos: (1) Se generará una acción aleatoria (de entre el conjunto de acciones) siguiendo una distribución uniforme. (2) Se comprueba si una distribución geométrica con $p = p_{stop}$ tiene éxito para parar de generar el patrón. De lo contrario se vuelve al paso (1).
\item Se colocan aleatoriamente todos los objetos en el espacio (incluyendo al agente evaluado), siguiendo una distribución uniforme entre 1 y el número de celdas $n_c$. Si Good y Evil se colocan en la misma celda, se recalculan dos nuevas celdas para ellos mientras sigan coincidiendo en la misma celda.
\item Se establecen las recompensas de las celdas a 0.
\end{enumerate}

Tras preparar el entorno, comienza el bucle principal para las interacciones con los agentes:

\begin{enumerate}
\item Recolocar a los agentes Good y Evil si ha transcurrido el número de interacciones de la cláusula ciclo (para entender el motivo de esta cláusula véase el Ejemplo~\ref{ex:motivo_ciclos} y la solución propuesta a continuación).
\item Los objetos Good y Evil dejan caer sus respectivas recompensas en sus celdas. Good deja caer una recompensa positiva mientras que Evil deja caer una recompensa negativa.
\item Se copia la estructura interna del espacio (con los objetos que se encuentran dentro) para proporcionarla como observación.

\item Para cada agente (incluyendo al agente evaluado):
\begin{itemize}
\item Se le entrega la recompensa de la interacción anterior. Si fuera la primera interacción se entrega una recompensa inicial igual a 0.
\item Se le entrega la observación (la copia de la estructura del espacio).
\item El agente indica la acción que pretende realizar siguiendo su propio comportamiento.
\end{itemize}

\item Se realizan simultáneamente las acciones de todos los agentes. Si Good y Evil pretenden colocarse en la misma celda se procede del siguiente modo:
\begin{itemize}
\item Si Good no se movió en esta interacción, el agente Evil fue el que intentó colocarse en la celda en la que se encontraba Good, por lo que Evil no se moverá.
\item Análogamente a la situación anterior, si Evil no se movió en esta interacción entonces Good no se moverá.
\item En cualquier otro caso, tanto Good como Evil se movieron a la misma celda sin que ninguno se encontrarse previamente en ella. En este caso se decide aleatoriamente cual de los dos objetos no se moverá en esta interacción.
\end{itemize}

\item Se calculan las recompensas para todos los agentes y se almacenan para ser entregadas en la siguiente interacción.
\item Se actualizan todas las recompensas de las celdas.
\item Si el ejercicio aún no ha terminado, seguir con el bucle.
\end{enumerate}

Una vez se ha terminado el ejercicio, se envían a los agentes las recompensas que quedaron pendientes de la última interacción del bucle y se calcula la media de las recompensas obtenidas del agente evaluado durante el ejercicio. Esta media representa el resultado (o la inteligencia demostrada) del agente en el ejercicio.

\section{Modos de interacción con el sistema de evaluación}
Los modos de interacción son los que permiten interactuar a los sujetos (humanos, algoritmos, \dots) con el test. Estos modos pueden estar diseñados como interfaces gráficas o simplemente como una interfaz (no gráfica) dependiendo de la naturaleza del sujeto para el que se ha construido.

En el trabajo previo se diseñaron dos de estos modos: Uno de ellos se diseñó como una interfaz gráfica para ayudar a los autores con el desarrollo del propio test (Modo investigación). El otro fue diseñado como interfaz (no gráfica) para permitir la interacción de los agentes de algoritmos software de IA (Modo algoritmo de IA). En esta tesis hemos añadido (como interfaz gráfica) un nuevo modo para que agentes humanos puedan interactuar con el sistema (Modo ser humano). A continuación detallamos los tres modos.

\subsection{Modo investigación}
Este modo se ha diseñado explícitamente para asistir durante el desarrollo y la implementación de los tests del sistema de evaluación. Por lo tanto, para poder probar mejor el sistema, permite generar cualquier tipo de ejercicio modificando cualquier parámetro del entorno y de los agentes implementados. Además, con este modo se proporciona una amplia interfaz con la que se puede ver completamente el estado del entorno.

\begin{figure}[h!]
\centering
\includegraphics[width=1\textwidth]{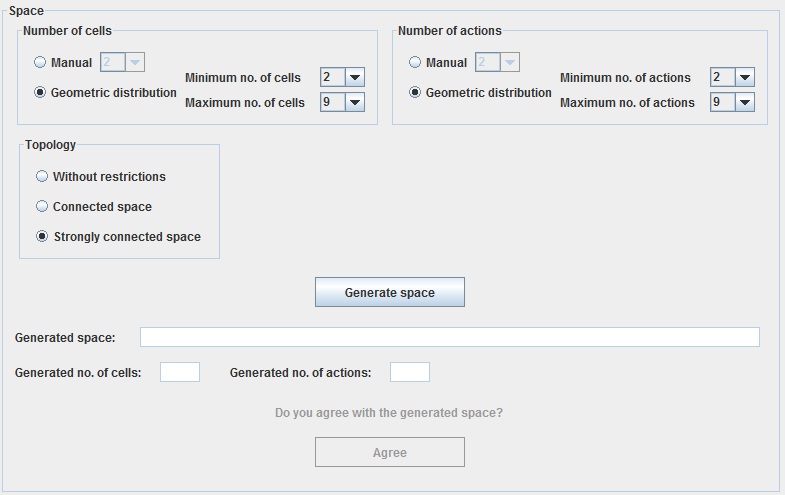}
\caption{Interfaz para modificar las características del espacio en el modo investigación.}
\label{fig:OpcionesEspacio}
\end{figure}

\begin{figure}[h!]
\centering
\includegraphics[width=1\textwidth]{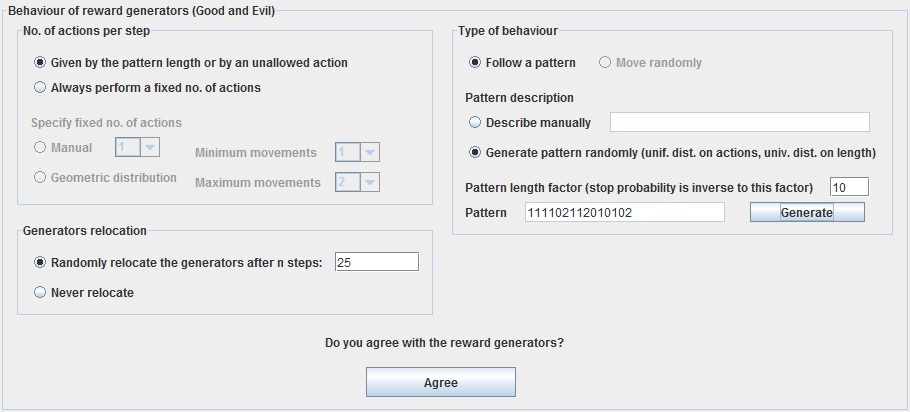}
\caption{Interfaz para modificar el comportamiento de $Good$ y $Evil$ en el modo investigación.}
\label{fig:OpcionesGoodEvil}
\end{figure}

A continuación vemos en las \mbox{Figuras \ref{fig:OpcionesEspacio} y \ref{fig:OpcionesGoodEvil}} las interfaces que permiten generar el espacio y el comportamiento para Good y Evil que se utilizarán durante la prueba.

\begin{figure}[h!]
\centering
\includegraphics[width=1\textwidth]{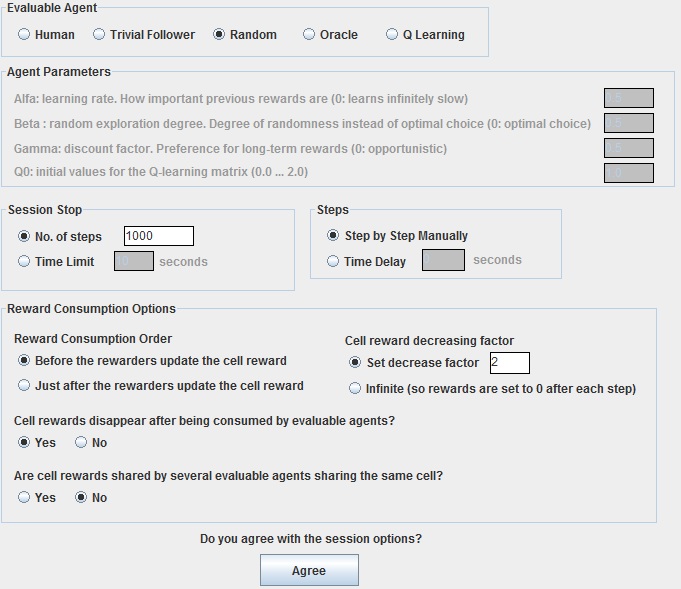}
\caption{Interfaz para modificar el ejercicio en el modo investigación.}
\label{fig:OpcionesSesion}
\end{figure}

En la Figura~\ref{fig:OpcionesSesion} se puede seleccionar (y configurar si es posible) al agente que se quiere probar y se puede especificar las características que poseerá el entorno para el ejercicio.

\begin{figure}[th!]
\centering
\includegraphics[width=1\textwidth]{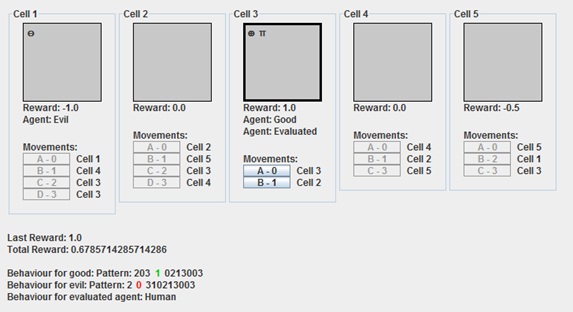}
\caption{Visión completa del entorno en el modo investigación.}
\label{fig:InterfazEntornoInvestigacion}
\end{figure}

En la Figura~\ref{fig:InterfazEntornoInvestigacion} podemos ver la interfaz que se muestra durante la prueba. Esta interfaz está formada por todos los detalles del entorno actual, desde la observación (incluyendo incluso las recompensas) hasta el comportamiento de los agentes Good y Evil y su situación actual.

\subsection{Modo algoritmo de IA}\label{sec:ModoAlgoritmoIA}
Ya que este modo se ha diseñado para algoritmos software, no posee ninguna interfaz visual de usuario. El test interactúa con los algoritmos mediante las observaciones y las recompensas. Las observaciones, cómo se indicó en la Sección~\ref{sec:DifObservaciones} en la Página~\pageref{sec:DifObservaciones}, constan de una copia del espacio con todos sus atributos, a partir del cual los algoritmos recogen la información que necesitan. Sin embargo, esta implementación está preparada (siguiendo la teoría del marco conceptual) para proporcionar la observación utilizando cadenas. Se les proporcionará a los agentes las recompensas utilizando números reales entre $[-1, 1]$.

\subsection{Modo ser humano}\label{sec:ModoHumano}
En esta tesis de máster hemos diseñado e implementado una primera interfaz para evaluar seres humanos. Esta interfaz se ha desarrollado con el objetivo en mente de que fuera lo más apropiada (a la vez que intuitiva) posible.

\begin{figure}[h!]
\centering
\includegraphics[width=1\textwidth]{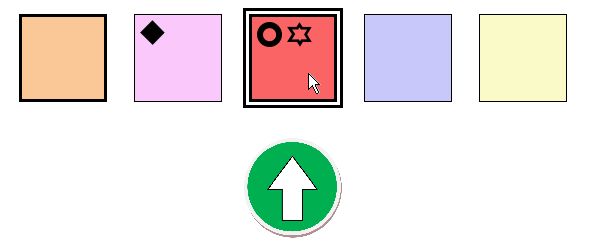}
\caption{Una captura de la interfaz para humanos. Podemos ver una situación donde el agente acaba de recibir una recompensa positiva, mostrado a través del círculo verde con una flecha hacia arriba. La imagen también muestra que el agente está situado en la celda 3, y Evil y Good se encuentran en las celdas 2 y 3 respectivamente. El agente se puede mantener en la celda 3 y también puede moverse a la celda 1. La celda 3 está destacada ya que es una celda accesible y el cursor se encuentra encima de ella.}
\label{fig:InterfazHumanos}
\end{figure}

La interfaz para humanos se ha diseñado con los siguientes principios en mente: (1) los símbolos utilizados para representar las observaciones no tienen que tener ningún significado explícito para el sujeto que va a realizar el test para evitar sesgos a su favor (p.~ej. no utilizar una calavera para el agente Evil) y (2) las acciones y las recompensas deberían ser fácilmente interpretables por el sujeto para evitar un mayor esfuerzo cognitivo. Podemos ver esta interfaz en la Figura~\ref{fig:InterfazHumanos}.

Las celdas se representan con cuadrados coloreados. Los agentes se representan con símbolos que tratan de ser lo más `neutrales' posibles (p.~ej. $\blacklozenge$ para $Evil$ y $\bigvarstar$ para $Good$ y $\bigcirc$ representando al sujeto\footnote{Otra posibilidad hubiera sido utilizar símbolos lingüísticos para representar a los agentes. Esta aproximación podría haber desigualado las condiciones del test a favor de los que conocieran estos símbolos, fallando al tratar de utilizar una interfaz lo menos sesgada posible. Por este motivo decidimos utilizar símbolos lo más neutros posibles.}. Las celdas accesibles tienen un borde más grueso que las no accesibles. Cuando el sujeto sitúa el puntero del ratón encima de una celda accesible, esta celda se destaca utilizando un borde doble e incrementando la saturación de los colores de fondo. Las recompensas positivas, neutrales y negativas se representan respectivamente con una flecha hacia arriba en un círculo verde, un cuadrado pequeño en un círculo gris y una flecha hacia abajo en un círculo rojo. El motivo de no utilizar los típicos símbolos representativos para el acierto $\checkmark$ y fallo $X$, es que las recompensas no se corresponden directamente con aciertos o fallos, ya que, por ejemplo, en una situación dada, una acción con mala recompensa podría ser la mejor acción posible al acercarse más a una zona en donde posteriormente se otorgarán recompensas positivas.

En esta interfaz no incluimos información sobre la recompensa acumulada que posee el examinado. El hecho de saber cómo va uno en la prueba, además de recibir información inmediata sobre el último ensayo, podría influir en cierta medida en la motivación o en la disposición con la que se enfrenta al siguiente ejercicio. Esto podría afectar de forma diferencial en función de cada persona, como por ejemplo en función de su ``personalidad'', y no directamente en la inteligencia. Por ejemplo, el hecho de saber que se ha fallado muchas veces podría bloquear a algunos y generar aun peores resultados, mientras que a otros podría motivarlos para conseguir recuperarse. También es posible que, para algunos, conseguir una recompensa acumulada muy elevada reduzca su motivación en la tarea.

\section{Agente de IA de referencia: Q-learning}\label{sec:QlearningImplementado}
Ya que este sistema de evaluación se ha construido utilizando la configuración de aprendizaje por refuerzo (RL), lo natural es comenzar evaluando algoritmos de RL. De hecho, como vimos en el Capítulo~\ref{cap:Precedentes} RL es una configuración lo suficientemente apropiada y general como para definir y analizar agentes de aprendizaje, los cuales interactúan con un entorno a través del uso de observaciones, acciones y recompensas.

Debido a la vasta cantidad de literatura que existe en algoritmos de RL, nos encontramos con un difícil problema al intentar obtener un algoritmo `representativo' al que empezar a evaluar. De acuerdo con \cite{Woergoetter:2008}, los tres algoritmos más influyentes en RL son Temporal Difference (TD) learning \cite{sutton1988learning}, Actor-Critics adaptable \cite{barto1983neuronlike} y Q-learning \cite{watkins1992q}.

Para probar la clase de entornos, hemos elegido a Q-learning como representante de los algoritmos de RL para evaluarlo. La razón de elegir Q-learning ha sido deliberada, ya que primero queríamos evaluar un algoritmo que fuera estándar y, más especialmente, porque (por el momento) no queremos evaluar algoritmos muy especializados en entornos ergódicos o algoritmos con mejores propiedades computacionales (p.~ej. delayed Q-learning \cite{Strehl:2006:PMR:1143844.1143955} hubiera sido una mejor elección si hubiéramos querido evaluar también la velocidad). Hemos utilizado una implementación estándar de Q-learning, tal y como se explica en \cite{watkins1992q} y \cite{sutton1998reinforcement}. Véase la Sección~\ref{sec:DefinicionQlearning} para una explicación más detallada de como funciona esta técnica.

Básicamente, Q-learning es una función de agente que decide, a partir del estado actual, cual es la acción que debe realizar para conseguir la mayor recompensa posible. Sin embargo, las observaciones que le proporciona el entorno en la aproximación implementada, resultan inapropiadas para representar los estados, por lo que hemos tenido que especificar una descripción de los estados más clara y concisa. Podemos ver un ejemplo de esta descripción en el Ejemplo~\ref{ejm:DescripciónEstadoQLearning}.

\begin{example}\label{ejm:DescripciónEstadoQLearning}
Un ejemplo de una descripción para un estado en la matríz Q es la cadena `$101|000|010$'.

Cada secuencia de $0$ y $1$ entre $|$ representa el contenido de una celda, por lo que esta descripción representa a un espacio con tres celdas. Dentro de cada celda, los números $0$ y $1$ representan la existencia o no de los objetos dentro de esa celda (con las posiciones de los objetos constantes entre celdas) del siguiente modo: siendo `$101$' el contenido de la primera celda, significa que `$1$' el primer objeto se encuentra dentro de la celda, `$0$' el segundo objeto no se encuentra en la celda y `$1$' el tercer objeto se encuentra en la celda. Así tenemos que tanto el primer como el tercero se encuentran dentro de la primera celda. Podemos ver que la siguiente celda (con descripción `$000$') no contiene ningún objeto. Finalmente, en la última celda (con descripción `$010$') se encuentra el segundo objeto.

Para seguir un orden, hemos establecido que el primer objeto siempre será el agente Good, el segundo objeto será el agente Evil y el tercer objeto será el agente que se está evaluando $\pi$ (siendo en este caso el propio Q-learning)\footnote{Nótese que aunque hayamos establecido este orden a priori, Q-learning no se aprovecha de esta información para adivinar qué posición ocupa cada agente. Esta descripción se ha utilizado únicamente como representación del estado para la matriz Q sin influenciar directamente en el comportamiento del agente.}.
\end{example}

\chapter{Evaluación de un algoritmo de IA}\label{cap:Evaluacion}
En este capítulo vamos a probar si la clase de entornos diseñada ofrece resultados coherentes al evaluar algoritmos de IA. En concreto evaluamos la técnica de aprendizaje por refuerzo conocida como Q-learning visto en el Capítulo~\ref{cap:Precedentes} e implementado en la herramienta de evaluación, tal como hemos explicado en el capítulo anterior.

La evaluación de Q-learning en este test de inteligencia general nos proporciona algunos puntos de vista interesantes sobre cómo podrían evaluarse los algoritmos de RL (con un test de inteligencia) y también sobre la viabilidad de estos tests como test de inteligencia general para la IA.

\section{Configuración de los experimentos}
Para estos experimentos vamos a evaluar Q-learning a través de un conjunto de entornos. Con respecto a las características que tendrán estos entornos, de las opciones descritas en capítulos anteriores, para este experimento hemos tomado algunas decisiones:

\begin{itemize}
\item A pesar de que \cite{HernandezOrallo09b} sugiere que las interfaces sean parcialmente observables, aquí hemos hecho que sean completamente observables, de modo que los agentes serán capaces de ver todas las celdas con su contenido y las acciones disponibles desde la celda en la que se encuentren y a qué celdas llevará cada acción.
\item Los agentes Good y Evil siguen un mismo patrón de comportamiento en cada ejercicio. Este patrón se recalcula para cada nuevo ejercicio.
\item Las recompensas de las celdas se consumen \emph{después} de actualizarlas y se mantienen solo durante una iteración, pasada la iteración se establece a 0. Esta configuración fuerza a que los agentes solamente obtengan recompensas cuando coincidan con Good o Evil en una celda. Nótese que, en esta configuración, los agentes deben adivinar cuál será la siguiente celda en la que se colocará Good para poder obtener buenos resultados.
\item Ya que en estos experimentos no queremos evaluar características especializadas (como, por ejemplo, la velocidad), cada ejercicio tiene un número máximo de interacciones durante las que los agentes interactúan con el entorno. Hemos establecido un número de interacciones lo suficientemente grande como para poder apreciar a qué resultados convergen los agentes.
\label{itm:CaracteristicasEntornos}
\end{itemize}

\section{Agentes evaluados}
Q-learning tiene dos parámetros clásicos que debemos establecer: \emph{learning rate} ($\alpha$) y \emph{discount factor} ($\gamma$). Los parámetros para Q-learning son $\alpha = 0.05$, $\gamma = 0.35$. Los elementos en la matriz $Q$ se establecen inicialmente a $2$ (las recompensas tienen un rango entre $-1$ y $1$, pero están normalizados entre $0$ y $2$ para que la matriz $Q$ sea siempre positiva). Los parámetros se han elegido para ser óptimos en este conjunto de experimentos probando 20 valores consecutivos para $\alpha$ y $\gamma$ entre $0$ y $1$. Estas $20$ x $20$ = $400$ combinaciones se han evaluado en 1.000 sesiones cada una utilizando entornos aleatorios de entre los tipos de entornos utilizados en este capítulo, por lo que estos parámetros han sido escogidos de modo que Q-learning obtenga los mejores resultados posibles en estos experimentos.

El estado de la matríz Q se describe a partir del contenido de las celdas (incluyendo la localización de los agentes) de la copia del espacio proporcionado en las observaciones. En la Sección~\ref{sec:QlearningImplementado} podemos ver cómo se describen los estados.

Para tener unos resultados de referencia, hemos implementado varios agentes distintos contra los que comparar los resultados de Q-learning. Estos agentes seguirán siempre el mismo comportamiento sin importar las recompensas que obtengan, por lo que, en principio, sus resultados deberán converger rápidamente. Veamos estos agentes:

\begin{itemize}
\item Aleatorio: Un agente aleatorio es, simplemente, un agente que elige de forma aleatoria de entre sus acciones disponibles utilizando una distribución uniforme.
\item Seguidor trivial: Este agente observa las celdas accesibles para ver si Good está en alguna de ellas. Si lo encuentra, entonces se mueve a esa celda (o se mantiene en la misma celda si Good compartía celda con él). De lo contrario, elige aleatoriamente de entre las celdas accesibles pero tratando de evitar colocarse en la celda en donde se encuentre Evil.
\item Oráculo: Este agente puede ver de antemano las celdas a donde Good y Evil pretenden moverse en la siguiente interacción y, si Good pretende moverse a una celda directamente accesible por el agente, entonces su movimiento será moverse a esa celda. De lo contrario, buscará de entre todas sus celdas accesibles cuál de todas ellas le ofrecerá (en la siguiente interacción) la mayor recompensa\footnote{Nótese que los demás agentes no pueden ver las recompensas que se encuentran dispersas por el entorno, ya que no forman parte de las observaciones.}. Aunque el Oráculo tenga esta `ventaja' sobre el resto de agentes, no es un agente óptimo, ya que solo actúa en función del siguiente movimiento, lo cual no siempre resulta una buena política. Además, debido a que el entorno es estocástico, no es capaz de predecir con precisión en qué celda acabará Good cuando coincida con Evil en la misma celda y, por lo tanto, es posible que pierda su rastro en estas situaciones.
\end{itemize}

\section{Un ejemplo simple}
\begin{figure}[h!]
\centering
\includegraphics[width=0.7\textwidth]{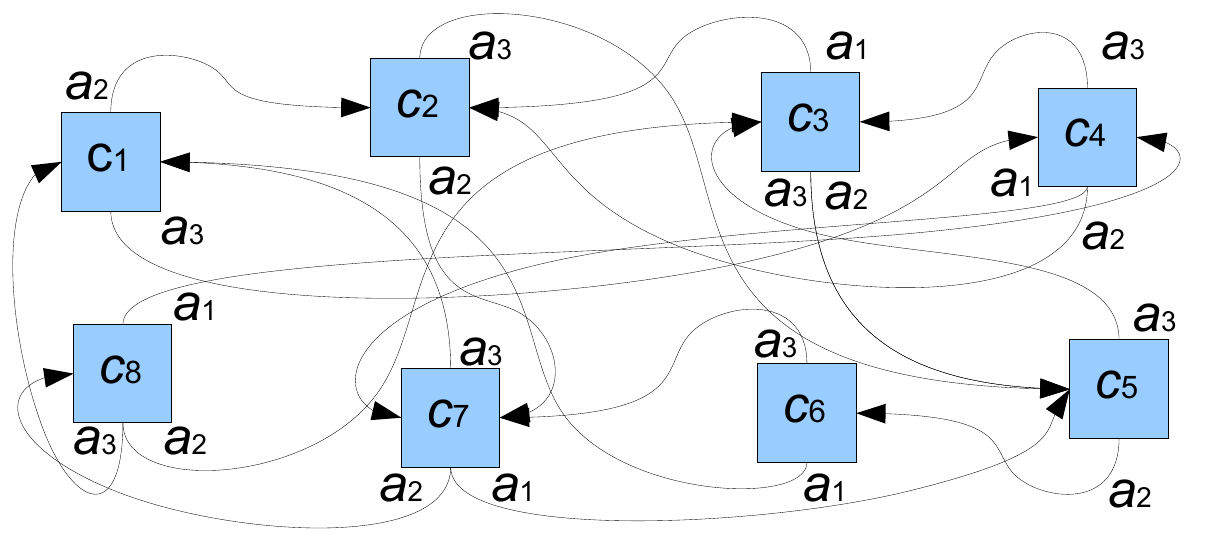}
\caption{Un espacio con 8 celdas y 4 acciones ($a_0, a_1, a_2, a_3$). No se muestra la acción reflexiva $a_0$.}
\label{fig:Espacio}
\end{figure}

Primero veamos los resultados de la evaluación de los agentes previos en un único entorno. El espacio es el de la Figura~\ref{fig:Espacio} y utilizamos el siguiente patrón de movimientos para Good y Evil: $203210200$. Se ha establecido el número de interacciones a 10.000.

\begin{figure}[h!]
\centering
\includegraphics[width=0.5\textwidth]{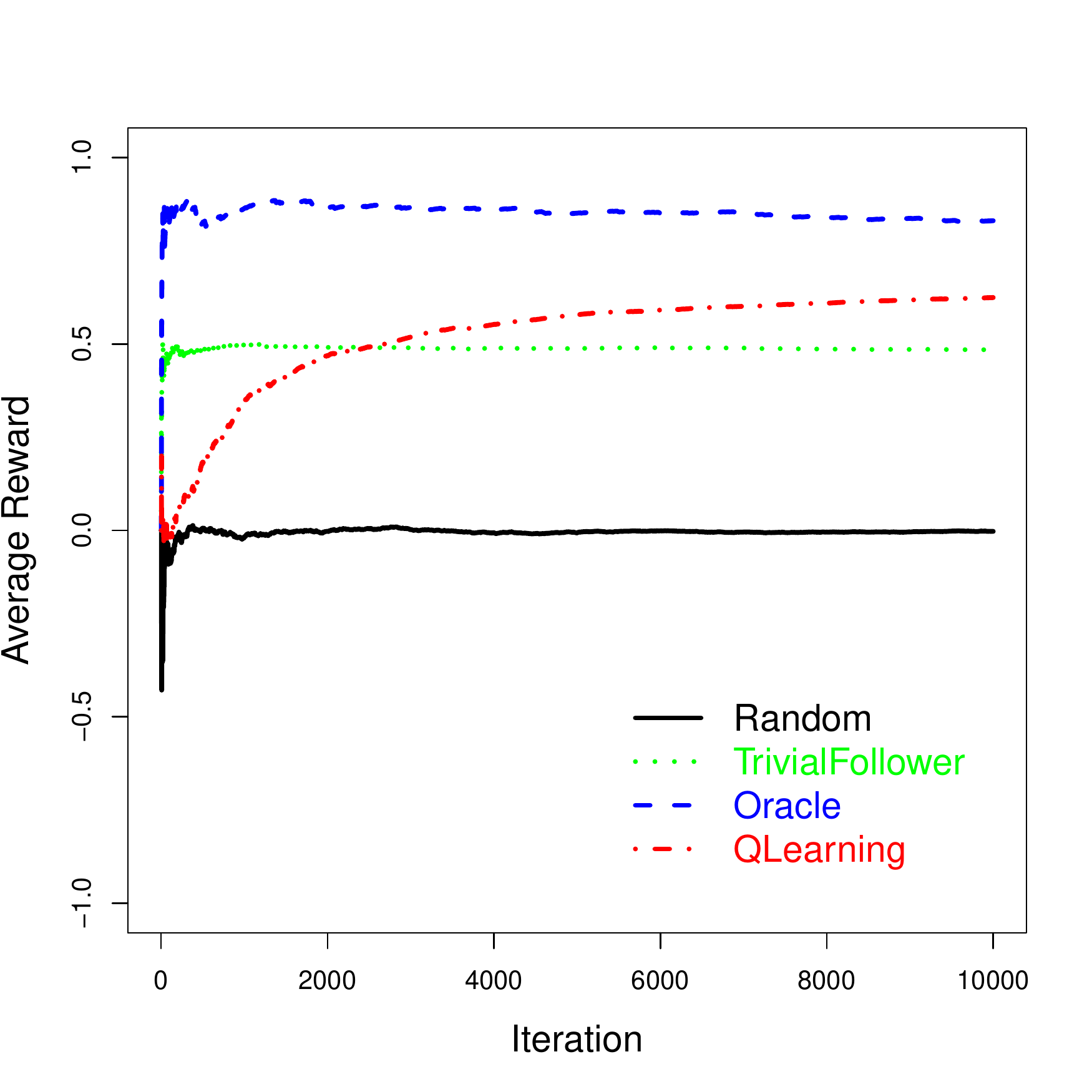}
\caption{Resultados para un entorno simple en el espacio de 8 celdas de la Figura~\ref{fig:Espacio} y con la secuencia de acciones que siguen Good y Evil: $203210200$.}
\label{fig:Simple}
\end{figure}

En la Figura~\ref{fig:Simple} podemos ver la media de las recompensas tras cada interacción para los distintos tipos de agentes (cada agente ha sido evaluado individualmente, pero vemos todos los resultados en el mismo gráfico). Esta figura muestra que la recompensa media inicialmente tiene muchas fluctuaciones para los cuatro tipos de agentes. El agente aleatorio pronto converge a su recompensa media esperada, la cual es 0. El seguidor trivial solo es capaz de obtener una puntuación cercana a $0.5$. En este entorno y con este patrón, seguir a Good es una buena política, pero solo hasta cierto punto. El oráculo converge a un valor alrededor de $0.83$. Como se menciona anteriormente, el oráculo es casi óptimo, y en muchos entornos no alcanzará el valor de $1$, pero si un buen valor. Q-learning resulta que converge de forma muy lenta hacia un valor alrededor de $0.625$. Aunque muy lentamente, en este caso podemos ver como al final supera al seguidor trivial.

Sin embargo, estos resultados solo se han dado para un entorno en particular, por lo que no podemos obtener ninguna conclusión fiable. En las siguientes secciones realizamos una batería de experimentos que intentan obtener algunas conclusiones sobre el comportamiento general de estos agentes.

\section{Experimentos sobre entornos generados}
\label{sec:ExperimentosEntornosGenerados}
En esta sección mantenemos la configuración y los parámetros anteriormente descritos, pero ahora promediamos los resultados entre varios entornos generados aleatoriamente. En particular hemos elegido 100 entornos de 3 celdas, 100 entornos de 6 celdas y 100 entornos de 9 celdas, los cuales nos permiten interpolar los resultados para un rango de celdas entre 3 y 9. Cada entorno tiene una generación aleatoria de la topología del espacio y una probabilidad aleatoria de parada $p_{stop}$ (utilizando una distribución geométrica) que controla el tamaño del patrón de Good y Evil, el cual se establece a $1/100$.

\begin{figure}[h!]
\centering
\includegraphics[width=0.49\textwidth]{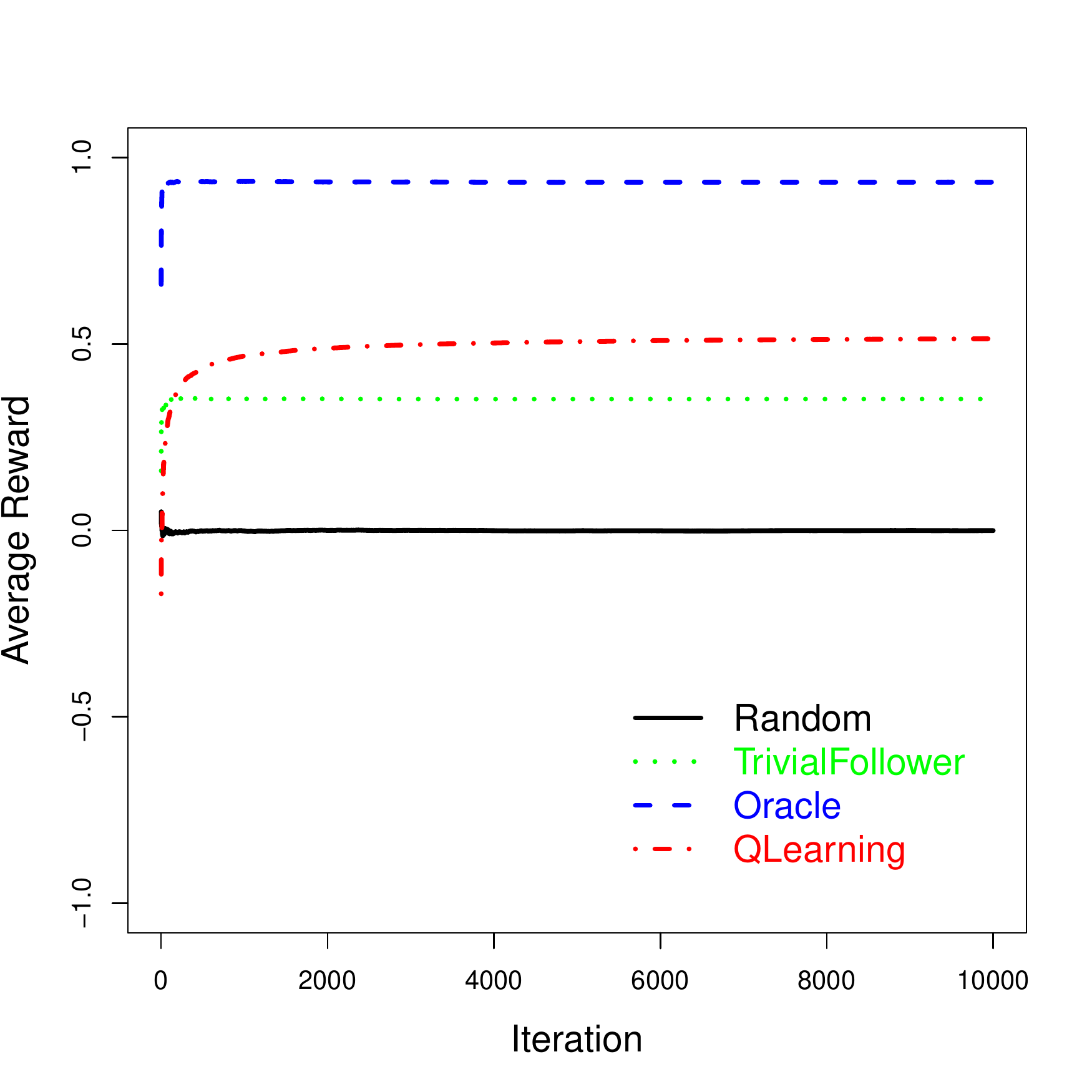}
\includegraphics[width=0.49\textwidth]{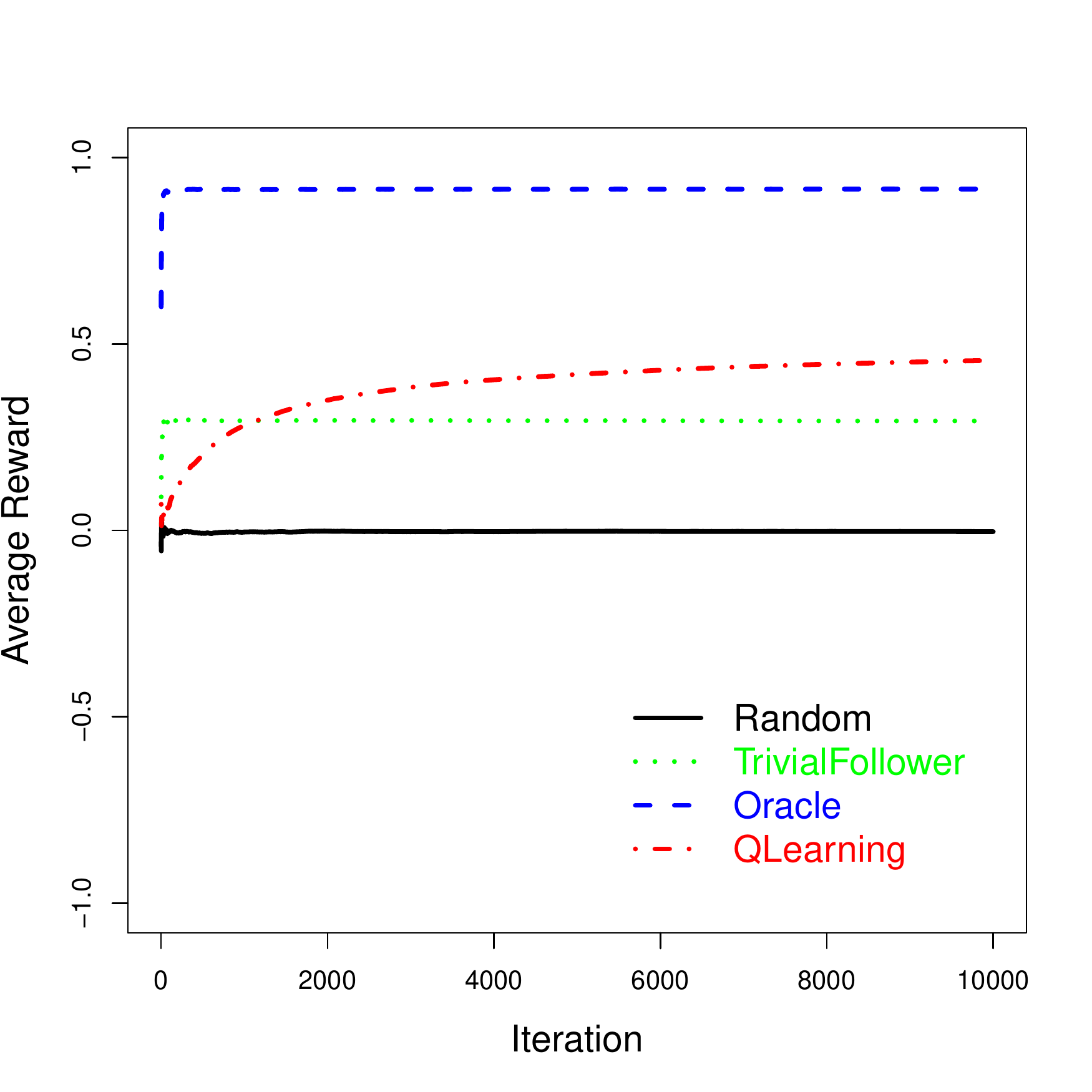}
\includegraphics[width=0.49\textwidth]{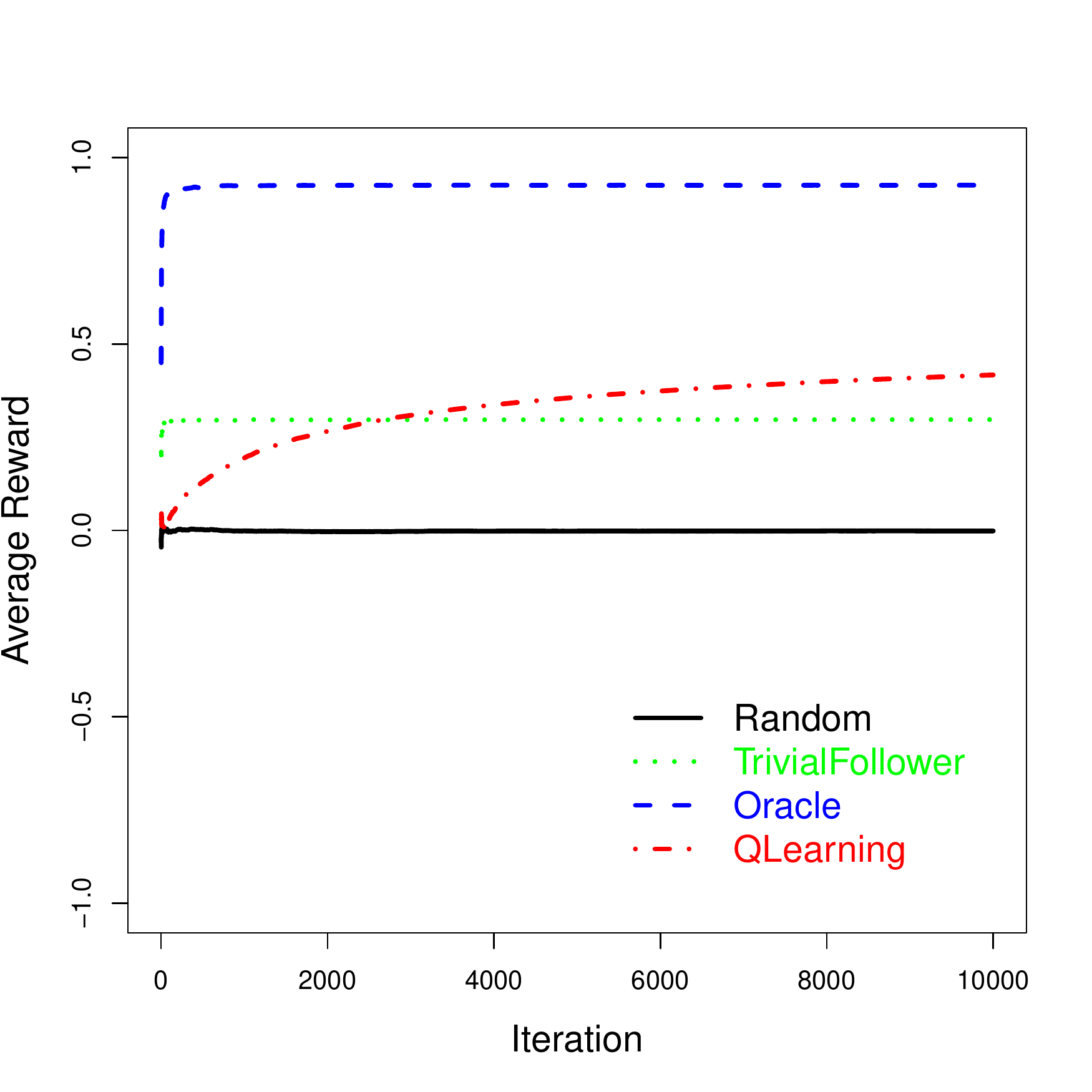}
\caption{Resultados para 100 entornos utilizando la configuración inicial.
Arriba-Izquierda: 3 celdas. Arriba-Derecha: 6 celdas. Abajo: 9 celdas.}
\label{fig:ResultadosEntornosGenerados}
\end{figure}

Se pueden ver los resultados en la Figura~\ref{fig:ResultadosEntornosGenerados}. A partir de estas gráficas obtenemos una idea general, la cual es consistente con el ejemplo simple visto en la sección anterior. Ahora no vemos fluctuaciones, ya que estos son resultados promediados a partir de cada uno de los 100 experimentos. Podemos ver que el tamaño del espacio no afecta mucho en esta configuración, ya que, una vez se ha establecido contacto con Good, la política es seguirlo (sin mucho éxito para el seguidor trivial y de manera muy exitosa para el oráculo). El (relativamente) mal comportamiento para el seguidor trivial en los entornos con 3 celdas, se explica porque en un espacio pequeño, existe una mayor probabilidad de que coincida con Evil en la misma celda. Q-learning es capaz de sobrepasar al seguidor trivial (tomando ventaja del patrón de Good y Evil) pero solo después de una lenta convergencia, la cual es más lenta conforme va creciendo el número de celdas (como era de esperar).

\begin{figure}[h!]
\centering
\includegraphics[width=0.49\textwidth]{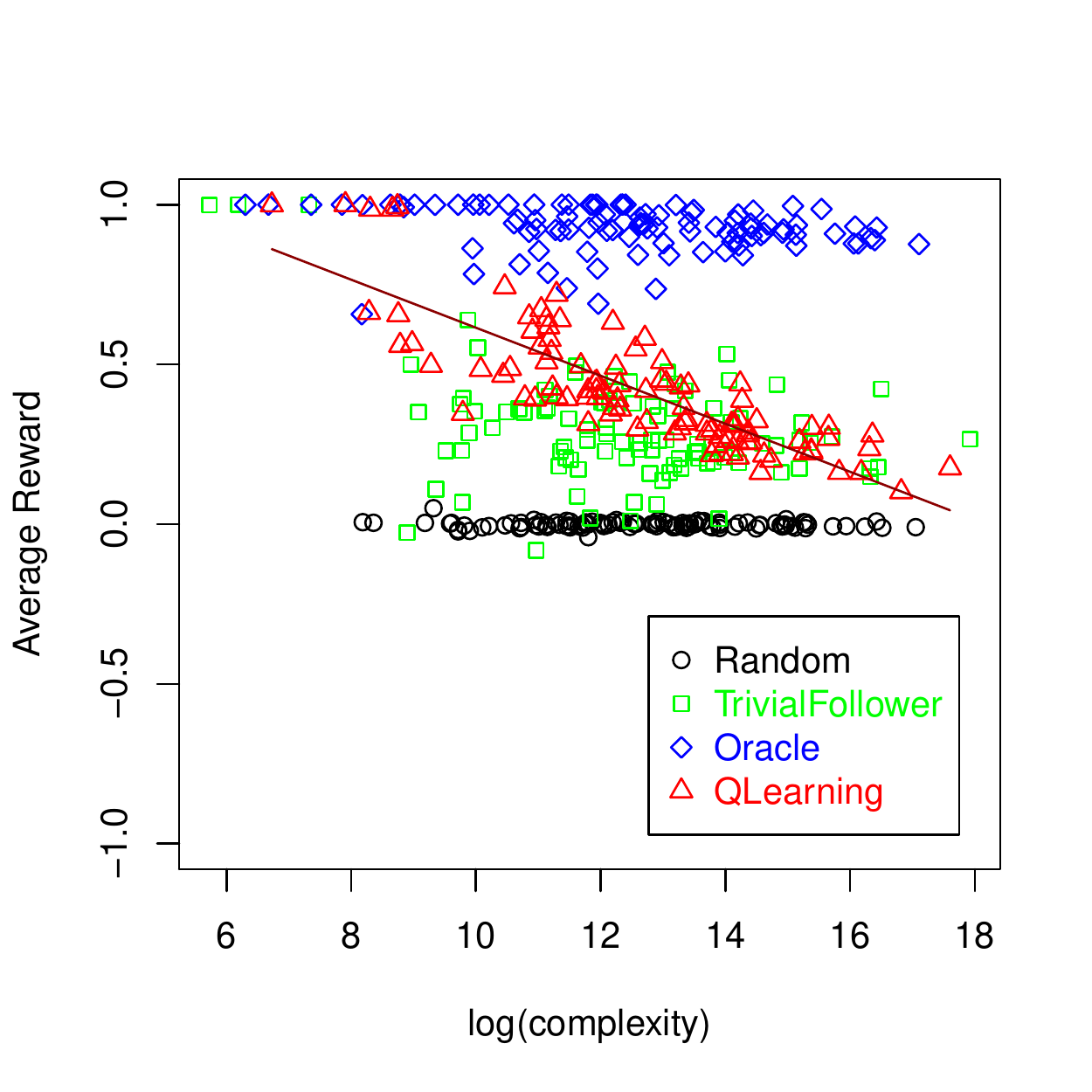}
\includegraphics[width=0.49\textwidth]{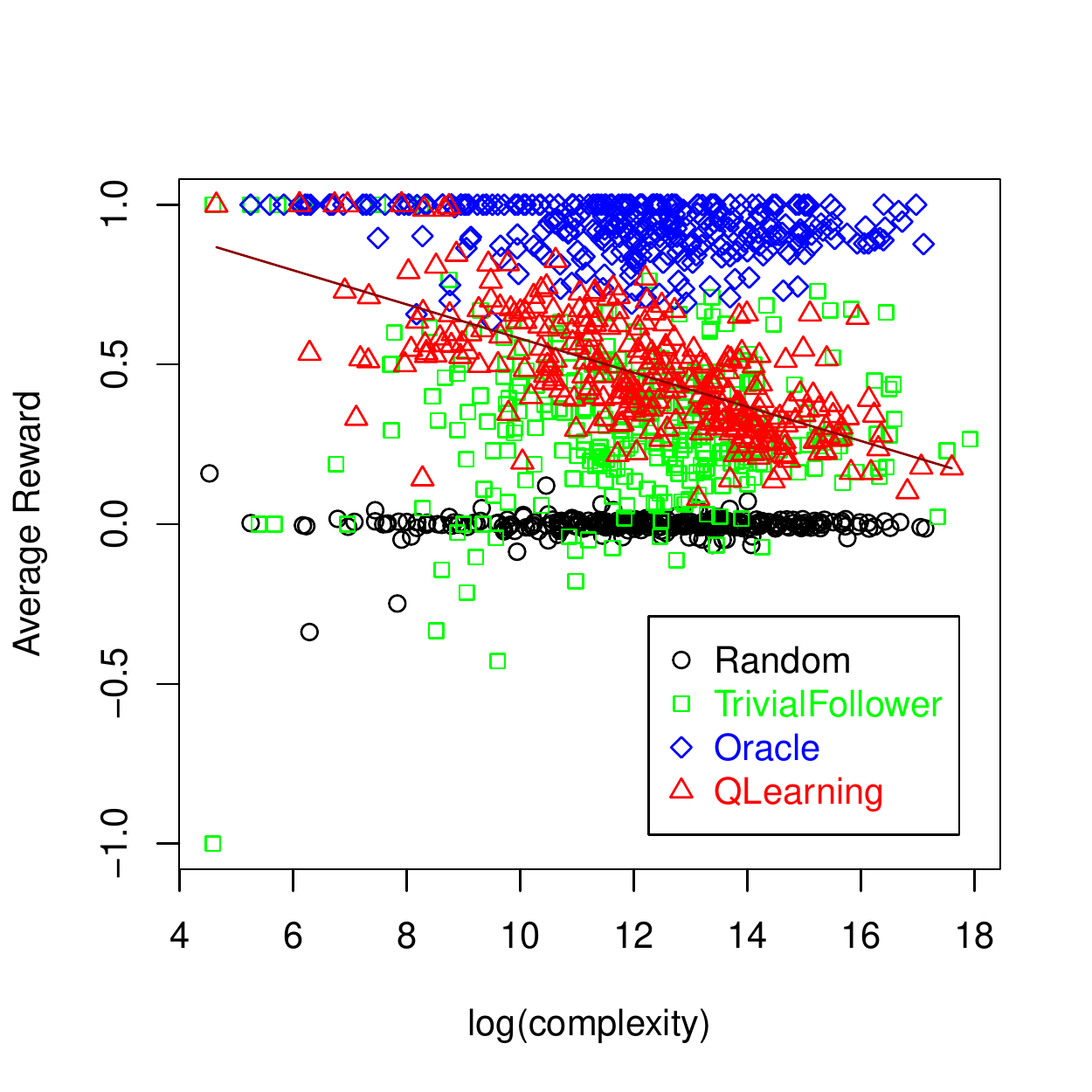}
\caption{Resultados tras 10.000 iteraciones comparadas con respecto a la complejidad del entorno. Izquierda: 100 entornos con 9 celdas. Derecha: 300 entornos con (3, 6, 9) celdas. La regresión lineal está formada a partir de los resultados de Q-learning.}
\label{fig:ComplejidadGenerados}
\end{figure}

Ahora analizamos el efecto de la complejidad. En la Figura~\ref{fig:ComplejidadGenerados}, vemos los resultados de las recompensas de los cuatro algoritmos tras 10.000 iteraciones comparados con la `complejidad' del entorno. Para aproximar esta complejidad, utilizamos el tamaño de la compresión de la descripción del espacio (la cual obviamente depende del número de celdas y de acciones), denotado por \emph{S}, y la descripción del patrón para Good y Evil, denotado por \emph{P}. Más formalmente, dado un entorno, aproximamos su complejidad (Kolmogorov), denotado por $K^{approx}$, como sigue:

\[ K^{approx} = LZ(concat(S,P)) \times |P| \]

Por ejemplo, si para describir el espacio de la Figura~\ref{fig:Espacio} utilizamos la cadena $S=$ ``{\footnotesize \ttfamily 12+3-{}-{}-{}-{}- | 12+++++3-{}-{}-{}-{}- | 1-2-{}-{}-{}-{}-{}-3++ | 1-{}-{}-{}-{}-2++++++3- | 12+3++++++ | 1-{}-{}-{}-{}-23-{}-{}-{}-{}-{}-{}- | 1++++++2-{}-{}-{}-{}-{}-{}-3++ | 1-{}-{}-{}-2+++3+}'' y el patrón para Good y Evil se describe como $P=$ ``{\footnotesize \ttfamily 203210200}'', concatenamos ambas cadenas (tamaño total 119 caracteres) y comprimimos la cadena resultante (utilizamos el método `gzip' de la función \emph{memCompress} en R, una implementación de la codificación Lempel-Ziv dentro del proyecto GNU). La longitud de la cadena comprimida en este caso es de 60.

Podemos ver que tanto al agente aleatorio, al oráculo y al seguidor trivial apenas les afecta la complejidad (como era de esperar en esta configuración). Por el contrario, Q-learning muestra un decrecimiento significativo en su rendimiento en cuanto la complejidad\footnote{Nótese que la complejidad no es una medida del número de celdas, sino una aproximación a la complejidad Kolmogorov del entorno.} se va incrementando. En cada gráfico mostramos un modelo de regresión lineal estándar en función de los datos de Q-learning, donde podemos apreciar mejor este decrecimiento.

\section{Experimentos sobre entornos `sociales'}
Hemos realizado otros experimentos donde incluimos a todos los agentes al mismo tiempo en el entorno, el objetivo es ver si pueden obtener alguna ventaja a partir de los movimientos de los otros agentes, o, por el contrario, esto incrementa la cantidad de información a procesar y se reduce el rendimiento de los agentes.

Comparamos este escenario `social' utilizando la misma configuración que en la Sección~\ref{sec:ExperimentosEntornosGenerados}. Incluimos todos los agentes a la vez, pero las recompensas no se dividen entre los agentes. En otras palabras, si la recompensa de una celda es de $0,5$ y dos agentes se sitúan en esa celda, entonces se les proporciona a ambos la recompensa de $0,5$. Esto nos permite analizar si el rendimiento en estos entornos (en donde existen más agentes al mismo tiempo) es beneficioso o no para cada tipo de agente.

\begin{figure}[h!]
\centering
\includegraphics[width=0.49\textwidth]{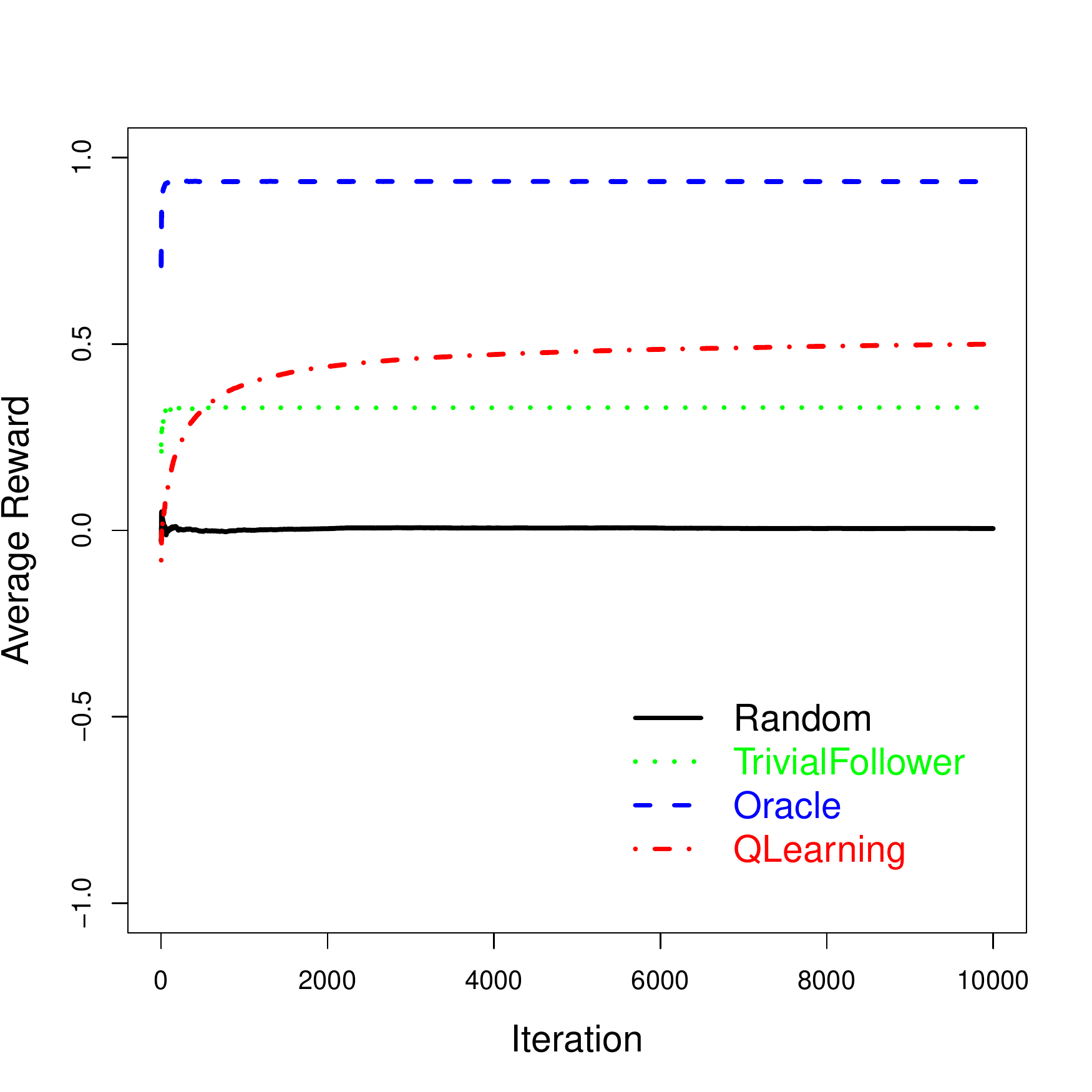}
\includegraphics[width=0.49\textwidth]{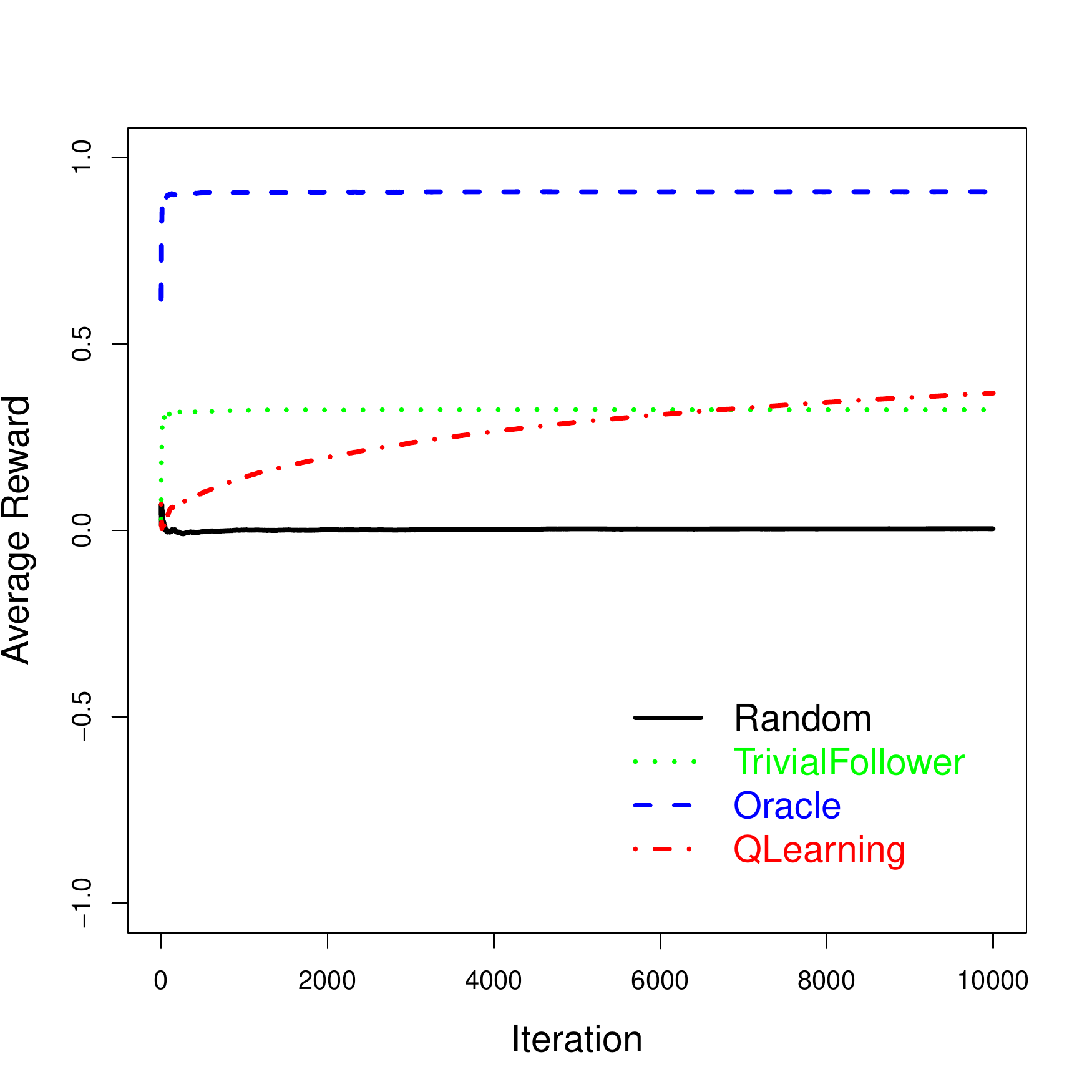}
\includegraphics[width=0.49\textwidth]{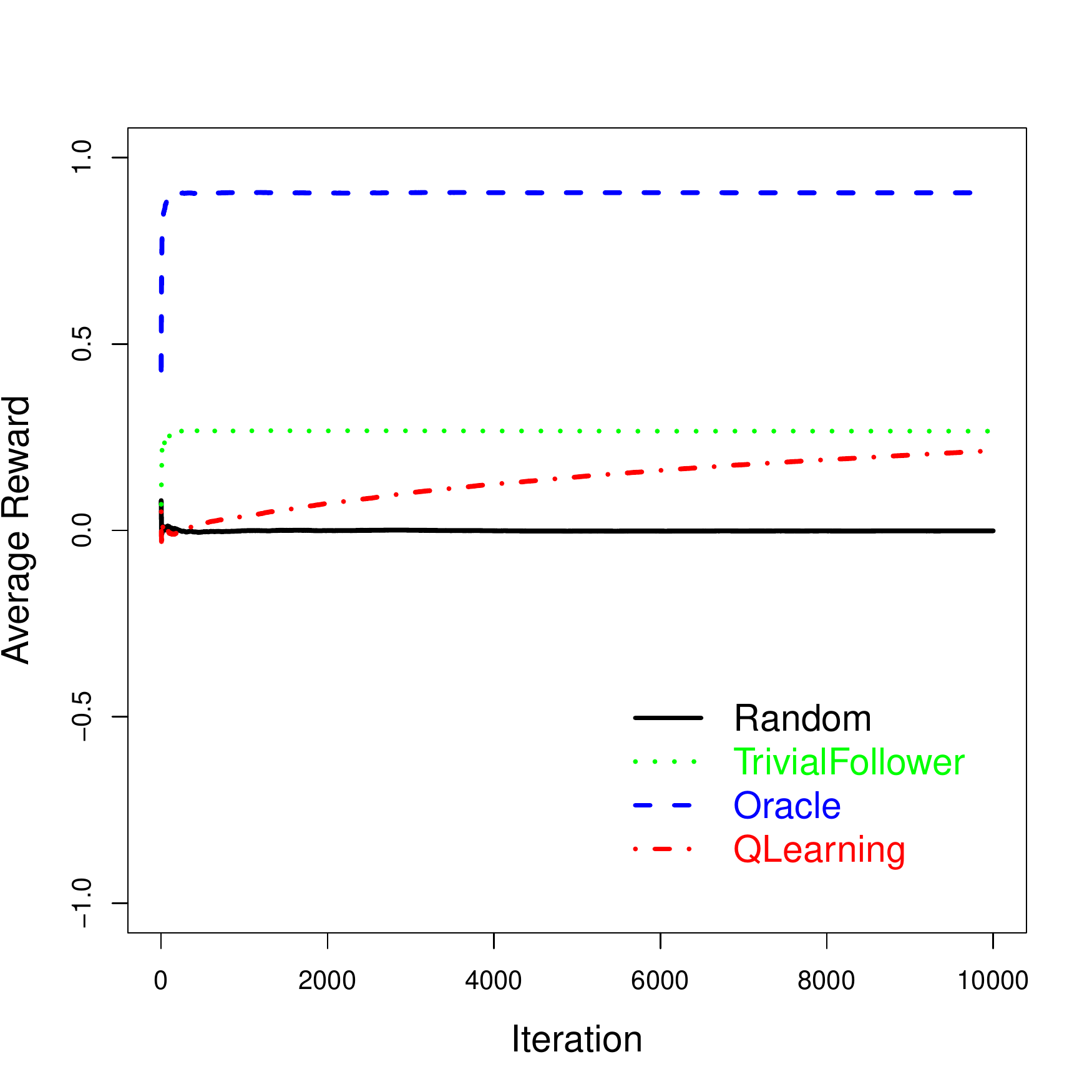}
\caption{Los resultados para $100$ entornos con comportamiento social.
Arriba-Izquierda: 3 celdas. Arriba-Derecha: 6 celdas. Abajo: 9 celdas.}
\label{fig:ResultadosEntornosSociales}
\end{figure}

En la Figura~\ref{fig:ResultadosEntornosSociales} podemos ver que los agentes oráculo y seguidor trivial no resultan afectados por este comportamiento social, sin embargo el rendimiento de Q-learning se reduce. La razón es que el tamaño de la observación (y por lo tanto de la matriz Q) se incrementa, ya que la información de las posiciones de los demás agentes hace que los entornos sean mucho más complejos para el Q-learning (mientras que en realidad no lo son tanto). Se puede ver la diferencia si comparamos las Figuras~\ref{fig:ResultadosEntornosGenerados} y \ref{fig:ResultadosEntornosSociales}. Q-learning no se aprovecha del comportamiento de los demás agentes para mejorar el suyo, sino todo lo contrario, obtiene peores resultados.

\begin{figure}[h!]
\centering
\includegraphics[width=0.49\textwidth]{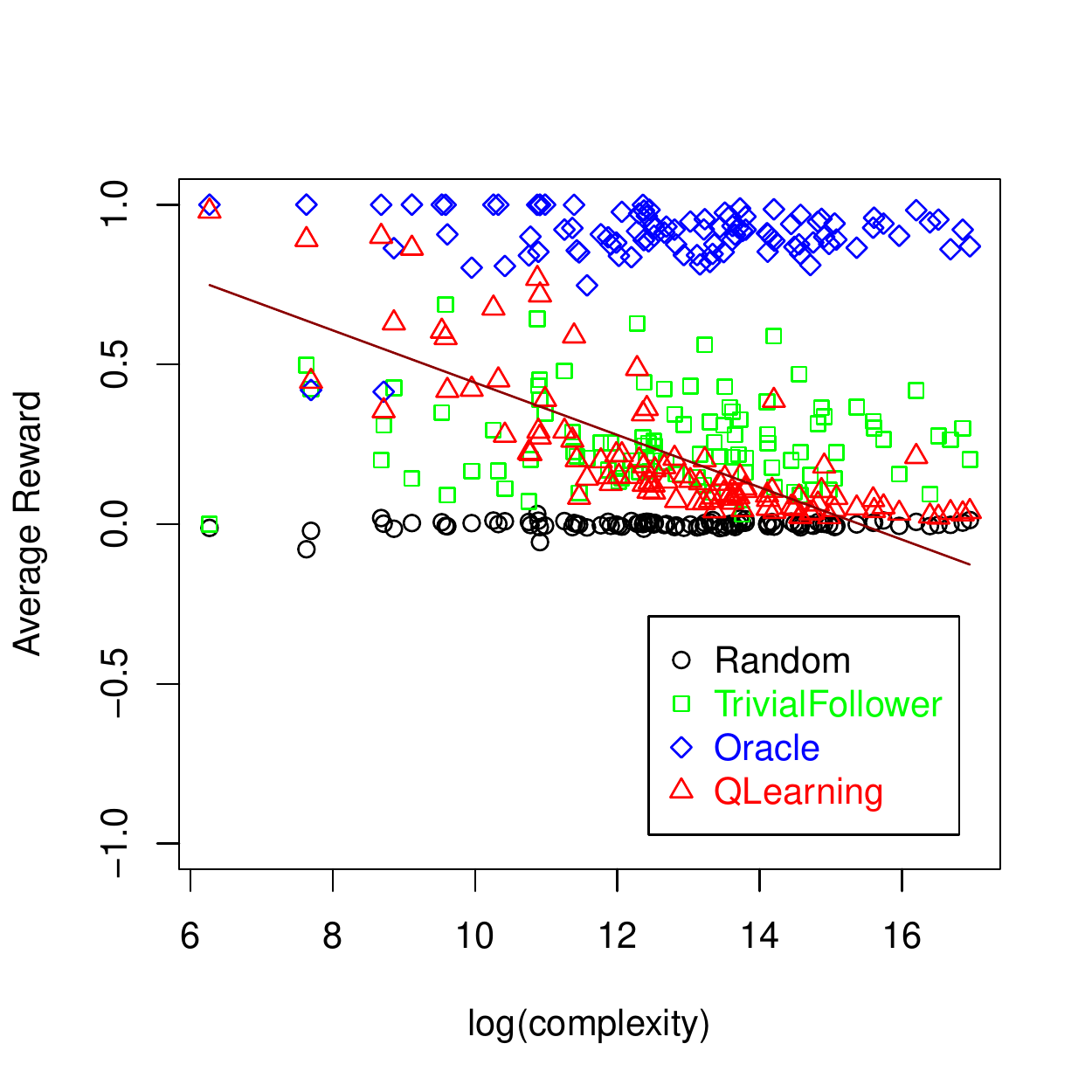}
\includegraphics[width=0.49\textwidth]{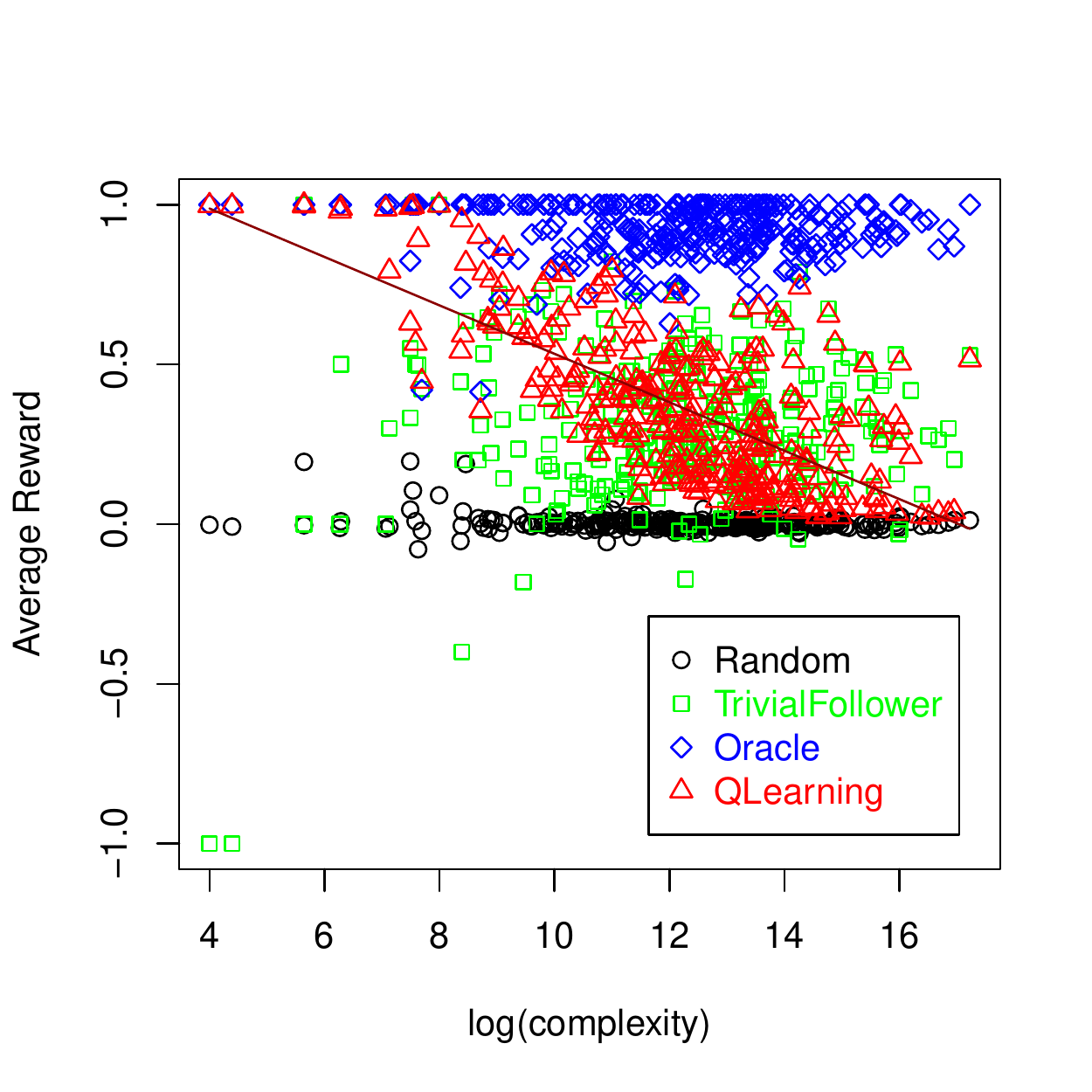}
\caption{Resultados tras 10.000 iteraciones comparadas con respecto a la complejidad del entorno con comportamiento `social'. Izquierda: 100 entornos con 9 celdas. Derecha: 300 entornos con (3, 6, 9) celdas. La regresión lineal está formada a partir de los resultados de Q-learning.}
\label{fig:ComplejidadSocial}
\end{figure}

La Figura~\ref{fig:ComplejidadSocial} muestra en este caso la evolución de los algoritmos con respecto a la complejidad en los entornos `sociales'. De nuevo, el gráfico muestra una relación inversa entre la complejidad y el rendimiento de Q-learning, siendo aun más acusada la pendiente de la regresión lineal. Sin embargo, esta relación de la complejidad no es tan directa, ya que no se ha tenido en cuenta la complejidad del resto de agentes en la medida.

\section{Conclusiones}
Los experimentos mostrados en este capítulo muestran que es posible utilizar esta implementación del test de inteligencia, utilizando la clase de entornos $\Lambda$, para evaluar sistemas de inteligencia artificial. Aunque para la implementación se han utilizado varias aproximaciones y simplificaciones, mostramos que la teoría funciona en capturar la esencia de la complejidad de las tareas independientemente de la aplicación o sistema de IA que se vaya a evaluar.

El objetivo de estos experimentos no era el de analizar las más que conocidas propiedades de Q-learning (tal como la convergencia, sobrecarga, etc.), ni tampoco designar un algoritmo `ganador'. El objetivo era mostrar que esta aproximación top-down para evaluar agentes de IA puede funcionar en la práctica.

Hemos utilizado esta implementación del test de inteligencia para evaluar a Q-learning como un algoritmo estándar típico en RL, y hemos podido ver que existe una relación (inversa) entre el rendimiento de Q-learning y la complejidad de los entornos y como este test es capaz de relacionarlo.

\paginablanco

\chapter{Evaluación de distintos sistemas inteligentes}\label{cap:EvaluacionDistintosAlgoritmos}
Tras haber comprobado (en el capítulo anterior) que estos tests pueden utilizarse para evaluar sistemas de inteligencia artificial, el siguiente paso a seguir es comprobar si los resultados obtenidos por distintos tipos de sistemas inteligentes reflejan realmente la diferencia esperada.

Para realizar este experimento hemos elegido dos tipos de sistemas distintos, los hemos evaluado y hemos comparado sus resultados. Por un lado, hemos elegido el mismo `representante estándar' de los algoritmos de IA que en el capítulo anterior (Q-learning) y, por otro lado, a un fácilmente accesible sistema biológico (Homo sapiens).

\section{Configuración de los experimentos}\label{sec:ConfigCompararIAvsBiologicos}
Para cada test, hemos utilizado una secuencia de 7 entornos, cada uno con un número de celdas ($n_c$) de 3 a 9. El espacio para cada ejercicio se genera al azar (respetando el número de celdas para cada ejercicio) durante el transcurso de los tests, al igual que el patrón de comportamiento que seguirán Good y Evil, cuyo tamaño se calcula en proporción (en promedio) al número de celdas, utilizando $p_{stop} = 1/n_c$. En cada entorno, vamos a permitir $10 \times (n_c - 1)$ interacciones, por lo que los agentes tienen la oportunidad de detectar cualquier patrón en el entorno (periodo de exploración) y también de tener algunas interacciones más para explotar sus descubrimientos (dado el caso de que realmente se descubra el patrón). La limitación del número de entornos y de interacciones se justifica porque los tests están pensados para aplicarse a agentes biológicos en un periodo razonable de tiempo (p.~ej. 20 minutos), estimando unos 4 segundos para cada acción. Q-learning ha sido evaluado en una batería de tests, mientras que cada humano ha sido evaluado en un único test. La Tabla~\ref{tab:ConfigTestCompararIAvsBiologicos} resume las decisiones que hemos tomado para el test.

\begin{table}[h!]
{\scriptsize
\centering
	\begin{tabular}{| c || c | c | c | }
		\hline
		Entorno \# & No. celdas ($n_c$) & No. interacciones ($m$) & $p_{stop}$ \\ \hline
		1 & 3 & 20 & $ 1/3 $ \\ \hline
		2 & 4 & 30 & $ 1/4 $ \\ \hline
		3 & 5 & 40 & $ 1/5 $ \\ \hline
		4 & 6 & 50 & $ 1/6 $ \\ \hline
		5 & 7 & 60 & $ 1/7 $ \\ \hline
		6 & 8 & 70 & $ 1/8 $ \\ \hline
		7 & 9 & 80 & $ 1/9 $ \\ \hline
		TOTAL & - & 350 & - \\ \hline
	\end{tabular}
\caption{Configuración para los 7 entornos que componen el test.}
\label{tab:ConfigTestCompararIAvsBiologicos}
}
\end{table}

Para que los resultados obtenidos fueran más comparables, hemos hecho que ambos tipos de agentes realizasen exactamente los mismos ejercicios, evaluándolos simultáneamente en los mismos entornos (siempre con los mismos movimientos para Good y Evil). Sin embargo, hemos evitado que pudieran verse durante su evaluación, de modo que los hemos evaluado por separado. Esto es debido a que si ambos agentes hubieran coincidido en el mismo ejercicio hubieran sido conscientes de la existencia del otro agente, lo cual hubiera introducido una mayor (e innecesaria) complejidad.

Con respecto al resto de características, hemos utilizado las mismas que en el capítulo anterior (véase Página~\pageref{itm:CaracteristicasEntornos}). De forma resumida: utilizamos interfaces completamente observables (omitiendo al resto de agentes evaluados), las recompensas se consumen \emph{después} de actualizarlas y se mantienen únicamente durante una interacción.

\section{Agentes evaluados y sus interfaces}
Ahora vamos a introducir a los agentes que vamos a evaluar, Q-learning y el Homo sapiens, así como las interfaces que ha utilizado cada uno.

\subsection{Un agente de IA: Q-learning}
Hemos elegido a Q-learning como `representante' de los algoritmos de IA para estos experimentos.

Para la descripción del estado de Q-learning, hemos utilizado la descripción del contenido de las celdas. Los parámetros elegidos son los mismos que en el capítulo anterior: $\alpha = 0,05$ (\emph{learning rate}) y $\gamma = 0,35$ (\emph{discount factor}). Los elementos en la matriz $Q$ se establecen inicialmente a $2$ (las recompensas tienen un rango entre $-1$ y $1$, pero están normalizados entre $0$ y $2$ para que la matriz $Q$ sea siempre positiva). Los parámetros se han elegido probando 20 valores consecutivos para $\alpha$ y $\gamma$ entre $0$ y $1$. Estas $20$ x $20$ = $400$ combinaciones se han evaluado en 1.000 sesiones cada uno utilizando entornos aleatorios de diferentes tipos y complejidades. Por supuesto, esta elección es beneficiosa para el rendimiento de Q-learning en los tests.

Para la interfaz de Q-learning se le proporciona una copia del espacio (completamente observable), incluyendo la localización de los agentes dentro de las celdas. A partir de esta copia, se construye la descripción del estado para la matriz Q. Podemos ver un ejemplo de esta descripción en el Ejemplo~\ref{ejm:DescripciónEstadoQLearning} de la Página~\pageref{ejm:DescripciónEstadoQLearning}.

\subsection{Un agente biológico: Homo sapiens}
Hemos elegido al Homo sapiens por razones obvias, en especial porque generalmente se asume que es la especie más inteligente del planeta. Hemos cogido 20 humanos de un departamento universitario (estudiantes de doctorado, personal de investigación y profesores) con edades comprendidas entre 20 y 50 años. Esta es una muestra en donde se espera que sus habilidades intelectuales sean en promedio mejores que la de la población general, por lo que tendremos esto en cuenta durante la evaluación de los resultados.

\begin{figure}[h!]
\centering
\includegraphics[width=1\textwidth]{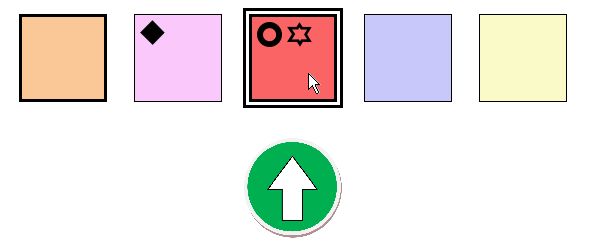}
\caption{Una captura de la interfaz para humanos. El agente acaba de recibir una recompensa positiva, mostrado a través del círculo verde con una flecha hacia arriba. La imagen también muestra que el agente está situado en la celda 3, y Good y Evil se encuentran en las celdas 3 y 2 respectivamente. El agente se puede mantener en la celda 3 y también puede moverse a la celda 1. La celda 3 está destacada ya que es una celda accesible y el cursor se encuentra encima de ella.}
\label{fig:InterfazHumanos2}
\end{figure}

Para la interfaz, hemos utilizado el modo de uso anteriormente descrito en la Sección~\ref{sec:ModoHumano}. Podemos ver una captura de esta interfaz en la Figura~\ref{fig:InterfazHumanos2}.

En estos test se han tomado varias decisiones sobre la interfaz: (1) Para cada número de ejercicio se utiliza el mismo orden de colores, (2) los símbolos para cada agente (al igual que los colores de las celdas) se mantienen para cada número de ejercicio y (3) cada ejercicio tiene pre-asignados los símbolos que se utilizan para Good y Evil. Estas decisiones hacen que la interfaz sea lo más homogénea posible y evitan posibles sesgos entre los humanos.

\begin{figure}[h!]
\centering
\includegraphics[width=1\textwidth]{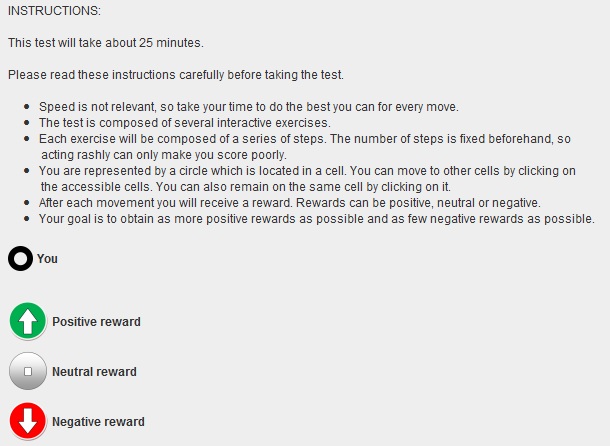}
\caption{Instrucciones mostradas a los seres humanos antes de comenzar su evaluación.}
\label{fig:InstruccionesHumanos}
\end{figure}

Al principio del test al sujeto se le presentan las instrucciones, las cuales estrictamente contienen lo que el usuario debe conocer, entre ellas: (1) que el test está compuesto de una serie de ejercicios interactivos, (2) la existencia de recompensas y su objetivo de conseguir el máximo número de recompensas positivas, (3) que no se mide la velocidad, por lo que actuar rápidamente solo conllevará (en general) malos resultados y (4) la existencia de un agente que lo representa y su símbolo. Podemos ver estas instrucciones en la Figura~\ref{fig:InstruccionesHumanos}.

\section{Resultados del experimento}
Como hemos indicado anteriormente, se han realizado 20 tests (con 7 ejercicios en cada uno) con la configuración mostrada en la Tabla~\ref{tab:ConfigTestCompararIAvsBiologicos}, administrado cada test a un humano distinto mientras que Q-learning se ha evaluado en todos ellos.

\begin{figure}[h!]
\centering
\includegraphics[width=0.49\textwidth]{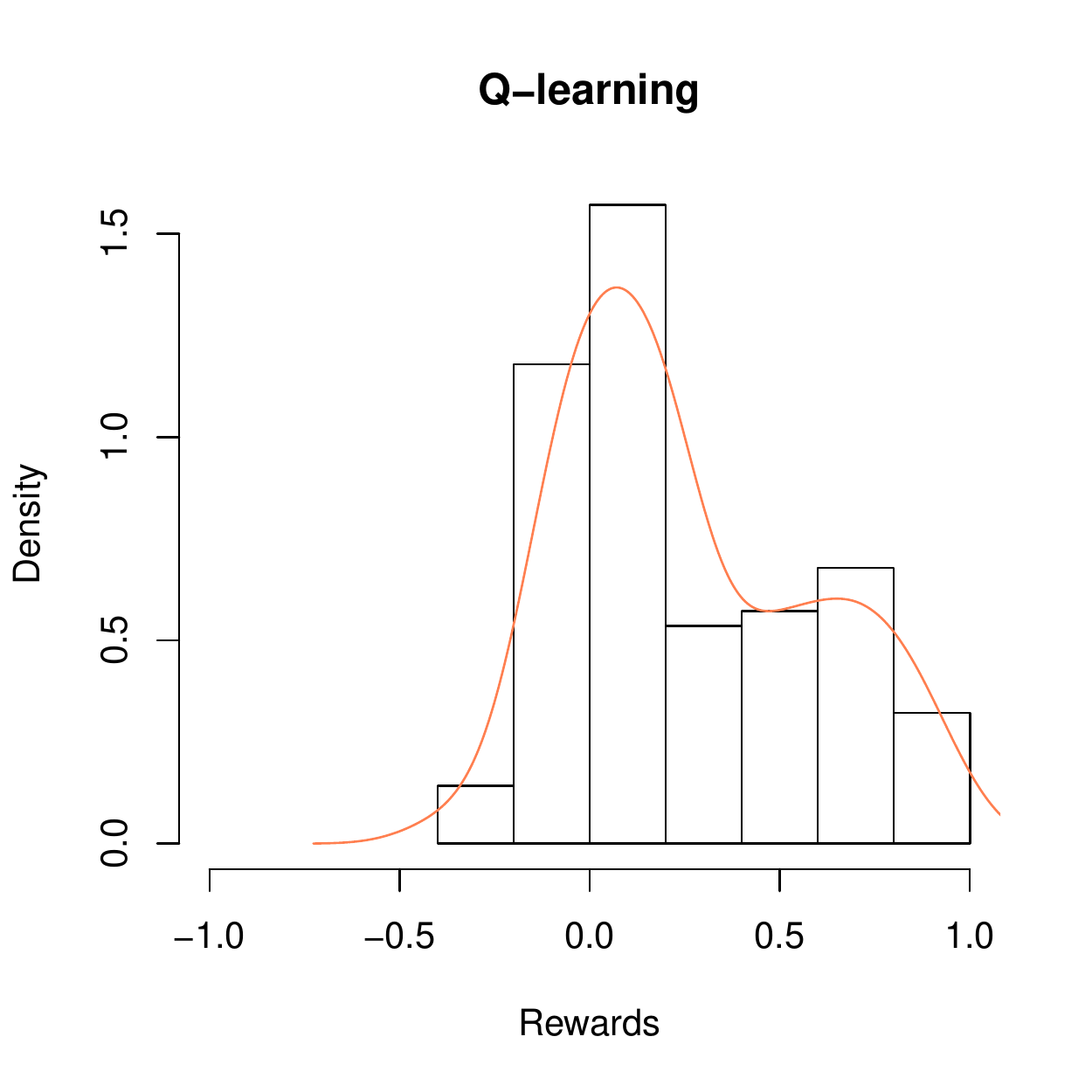}
\includegraphics[width=0.49\textwidth]{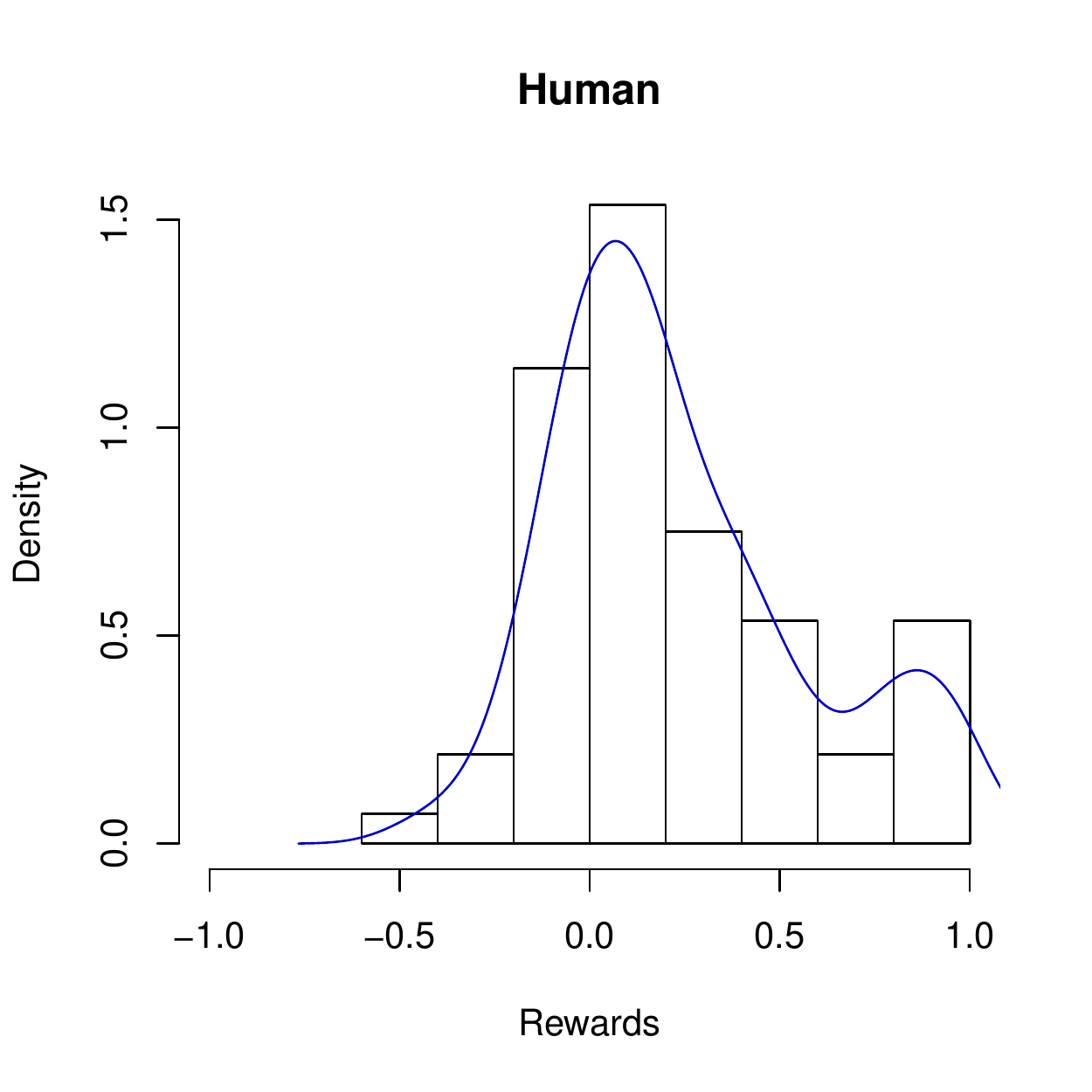}
\caption{Histogramas de los $20 \times 7 = 140$ ejercicios para Q-learning (izquierda) y humanos (derecha). Las lineas muestran las probabilidades de densidad.}
\label{fig:HistogramaCompararIAvsBiologicos}
\end{figure}

La Figura~\ref{fig:HistogramaCompararIAvsBiologicos} muestra los histogramas y las densidades de probabilidad (estimados con el paquete R). La primera observación que podemos ver a partir de este conjunto de pares de resultados son las medias. Mientras que Q-learning tiene una media global de 0,259, los humanos muestran una media de 0,237. Las desviaciones estándares son 0,122 y 0,150 respectivamente. Ambas imágenes son bastante parecidas (aparte de la mayor variación de los humanos) y no se percibe ninguna diferencia especial.

\begin{figure}[h!]
\centering
\includegraphics[width=0.49\textwidth]{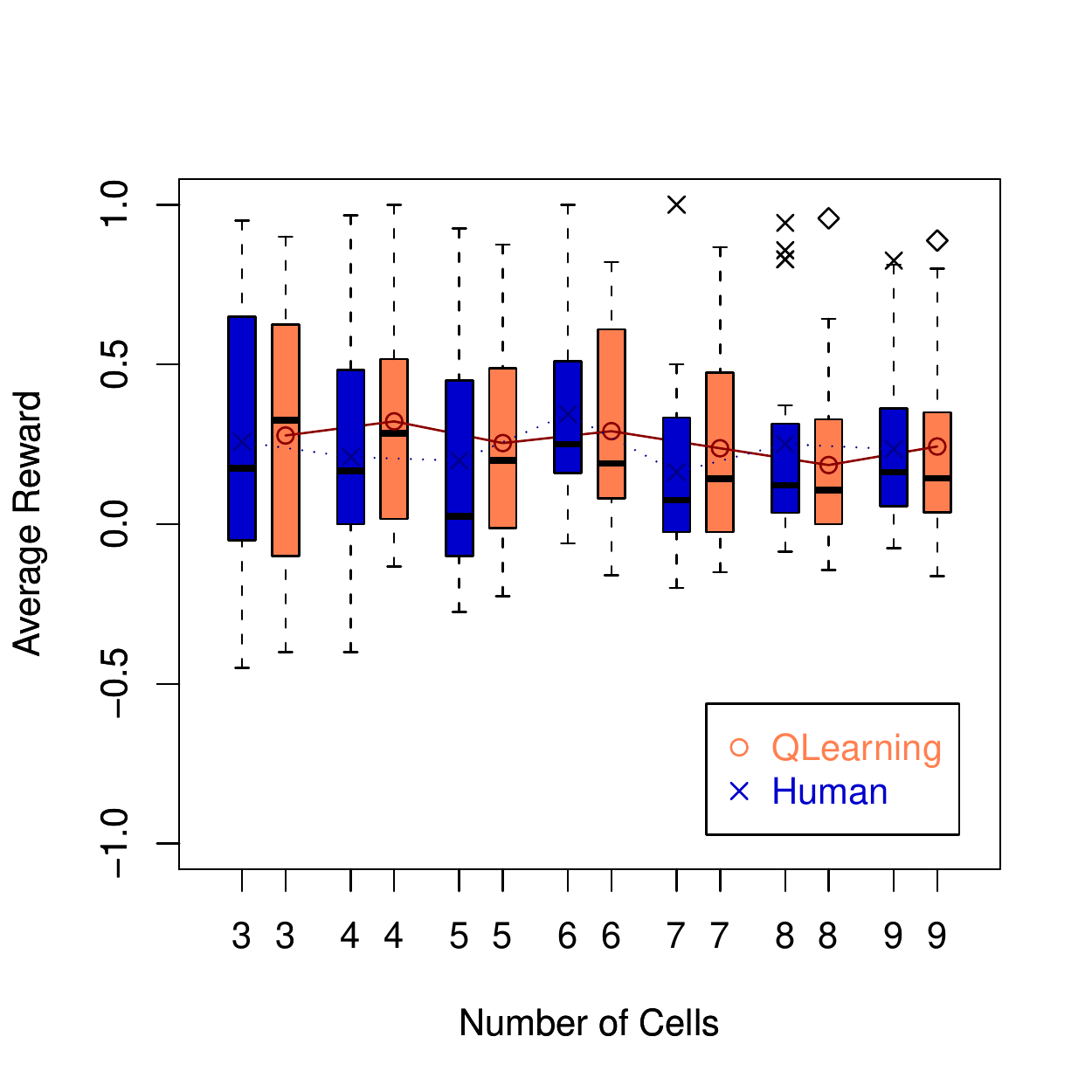}
\includegraphics[width=0.49\textwidth]{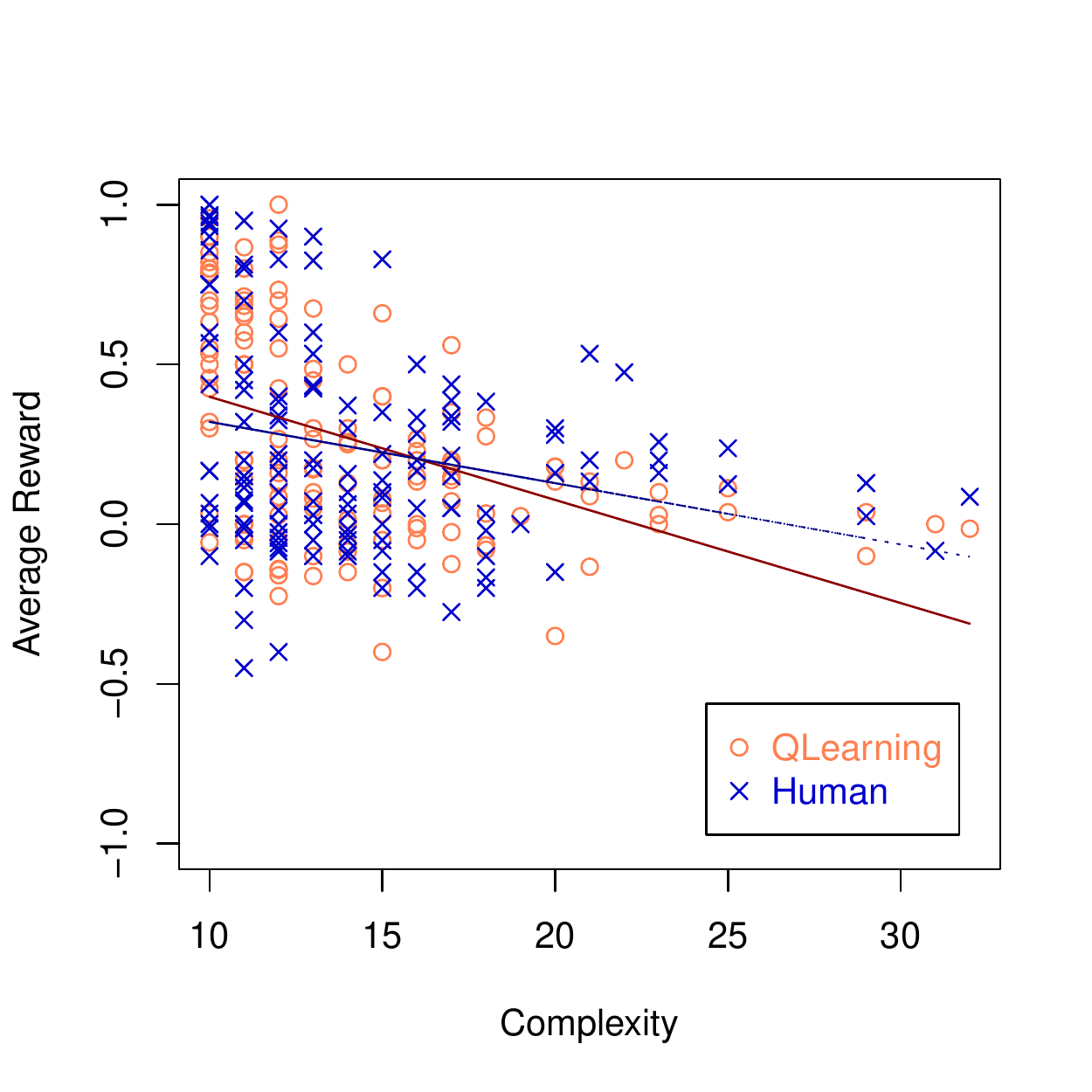}
\caption{Izquierda: Diagrama de cajas (con bigotes) para los siete ejercicios dependiendo del agente. Las medianas se muestran en la caja como un pequeño segmento negro, mientras que las medias se conectan entre ellas por una linea continua para Q-learning y una linea discontinua para los humanos. Derecha: Los resultados medios de las recompensas para los \mbox{$20 \times 7 \times 2 = 280$} ejercicios utilizando $K^{approx}$ como medida de la complejidad. También se muestra una regresión lineal para ambos tipos de agentes.}
\label{fig:DiagramaCajas}
\end{figure}

Para ver los resultados con mayor detalle (en términos de los ejercicios), la Figura~\ref{fig:DiagramaCajas} (izquierda) muestra los resultados agregando por ejercicio (un ejercicio para cada número de celdas entre $3$ y $9$, en total $7$ ejercicios por test). Esta figura muestra la media, la mediana y la dispersión tanto de Q-learning como de los humanos para cada ejercicio. Si miramos en los diagramas de cajas para cada tamaño del espacio, podemos ver que tampoco existe una diferencia significativa en términos de cómo funcionan Q-learning y los humanos en cada uno de los siete ejercicios. Mientras las medias son alrededor de 0,2 y 0,3, las varianzas son más pequeñas cuanto más grande es el número de celdas. Esto se debe a que en los ejercicios con mayor cantidad de celdas se dispone de un mayor número de iteraciones (véase la Tabla~\ref{tab:ConfigTestCompararIAvsBiologicos}).

Con los resultados obtenidos, hemos aplicado un ANOVA repetido de dos variantes (two-way repeated ANOVA): \mbox{agente $\times$ número de celdas}. El ANOVA realizado no muestra ningún efecto estadístico significativo para el agente (\mbox{$F_{1,19} = 0,461$}, \mbox{$P = 0,506$}) ni tampoco para el número de celdas (\mbox{$F_{6,114} = 0,401$}, \mbox{$P = 0,877$}). Tampoco se ha encontrado ningún efecto en las interacciones estadísticamente significante (\mbox{$F_{6,114} = 0,693$}, \mbox{$P = 0,656$}).

Finalmente, ya que el número de celdas no es una medida suficiente de la complejidad, hemos explorado la relación con la complejidad de los entornos. Para aproximar esta complejidad, de manera similar al capítulo anterior hemos utilizado el tamaño del patrón comprimido para Good y Evil, denotado como $P$. Más formalmente, dado un entorno, calculamos una aproximación\footnote{Esta es una aproximación claramente tosca de la complejidad Kolmogorov (o incluso de una variante computable como el $Kt$ de Levin). Una descripción de un entorno muy largo (e incomprensible) puede tener un comportamiento muy similar (o incluso igual) que un entorno mucho más simple durante un largo número de iteraciones. Una alternativa sería evaluar la complejidad Kolmogorov del ejercicio. Algunas de estas cuestiones se discuten en \cite{HernandezOralloDowe2010}.} a su complejidad (Kolmogorov), denotado $K^{approx}$ como sigue:

\[ K^{approx} = LZ(P) \]

Por ejemplo, si un patrón para \emph{Good} y \emph{Evil} se describe con la cadena \mbox{$P=$ ``{\footnotesize \ttfamily 20122220022222200222222002}''}, comprimimos la cadena (utilizando la función $memCompress$ de R, con una implementación de Lempel-Ziv dentro del proyecto GNU). La longitud de esta cadena comprimida es 19.

La Figura~\ref{fig:DiagramaCajas} (derecha) muestra cada uno de los $20 \times 7 = 140$ ejercicios para cada tipo de agente. De nuevo vemos una mayor dispersión para los humanos que para Q-learning, lo cual es debido a que los 20 humanos son distintos, mientras que Q-learning es exactamente el mismo algoritmo para cada uno de los 20 tests. Si realizamos una regresión lineal para cada tipo de agente, vemos que la pendiente para los humanos es menos pronunciada  que la de Q-learning.

Hemos calculado el coeficiente de correlación de Pearson entre la complejidad y las recompensas. Ahora encontramos una correlación estadísticamente significativa tanto para los humanos (\mbox{$r = -0,257$}, \mbox{$n = 140$}, \mbox{$P= 0,001$}) como para Q-learning (\mbox{$r = -0,444$}, \mbox{$n = 140$}, \mbox{$P < 0,001$}). También analizamos estas correlaciones por el número de celdas, como se muestra en la Tabla~\ref{tab:Correlaciones}. Esta tabla muestra el coeficiente de correlación de Pearson y los niveles de asociación significativos entre ``complejidad'' y ``recompensa'' por ``número de celdas'' para cada agente, todos con \mbox{$n = 20$}.

\begin{table}[h!]
{
\centering
\resizebox*{1\textwidth}{!}{
	\begin{tabular}{| c || c | c | c | c | c | c | c |}
	\hline
	Agente & 3 celdas & 4 celdas & 5 celdas & 6 celdas & 7 celdas & 8 celdas & 9 celdas \\ \hline
	Humano & -.474 (.017) & -.134 (.286) & -.367 (.056) & -.515 (.010) & -.282 (.114) & -.189 (.213) & -.146 (.270) \\ \hline
	Q-learning & -.612 (.002) & -.538 (.008) & -.526 (.009) & -.403 (.039) & -.442 (.026) & -.387 (.046) & -.465 (.019) \\ \hline
	\end{tabular}}
\caption{Coeficiente de correlación de Pearson y los valores de $p$ (entre paréntesis) entre ``complejidad'' y ``recompensa'' por ``número de celdas''.}
\label{tab:Correlaciones}
}
\end{table}

Podemos ver que las correlaciones son más fuertes y siempre significativas para Q-learning, mientras que son más leves (y no siempre significativas) para los humanos. Esto podría deberse a que los humanos no son `reseteados' entre ejercicios. En general, necesitaríamos más datos (más tests y más interacciones para cada ejercicio) para confirmar o refutar esta hipótesis.

\section{Impresiones del Homo sapiens}
Tras el test, se les pidió a los humanos rellenar un cuestionario para obtener información con respecto a ciertos aspectos del test, como sus emociones (p.~ej. si se sentían motivados para realizar el test), la usabilidad del interfaz y las estrategias que siguieron durante el test (p.~ej. como actuaban para conseguir maximizar sus recompensas).

Entre sus observaciones podemos destacar que la mayoría lo encontró motivante, por lo que desempeñaban un rendimiento constante durante todo el test. Sin embargo, reconocen que el test (tal y como se planeaba) era algo difícil de realizar y de ahí justifican (en general) sus bajos resultados.

La mayoría de los evaluados reconocieron no comprender la mecánica de los ejercicios (agentes Good y Evil que proporcionaban las recompensas y que seguían un mismo patrón de movimientos) hasta la mitad del test. Esta es una explicación más que justifica los bajos resultados y la amplia varianza en los primeros ejercicios de la Figura~\ref{fig:DiagramaCajas} (izquierda), ya que (mayormente) su estrategia se centraba en realizar acciones al azar hasta comprender lo que debían hacer (siendo aún mejores que el valor esperado para un agente aleatorio ($0$), demostrando que sus acciones comenzaban a condicionarse por las recompensas positivas).

La amplia mayoría descubrió la existencia de los agentes Good y Evil, y también descubrieron que seguían cada uno un patrón de movimientos, los cuáles consideraron bastante difíciles de memorizar (hay que tener en cuenta que (según sus propias impresiones) no descubrieron la mecánica de los ejercicios hasta la mitad del test, en donde (en general) ya comenzaban a complicarse los patrones). Ningún encuestado descubrió ninguna otra regla del comportamiento de los entornos (entorno estocástico, mismo comportamiento para Good y Evil, \dots) a excepción de las ya mencionadas.

Con respecto a la usabilidad del interfaz, la mayoría se puso de acuerdo en que su diseño era el adecuado y que ofrecía la información necesaria sin la existencia de ruido que molestase para realizar el test. La única excepción fueron los colores de las celdas que, según sus comentarios, distrajeron bastante a la hora de descifrar el motivo por el cual se obtenían recompensas positivas. Quizás eliminando esta complejidad añadida los humanos hubieran obtenido mejores puntuaciones.

\section{Conclusiones}
A partir de este experimento hemos obtenido muchas conclusiones interesantes:

\begin{itemize}
\itemsep=0px
\item Es posible hacer el mismo test tanto a máquinas como a humanos sin necesidad de que sea antropomórfico. El test es exactamente el mismo para ambos y se deriva a partir de conceptos computacionales. Simplemente hemos ajustado la interfaz (cómo se muestran las recompensas, acciones y observaciones) dependiendo del tipo de sujeto.
\item Los humanos no son mejores que Q-learning en este test, incluso aunque el test (a pesar de varias simplificaciones) está basado en una distribución universal de entornos sobre una clase de entornos muy general. De hecho, creemos que el uso de una distribución universal es precisamente una de las razones por las que los humanos no son especialmente buenos en este test.
\item Ya que estos resultados son similares a \cite{sanghidowe2003computer} (el cual muestra que las máquinas pueden obtener buenos resultados en IQ tests), nos encontramos con una evidencia adicional de que un test que es válido para seres humanos o para máquinas por separado podría no ser útil para distinguirlos o para situarlos en una misma escala y, por lo tanto, no pueda servir como test de inteligencia universal.
\end{itemize}

Estos tests pretenden ser lo más generales posibles. Es verdad que los autores han realizado muchas simplificaciones a la clase de entornos, como por ejemplo que Good y Evil no reaccionan al entorno (simplemente ejecutan una secuencia cíclica de acciones siguiendo un patrón), y también hemos utilizado una aproximación muy simple para la complejidad en lugar de mejores aproximaciones a la complejidad Kolmogorov o a la $Kt$ de Levin. Además, los humanos no se `resetean' (ni tampoco se pueden) entre ejercicios y la muestra utilizada tiene unas habilidades intelectuales (en principio) superiores en promedio al de la población general. A pesar de todos estos problemas, los cuales (la mayoría) dan más mérito a Q-learning, pensamos (aunque no podemos concluirlo) que los tests no son lo suficientemente generales. Q-learning no es el mejor algoritmo de IA disponible hoy en día (de hecho no consideramos que Q-learning sea muy inteligente), por lo que los resultados no son representativos de las diferencias reales de inteligencia entre los humanos y Q-learning. Esta pérdida de generalidad puede deberse (parcialmente) a las extensas simplificaciones realizadas con respecto a la idea original del test anytime y a la clase de entornos.

Una posibilidad es que quizás el tamaño de nuestra muestra sea demasiado pequeño. Si hubiéramos tenido más entornos con mayor complejidad y dejando que los agentes interactuasen más tiempo en cada uno de ellos, quizás veríamos una imagen distinta. Sin embargo, no está claro que los humanos puedan escalar mejor en este tipo de ejercicios, especialmente si no se pueden reutilizar partes de ejercicios anteriores en los ejercicios nuevos. Primero, algunos de los patrones que han aparecido en muchos de los ejercicios complejos, han sido considerados muy difíciles por los humanos. Segundo, Q-learning requiere muchas interacciones para converger, por lo que quizás solo se exagerase la diferencia a favor de Q-learning. En cualquier caso, esto debería analizarse con más experimentos.

Un problema más fundamental, y en nuestra opinión más relevante, es si se están haciendo pruebas en el tipo de entornos equivocado. En concreto, nos referimos a la clase de entornos o a la distribución del entorno. La clase de entornos es una clase general que incluye dos agentes simétricos, Good y Evil, los cuales están a cargo de las recompensas. No creemos que esta clase de entornos esté, en ningún caso, sesgada en contra de los humanos (se podría decir lo contrario). En el fondo, es difícil de responder a la pregunta de si un test está sesgado o no, ya que cualquier simple elección implica un cierto sesgo. En nuestra opinión, el problema podría estar en la distribución de entornos. Eligiendo una distribución universal se da mayor probabilidad a entornos muy simples con patrones muy simples, pero lo que es más importante, hace que cualquier tipo de interacción más `rica' sea imposible incluso en entornos con una alta complejidad Kolmogorov. Por lo que una mejor distribución de entornos (o quizás clase) debería dar mayor probabilidad a la adquisición incremental de conocimiento, a las capacidades sociales y a entornos más reactivos.

Una de las cosas que hemos aprendido es que el cambio de distribuciones universales desde entornos pasivos (como se propuso originalmente en \cite{DoweHajek98} y \cite{HernandezOrallo00a}) a entornos interactivos (como también se sugiere en \cite{HernandezOrallo00a} y se desarrolla completamente en \cite{legg2005universal} y \cite{LeggHutter07}) se encuentra en la dirección correcta, pero no es (por sí sola) la solución. Está claro que permite una mayor interpretación natural de la noción de inteligencia como el rendimiento en un amplio rango de entornos, y facilita la aplicación de tests más allá de los humanos y las máquinas (niños, chimpancés, \dots), pero hay otros problemas en los que hay que centrarse para poder dar una definición más apropiada de la inteligencia y poder proporcionar un test más práctico. La propuesta de un test adaptativo \cite{HernandezOralloDowe2010} introduce muchas nuevas ideas sobre cómo crear tests de inteligencia más prácticos; sustituyendo la distribución universal por una distribución adaptativa permitirá una mayor convergencia hacia los niveles de complejidad más apropiados para los agentes. A partir de estos resultados creemos que la prioridad es definir nuevas distribuciones de entornos que puedan dar mayor probabilidad a entornos donde la inteligencia pueda mostrar todo su potencial (véase, p.~ej. \cite{AGI2011DarwinWallace}).

\chapter{Perspectivas}\label{cap:Perspectivas}
Con esta tesis de máster, hemos podido observar (con unos experimentos iniciales) como los tests desarrollados son válidos para evaluar varios tipos de sistemas por separado, pero no podemos concluir que (en su estado actual) sean apropiados para situar a distintos tipos de sistemas en una misma escala. Según los resultados obtenidos, no podemos asegurar de ninguna forma cuál de los dos sistemas evaluados es el más inteligente. Sin embargo, sin necesidad de realizar ningún test, damos por sentado que el ser humano es bastante superior al algoritmo de IA con el que lo hemos comparado.

Uno de los problemas de este sistema de evaluación es que, con la distribución universal utilizada para generar los entornos, existe una probabilidad prácticamente nula de que se generen entornos en donde se necesiten de unas capacidades cognitivas superiores para conseguir resolverlos. Esto se debe a que los entornos más simples se generan con mayor probabilidad, mientras que será muy improbable que se genere algún entorno mínimamente complejo. Por ejemplo, si generamos entornos siguiendo esta distribución, es muy difícil que (a partir de un entorno) consigamos obtener alguna información útil que nos ayude en futuros entornos. De igual modo, resulta prácticamente imposible llegar a generar algún tipo de comportamiento social. El comportamiento social ha necesitado de muchísimos millones de años para surgir en nuestro mundo y, además, solo ha sido capaz de aparecer a partir de la evolución de las especies tras múltiples mutaciones, por lo que pretender generarlo desde cero resulta prácticamente ridículo.

Si vemos como ha ido evolucionando nuestro mundo, podemos darnos cuenta de algunos motivos por los que los seres humanos han sido capaces de situarse como el sistema mejor adaptado y supuestamente más inteligente:

\begin{enumerate}
\itemsep=0px
\item Su capacidad para identificar, a partir de problemas o situaciones anteriores, una relación (si existiera) con el problema al que se están enfrentando y utilizar esta información a su favor.
\item Su capacidad para crear rápidamente nuevas estrategias acordes a los nuevos problemas a los que se enfrenten. El ser humano es capaz de construir, a partir de pocos intentos, una estrategia (aunque normalmente no óptima) que le permite resolver satisfactoriamente cada nuevo problema al que se enfrenta.
\item Su facilidad para relacionarse con distintos individuos y tomar ventaja de ello. Estas relaciones han permitido que los seres humanos cooperen para conseguir resolver con mayor eficacia los problemas a los que se han ido enfrentando durante su existencia.
%\item Su facilidad para intuir heurísticas que faciliten la resolución de los problemas. Cada vez que los seres humanos se enfrentan a un nuevo problema, en pocos intentos son capaces de darse cuenta de cuales fueron sus malas decisiones, a la vez que son capaces de intuir una estrategia (aunque normalmente no óptima) con la que conseguir resolverlo.
\end{enumerate}

Si estos son algunos de los motivos que han permitido a los seres humanos situarse como el sistema (supuestamente) más inteligente de nuestro planeta, resulta sensato que un sistema de evaluación de inteligencia proporcione (como mínimo) entornos en donde cualquier sujeto, en función de su capacidad sobre alguna o todas estas características, pueda resolverlos con mayor o menor eficacia.

Otro de los problemas de este sistema de evaluación, es que la aproximación del nivel de inteligencia que ofrece resulta poco fiable ya que, al parar la evaluación demasiado pronto es posible que el resultado obtenido diste de su inteligencia real. Además, este test debería ser capaz de medir de alguna forma la velocidad de aprendizaje\footnote{Esta velocidad no debería medirse con respecto al tiempo físico, ya que si dos máquinas tratasen de resolver un mismo problema utilizando la misma técnica (por ejemplo, fuerza bruta) no se demostraría mayor inteligencia si una terminase en menos tiempo por poseer un hardware más rápido. En su lugar, esta velocidad debería medirse más bien con respecto al número de interacciones empleados en su resolución.}, ya que, conforme menos tiempo requieran los sujetos para resolver nuevos problemas, más inteligentes serán.

A partir de esta discusión podemos concluir que como no todos los tipos de sistemas inteligentes tienen las mismas capacidades: (1) necesitamos una distribución que nos permita obtener entornos que se puedan adaptar mejor a los sujetos que estamos evaluando (una distribución adaptativa), (2) conseguir introducir entornos más generales que incluyan (entre otras cosas) algún tipo de comportamiento social y adquisición incremental de conocimiento y (3) una interpretación fiable de los resultados obtenidos durante la evaluación. A continuación vemos estas perspectivas.

\section{Mejor adaptación de la complejidad}
Con la forma actual de incrementar la complejidad de los entornos que se les proporciona a los agentes, se requerirá de mucho tiempo para llegar a sus niveles de inteligencia.

Para comprender mejor el alcance de este problema imaginemos, por ejemplo, que se están evaluando a dos tipos distintos de agentes: humanos y burros (por ser un animal supuestamente poco inteligente), y que los entornos que ofrece esta clase de entornos son laberintos cuya complejidad viene dada en función del número de paredes que posee; de modo que si existen $n$ paredes, entonces el laberinto será de complejidad $n$. Según la distribución utilizada (incrementar de 1 en 1 el número de paredes que posee el laberinto), se llegará (relativamente) rápido al número de paredes necesarias para evaluar a los burros. Sin embargo, los humanos también deberán ser evaluados en los mismos laberintos en los que fueron evaluados los burros, cuando los laberintos que poseen el número de paredes necesario para llegar a la inteligencia de los humanos se encuentran muy alejados a los que está realizando. Esta forma de proporcionar los entornos supone una pérdida de tiempo al tratar de evaluar a sistemas más inteligentes, ya que inicialmente deberán enfrentarse a problemas muy simples durante mucho tiempo.

Debido a esto, es necesario disponer de un sistema que adapte más rápidamente la complejidad de los entornos para converger más rápidamente a los niveles de inteligencia de cualquier sistema inteligente que se esté evaluando.

\section{Comportamiento social}
En \cite{AGI2011DarwinWallace} aparece una posible solución al problema de insertar comportamiento social en la clase de entornos. En este artículo se discute que, como resultaría una tarea de miles de años conseguir generar agentes con comportamiento social, una solución sería utilizar los propios algoritmos evaluados como agentes en los nuevos entornos. Utilizando estos agentes podemos disponer instantáneamente de comportamiento complejo en lugar de tener que esperar tantos años para conseguir algunos agentes, los cuales, con casi total probabilidad, serán muchísimo menos inteligentes. Además, con esta aproximación se conseguirá obtener un amplio número de agentes con complejidades muy variadas, que permitirán construir un amplio rango de entornos con muy distintas complejidades.

%Otro de los problemas que aun no se han tratado exhaustivamente en este proyecto, es una forma adecuada de medir los resultados de los agentes. Utilizar una media de las recompensas para evaluar el resultado de los agentes en un entorno parece bastante acertado, ya que, al desconocer cuándo terminará la evaluación, pone en compromiso a las estrategias completamente oportunistas y a las completamente explorativas, obligando a los agentes a utilizar estrategias híbridas a la vez que flexibles para interactúen en los entornos. Sin embargo, el problema surge al intentar comparar resultados de distintos entornos. Este problema nace de la distribución utilizada como referencia para obtener entornos. Si obtenemos un entorno y medimos su complejidad es posible que no se corresponda con su complejidad Kolmogorov real (véase \ref{sec:ComplejidadKolmogorov}) y, por lo tanto, sea realmente menos complejo que otro entorno obtenido con la misma complejidad de referencia. Este problema hace que dos resultados obtenidos en entornos con la misma complejidad no sean directamente equiparables, por lo que habrá que encontrar alguna forma de situar el resultado en un entorno en función de su complejidad real.

\section{Fiabilidad de los resultados}
Los tests adaptativos se van autoajustando en función de los resultados de los agentes, de modo que las complejidades de los ejercicios que se proponen van sufriendo grandes altibajos durante el comienzo del test, hasta que, finalmente, llegan al nivel del sujeto que están evaluando. Al igual que las complejidades de estos ejercicios, las estimaciones de la inteligencia obtenidas utilizando esta distribución sufrirán también grandes altibajos.

Este proyecto propone utilizar un valor con el que aproximar la inteligencia en función de los resultados obtenidos en los ejercicios. Sin embargo, solo con este valor no se puede aproximar su inteligencia de forma fiable.

\begin{figure}[h!]
\centering
\includegraphics[width=0.49\textwidth]{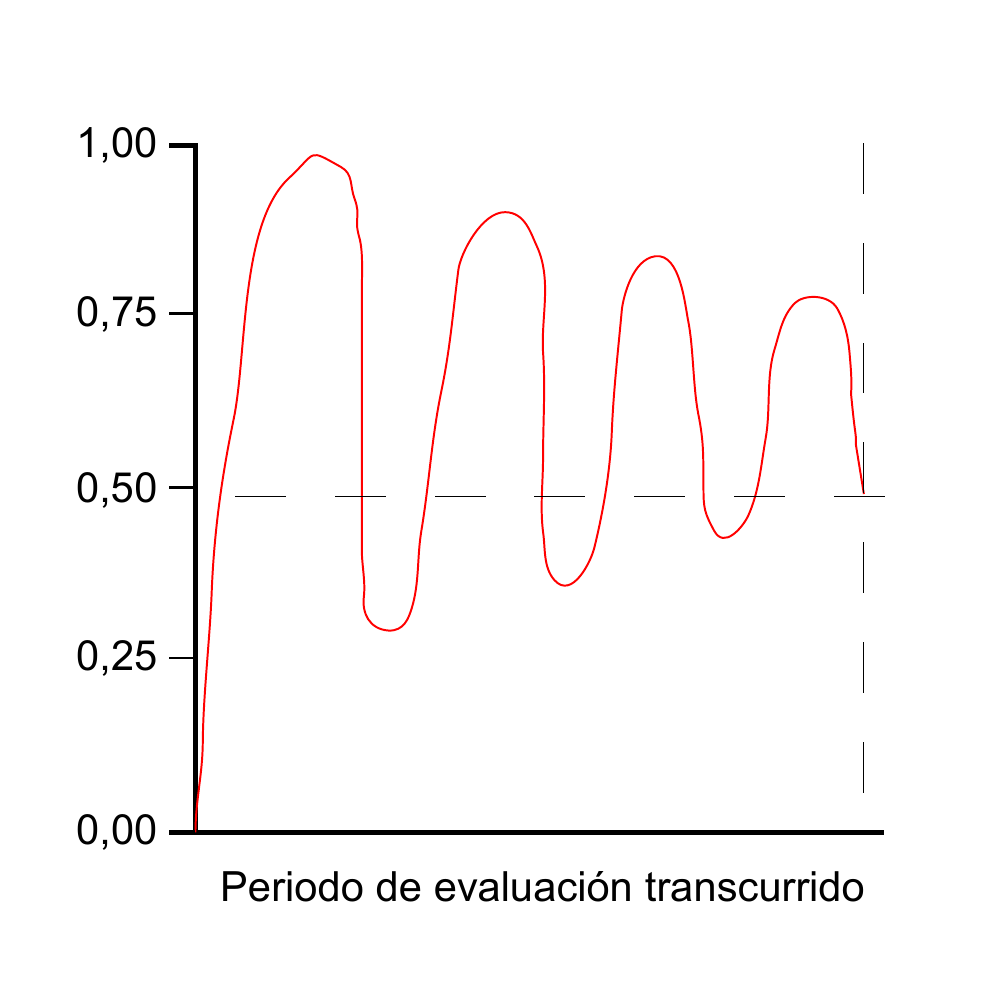}
\includegraphics[width=0.49\textwidth]{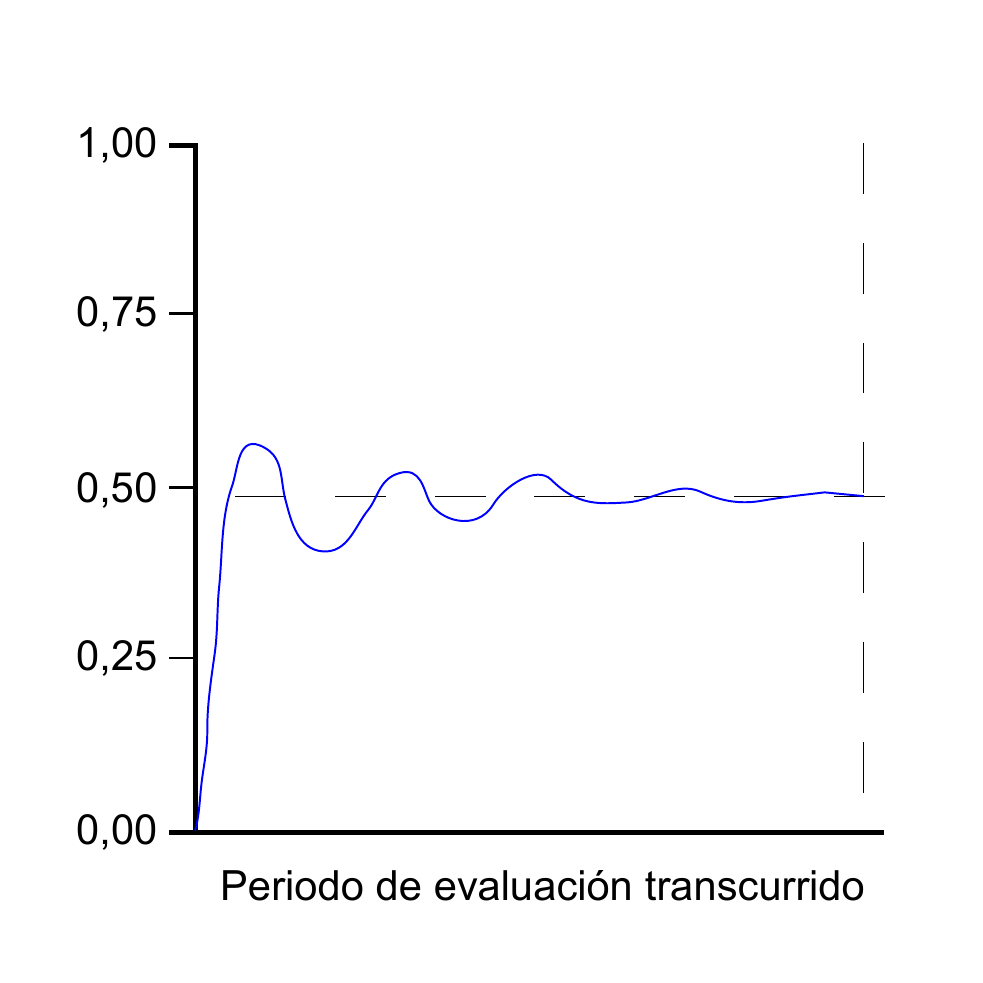}
\caption{Inteligencia estimada tras evaluar a dos agentes durante el mismo periodo. Izquierda: Resultado oscilante. Derecha: Resultado convergente.}
\label{fig:FiabilidadParada}
\end{figure}

Veamos dos situaciones que reflejan este problema: (1) Un agente se ha evaluado en varios entornos y, mientras sus resultados continúan oscilando se detiene su evaluación, obteniéndose así un resultado para su inteligencia de 0,47. (2) Otro agente ha sido evaluado en el mismo número de entornos (y posiblemente de tiempo) y claramente se puede observar que su resultado es más estable en torno a 0,47. Podemos ver ambas situaciones en la Figura~\ref{fig:FiabilidadParada}.

Obviamente no podemos asegurar que ambos agentes sean igual de inteligentes, aun obteniendo el mismo resultado en la misma cantidad de entornos.

\begin{figure}[h!]
\centering
\includegraphics[width=0.75\textwidth]{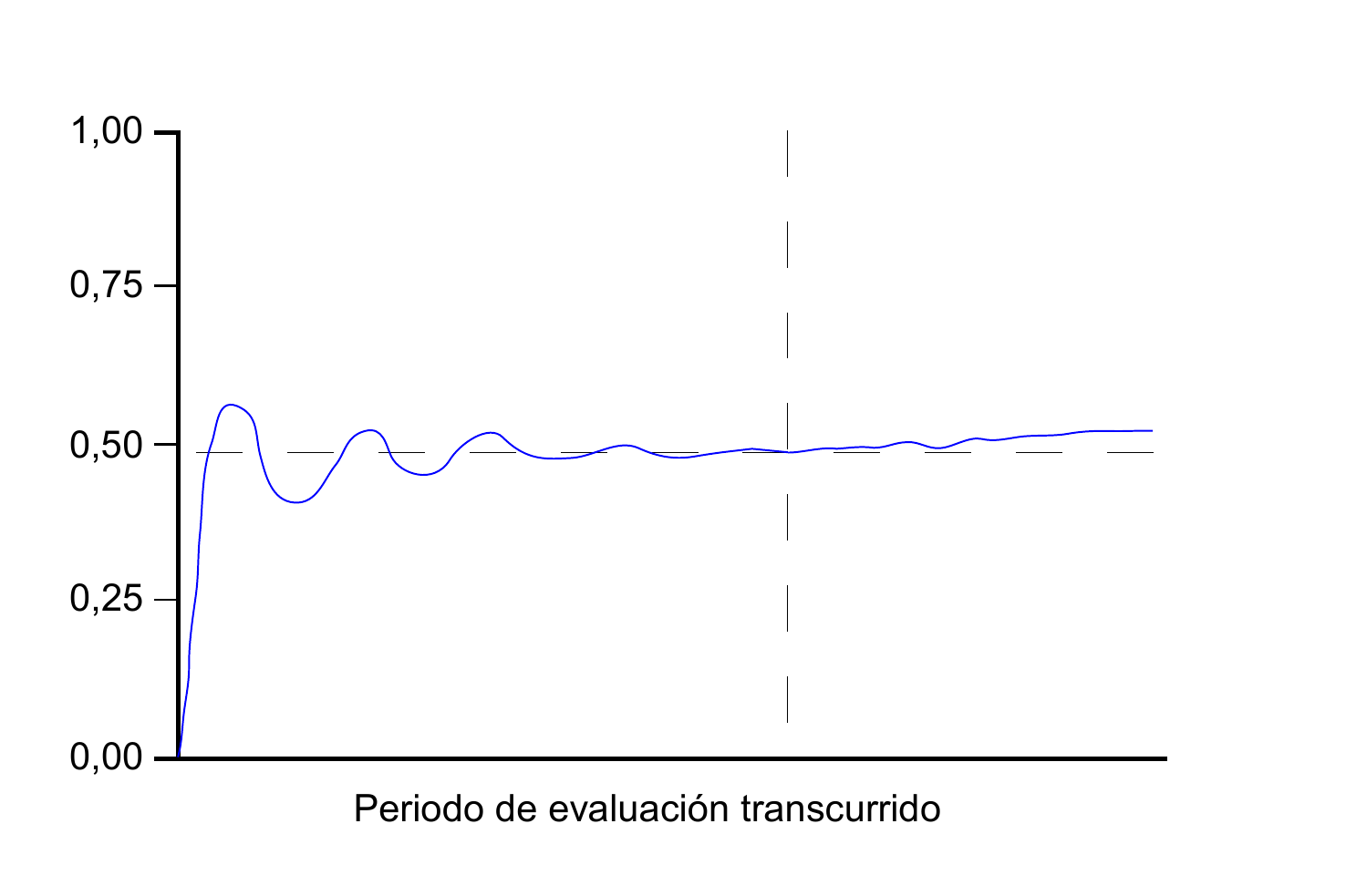}
\caption{Mejora de rendimiento creciente tras comprender mejor el funcionamiento de los ejercicios.}
\label{fig:FiabilidadEspecializarse}
\end{figure}

Imaginemos que al agente de la segunda situación se le sigue evaluando, es posible que sus resultados comiencen a mejorar lentamente ya que, al haber sido evaluado en tantos entornos, comienza a comprender mejor cómo obtener mejores resultados. Podemos ver esta situación en la Figura~\ref{fig:FiabilidadEspecializarse}.

Es de esperar que esta mejora paulatina acabe finalmente estabilizándose en un valor, ya que el agente no será lo suficientemente inteligente como para poder mejorar más en entornos muy complejos. No obstante, mientras tanto no se haya estabilizado completamente, el resultado obtenido no se corresponderá con su inteligencia real.

Necesitamos, además del valor aproximado de la inteligencia, un nuevo valor que nos indique lo cerca que se encuentra esta aproximación con respecto a su inteligencia real. Una posible solución sería calcular la varianza de los resultados, de modo que cuanto más se estabilicen los resultados de los agentes, más cerca se encontrará el valor aproximado de su inteligencia real. Algunas ideas desarrolladas en los IRT (Item Response Theory) podrían ser útiles para resolver esta falta de fiabilidad.

\paginablanco

\chapter{Conclusiones y trabajo futuro}\label{cap:Conclusiones}
\section*{Conclusiones}
Podemos analizar los resultados de esta tesis de máster conforme a dos perspectivas diferentes: (1) el cumplimiento de los objetivos marcados y (2) el conocimiento que se ha adquirido a partir del desarrollo del proyecto y de los experimentos.

\begin{enumerate}
\itemsep=0px
\item Se han cumplido los objetivos iniciales de la Sección~\ref{sec:Objetivos}.
\begin{itemize}
\itemsep=0px
\item La clase de entornos desarrollada ofrece resultados coherentes al evaluar un algoritmo de IA: Se puede apreciar como el algoritmo obtiene mejores resultados conforme va adaptando su comportamiento a los entornos. Además, tras cierto periodo (de interacciones o de tiempo), los resultados obtenidos por un algoritmo claramente convergen a un valor.
\item Los experimentos realizados nos muestran que el test no es capaz de discriminar correctamente a distintos tipos de sistemas en función de su inteligencia: Damos por sentado que los humanos son más inteligentes que el algoritmo de IA contra el que los hemos comparado. Sin embargo, este test no ha sido capaz de discriminar correctamente esta diferencia, por lo que (en su estado actual) falla al tratar de evaluar la inteligencia universal. 
\end{itemize}
\item Hemos adquirido nuevo conocimiento a partir del trabajo realizado en esta tesis de máster.
\begin{itemize}
\itemsep=0px
\item Es posible realizar el mismo test a distintos tipos de sistemas inteligentes sin necesidad de que sea antropomórfico: Ya que a cada tipo de sistema le resulta más natural interactuar con el test de manera distinta, lo único que hay que hacer es proporcionar una interfaz adecuada al sujeto que se va a evaluar, sin necesidad de cambiar el test en sí.
\item Tras evaluar un algoritmo de IA en distintos entornos con distintas complejidades, hemos podido ver que existe una relación (inversa) entre el rendimiento de Q-learning y la complejidad de los entornos y como este test es capaz de relacionarlo.
\item El cambio de distribuciones universales desde entornos pasivos a entornos interactivos se encuentra en la dirección correcta, pero no es, por sí sola, la solución. A partir de los resultados obtenidos, podemos ver que estos test son capaces de medir `algo' relacionado con la inteligencia, pero no podemos asegurar que lo que se mida sea realmente inteligencia, por lo que será necesario generalizar aun más estos tests para conseguirlo.
\item Estos tests necesitan una distribución (o clase de entornos) que ofrezcan una mayor probabilidad a la adquisición incremental de conocimiento, a las capacidades sociales y a entornos más reactivos. Sin embargo, tal y como se generan actualmente los entornos, resulta prácticamente imposible conseguir obtener alguna de estas propiedades (y aun menos todas).
\item Los entornos generados no permiten (en la práctica) utilizar todo el potencial de la inteligencia, por lo que será necesario generarlos siguiendo una nueva distribución que se vaya adaptando mejor al nivel de inteligencia de los agentes evaluados, proporcionando entornos más acordes a sus capacidades.
\end{itemize}
\end{enumerate}

\section*{Publicaciones}
Estas son las publicaciones generadas a partir de esta tesis de máster:

\begin{itemize}
\itemsep=0px
\item Evaluating a reinforcement learning algorithm with a general intelligence test \cite{CAEPIA2011Evaluating}.
\item Comparing humans and AI agents \cite{AGI2011ComparingHumansAI}.
\item On more realistc environment distribution for defining, evaluating and developing intelligence \cite{AGI2011DarwinWallace}.
\end{itemize}

\section*{Trabajo futuro}
A partir de las perspectivas vistas en el capítulo anterior y las conclusiones de este capítulo, podemos ver cual es el trabajo futuro que le depara a este proyecto.

\begin{itemize}
\itemsep=0px
\item Generar completamente los entornos de manera automática sin necesidad de ninguna intervención humana. Se deberán poder generar y construir automáticamente cualquier elemento de los entornos (espacios, objetos, agentes, \dots).
\item Construir los tests a partir de ejercicios, utilizando como base la complejidad de los entornos y los resultados de los agentes.
\item Estimar la fiabilidad de los resultados obtenidos.
\item Utilizar una distribución más adaptativa con la que converger mejor al nivel de inteligencia de los agentes.
\item Introducir automáticamente comportamiento social en los entornos, a partir de los algoritmos previamente evaluados.
\item Evaluar nuevamente distintos tipos de sistemas inteligentes y comprobar si los cambios realizados ofrecen resultados más coherentes.
\end{itemize}

\paginablanco

\appendix
\chapter{Diagrama de clases}\label{apx:DiagramaDeClases}
Para hacernos una idea general de cómo está diseñado e implementado el sistema de evaluación del proyecto ANYNT, aquí vemos el diagrama de clases de los propios tests. Además, este diagrama de clases también contiene las clases diseñadas en esta tesis de máster. Nos centramos únicamente en las clases más importantes.

\begin{figure}[h!]
\centering
\includegraphics[width=1\textwidth]{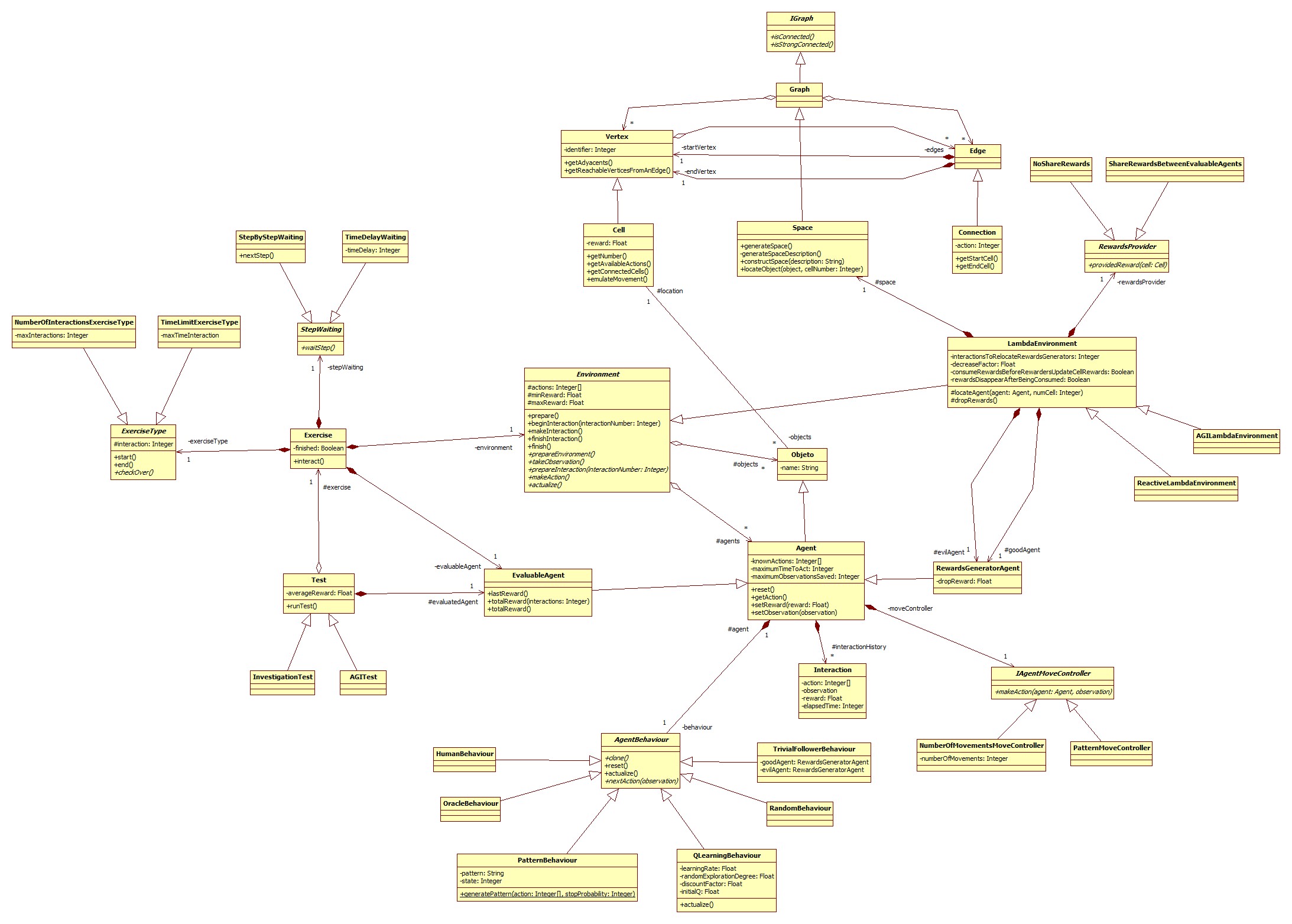}
\caption{Diagrama de clases completo de la estructura interna del sistema de evaluación.}
\label{fig:DiagramaClases}
\end{figure}

En la Figura~\ref{fig:DiagramaClases} podemos ver el diagrama de clases completo.

En los siguientes apartados vemos las clases más importantes implementadas: Test, Exercise, LambdaEnvironment, Space y Agent.

\section{Test}
Esta clase representa al propio test en sí. Tiene una referencia al ejercicio que se está realizando y al agente que se evalúa. Aunque aun no está completamente implementado, Test será el que estime la inteligencia de los agentes y se encargará de seleccionar la complejidad de los ejercicios de un modo adaptativo.

\begin{figure}[h!]
\centering
\includegraphics[width=1\textwidth]{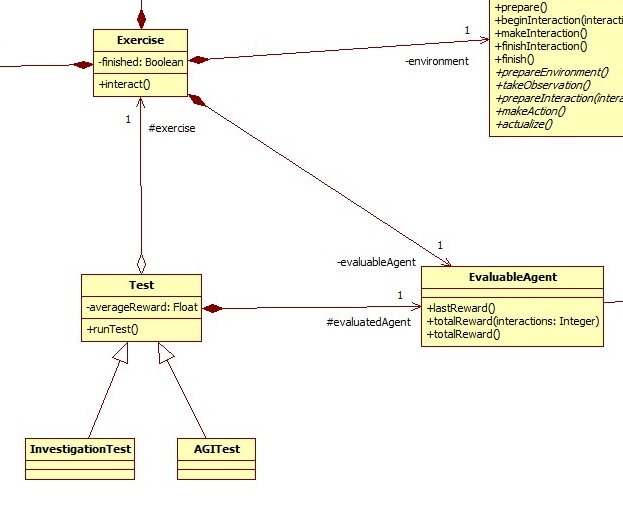}
\caption{Diagrama de clases mostrando únicamente a Test y las clases con las que se relaciona.}
\end{figure}

\section{Exercise (Ejercicio)}
Los ejercicios son los problemas a los que deberán enfrentarse los agentes durante el test. Estos ejercicios se componen del agente que se está evaluando y del entorno con el que deberá interactuar.

\begin{itemize}
\itemsep=0px
\item Estos ejercicios tienen ciertos parámetros que se pueden configurar:
\begin{description}
\itemsep=0px
\item[StepWaiting] Indica cómo se esperará para pasar a la siguiente interacción si el agente que se está evaluando no es humano.
\item[ExerciseType] Indica la condición que se tendrá en cuenta para dar el ejercicio por concluido.
\end{description}
\end{itemize}

\begin{figure}[h!]
\centering
\includegraphics[width=1\textwidth]{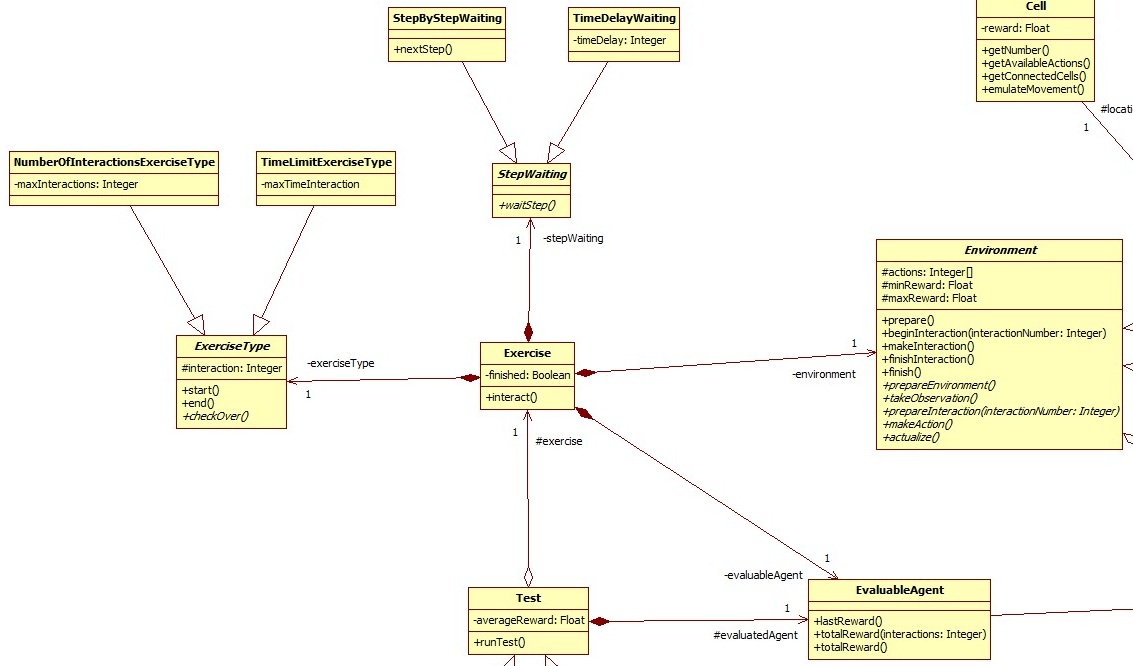}
\caption{Diagrama de clases mostrando únicamente a Exercise y las clases con las que se relaciona.}
\end{figure}

\section{LambdaEnvironment (Clase de entornos $\Lambda$)}
Esta es la clase de entornos $\Lambda$ diseñada. Cada uno de estos entornos está compuesto por un espacio y los agentes y objetos presentes durante el ejercicio. Como mínimo tendrá a los dos agentes especiales Good y Evil y al agente evaluado. Estos entornos tratan de ser lo más generales posibles, por lo que se pueden añadir (además de los tres agentes mencionados) todo tipo de objetos y agentes con los que interactuar.

\begin{itemize}
\itemsep=0px
\item Los entornos tienen ciertos parámetros que se pueden configurar:
\begin{description}
\itemsep=0px
\item[RewardsGeneratorAgent] Agentes que proporcionan las recompensas en el entorno (Good y Evil).
\item[RewardsProvider] Calcula la recompensa que reciben los agentes en función de la política que se haya elegido.
\item[DecreaseFactor] Factor por el que se dividirán las recompensas tras cada interacción.
\item[ConsumeRewardsBefore\dots] Las recompensas se consumen antes o después de ser actualizadas.
\item[RewardsDisappear\dots] Las recompensas desaparecen o no tras ser consumidas por un agente.
\end{description}
\end{itemize}

Además, esta clase puede extenderse para construir, por ejemplo, entornos reactivos, en donde los objetos reaccionasen a las acciones de otros objetos.

\begin{figure}[h!]
\centering
\includegraphics[width=1\textwidth]{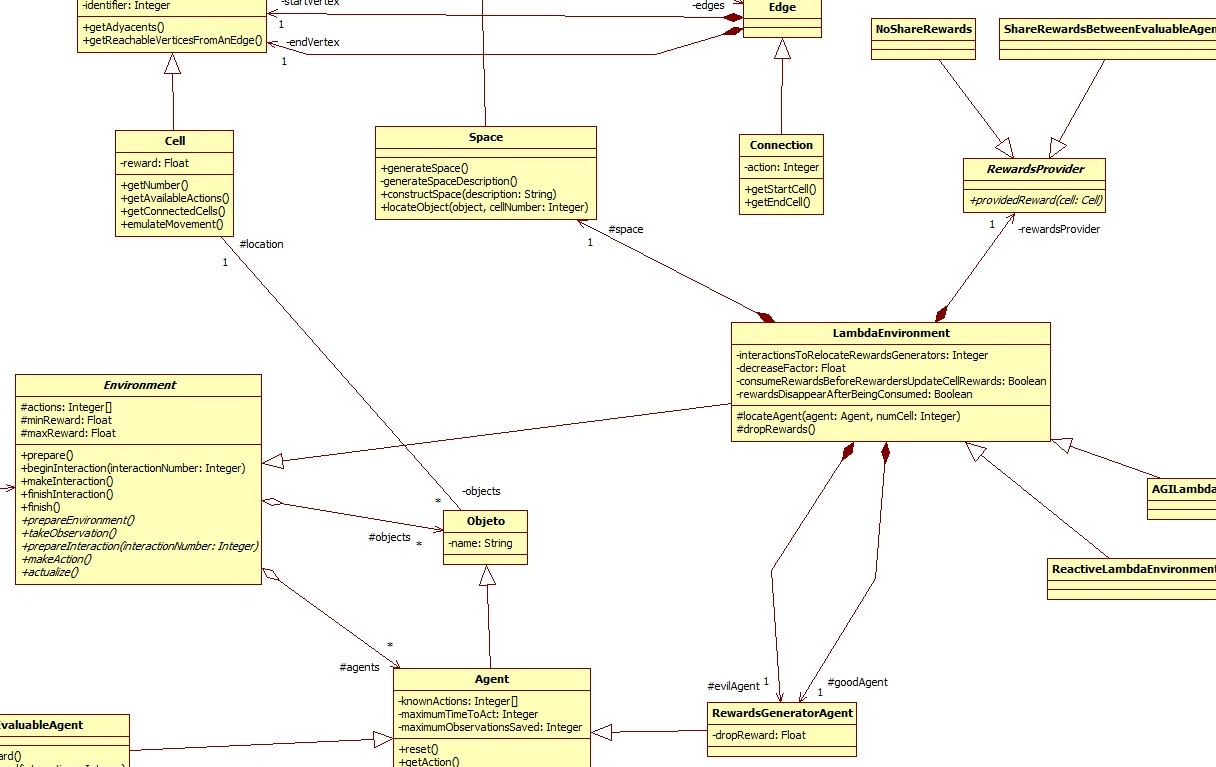}
\caption{Diagrama de clases mostrando únicamente a LambdaEnvironment y las clases con las que se relaciona.}
\end{figure}

\section{Space (Espacio)}
El espacio está construido como un grafo dirigido con sus vértices (celdas) y aristas (acciones/conexiones). Cada celda del espacio puede contener cualquier cantidad de objetos y contiene la recompensa de la celda.

\begin{itemize}
\itemsep=0px
\item Esta clase es la encargada de generar y construir la topología (celdas y acciones) del espacio, para ello utiliza las siguientes funciones:
\begin{description}
\itemsep=0px
\item[GenerateSpaceDescription] Genera la descripción de un nuevo espacio.
\item[ConstructSpace] A partir de la descripción de un espacio, construye su topologia.
\item[GenerateSpace] Genera un nuevo espacio utilizando las dos funciones anteriormente descritas.
\end{description}
\item Además, la clase podrá comprobar si el espacio construido a partir de la descripción generada está:
\begin{description}
\itemsep=0px
\item[Conectado] Todas las celdas están conectadas unas a otras a través de acciones.
\item[Fuertemente conectado] Existe un camino (siguiendo la dirección de las acciones) desde cualquier celda que conduce a cualquier otra celda del espacio.
\end{description}
\end{itemize}

\begin{figure}[h!]
\centering
\includegraphics[width=1\textwidth]{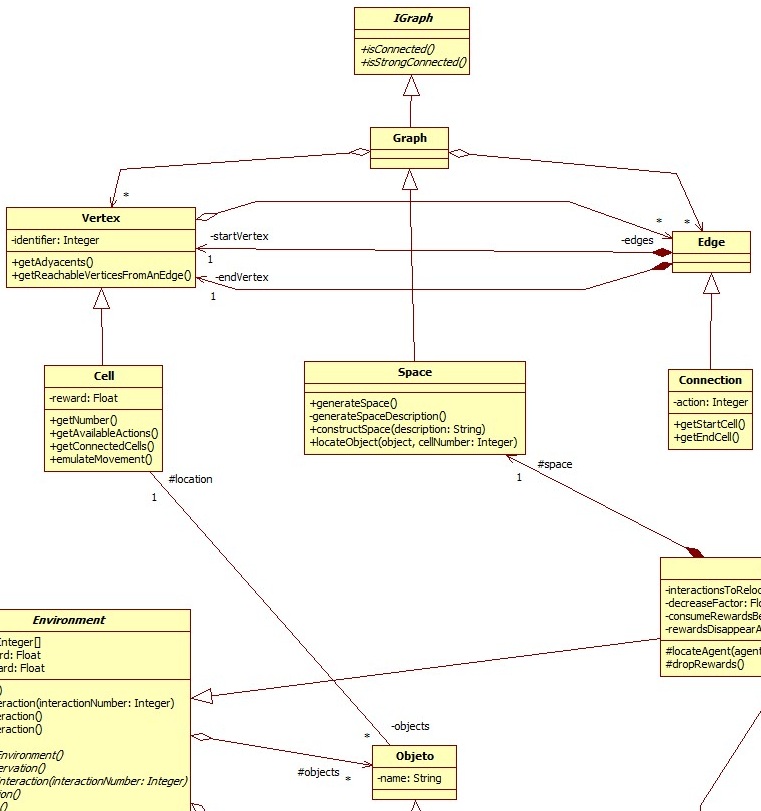}
\caption{Diagrama de clases mostrando únicamente a Space y las clases con las que se relaciona.}
\end{figure}

\section{Agent (Agente)}
Según nuestra aproximación, todos los agentes son objetos animados, por lo que tienen comportamiento propio con el que moverse a través del espacio.

\begin{itemize}
\itemsep=0px
\item Podemos ver los distintos tipos de agentes que existen:
\begin{description}
\itemsep=0px
\item[EvaluableAgent] Agente que se va a evaluar.
\item[RewardGeneratorAgent] Agentes recompensadores que dejan caer las recompensas en el entorno (Good y Evil).
\end{description}
\item Para poder configurar el comportamiento de los agentes de una forma más versátil, hemos dividido su comportamiento en dos clases:
\begin{description}
\itemsep=0px
\item[AgentBehaviour] A partir de la observación que se le proporcione, devolverá el siguiente movimiento a realizar.
\item[IAgentMoveController] Controla cuáles y cuántos de los movimientos que realizará el agente en cada interacción.
\end{description}
\end{itemize}

Además, el agente dispondrá de un historial con información de sus últimas interacciones. Se puede controlar la cantidad de interacciones que el agente es capaz de `recordar'; desde únicamente la última interacción realizada, hasta todas ellas.

\begin{figure}[h!]
\centering
\includegraphics[width=1\textwidth]{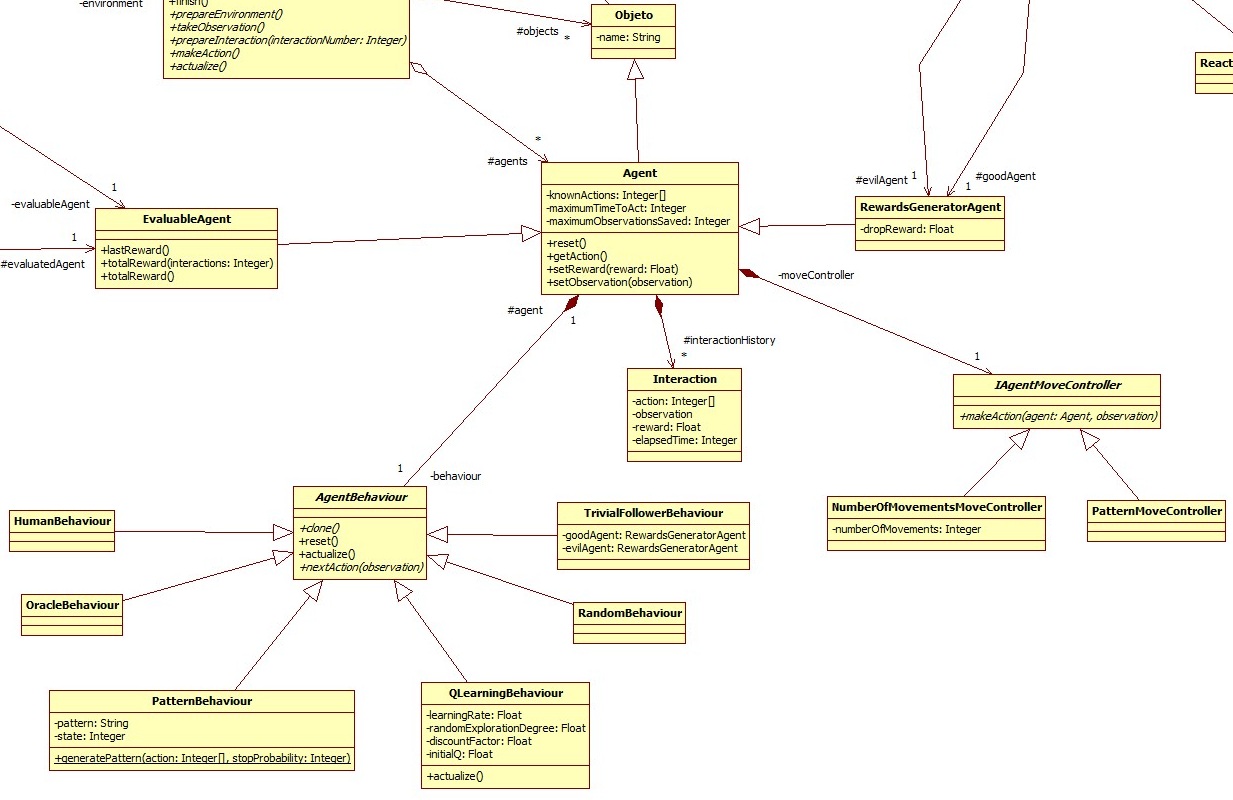}
\caption{Diagrama de clases mostrando únicamente a Agent y las clases con las que se relaciona.}
\end{figure}

\bibliography{biblio}
\bibliographystyle{plain}

% Termina el documento
\end{document}